\documentclass[final]{cvpr}
\usepackage{times}
\usepackage{epsfig}
\usepackage{graphicx}
\usepackage{amsmath}
\usepackage{amssymb}
\usepackage{cuted}

\usepackage{algorithm}
\usepackage{algpseudocode}
\def\code#1{\texttt{#1}}

\usepackage[pagebackref=true,breaklinks=true,letterpaper=true,colorlinks,bookmarks=false]{hyperref}

\def\cvprPaperID{4722} 
\def\confYear{CVPR 2021}

\newcommand\Mark[1]{\textsuperscript{#1}}

\usepackage{placeins}
\usepackage{stfloats}
\usepackage{multicol}

\begin{document}

\title{Magic Layouts:  Structural Prior for Component Detection \\in User Interface Designs}

\author{Dipu Manandhar\Mark{1}, Hailin Jin\Mark{2} , John Collomosse\Mark{1,2}\\
\Mark{1}CVSSP, University of Surrey, — Guildford, UK\\
\Mark{2}Adobe Research, Creative Intelligence Lab — San Jose, CA. \\
{\tt\small d.manandhar@surrey.ac.uk,  \{hljin, collomos\}@adobe.com}
}

\maketitle

\begin{abstract}

We present Magic Layouts; a method for parsing screenshots or hand-drawn sketches of user interface (UI) layouts.  Our core contribution is to extend existing detectors to exploit a learned structural prior for UI designs, enabling robust detection of UI components; buttons, text boxes and similar.  Specifically we learn a prior over mobile UI layouts, encoding common spatial co-occurrence relationships between different UI components. Conditioning region proposals using this prior leads to performance gains on UI layout parsing for both hand-drawn UIs and app screenshots, which we demonstrate within the context an interactive application for rapidly acquiring digital prototypes of user experience (UX) designs.

\end{abstract}

\section{Introduction}

User interface (UI) layout is a critical component in user experience (UX) design.  UI Layouts are commonly ideated and developed through sketched (`wireframe’) designs, or by mocking up screenshots.  Digital prototypes are then built using sequences of such layouts, to evaluate the UX and rapidly iterate on layout design.  The ability to quickly move from such prototypes ( sketches or screenshots) to digital prototypes in which components may be modified or rearranged, is valuable in expediting the design process.

This paper presents Magic Layouts; a technique for parsing existing UI layouts (for example wireframe sketches, or UI screenshots) into their UI components.  Our technical contribution is a deep learning method for detecting UI components within UI layouts that exploits common spatial relationships of components as a learned prior knowledge to improve detection accuracy.  

For example, UI elements often occur together and have a meaning underpinning that co-occurrence relationship.  A `text input field' and a `button' occuring side-by-side in a UI is often a  query-text and a response-button. We propose to explore the use of such co-occurrence information as an external knowledge graph to learn these component relationships, and incorporate this learning knowledge to boost the performance of state of the art detection algorithms.

\begin{figure}
    \centering
    \includegraphics [width=0.9\linewidth]{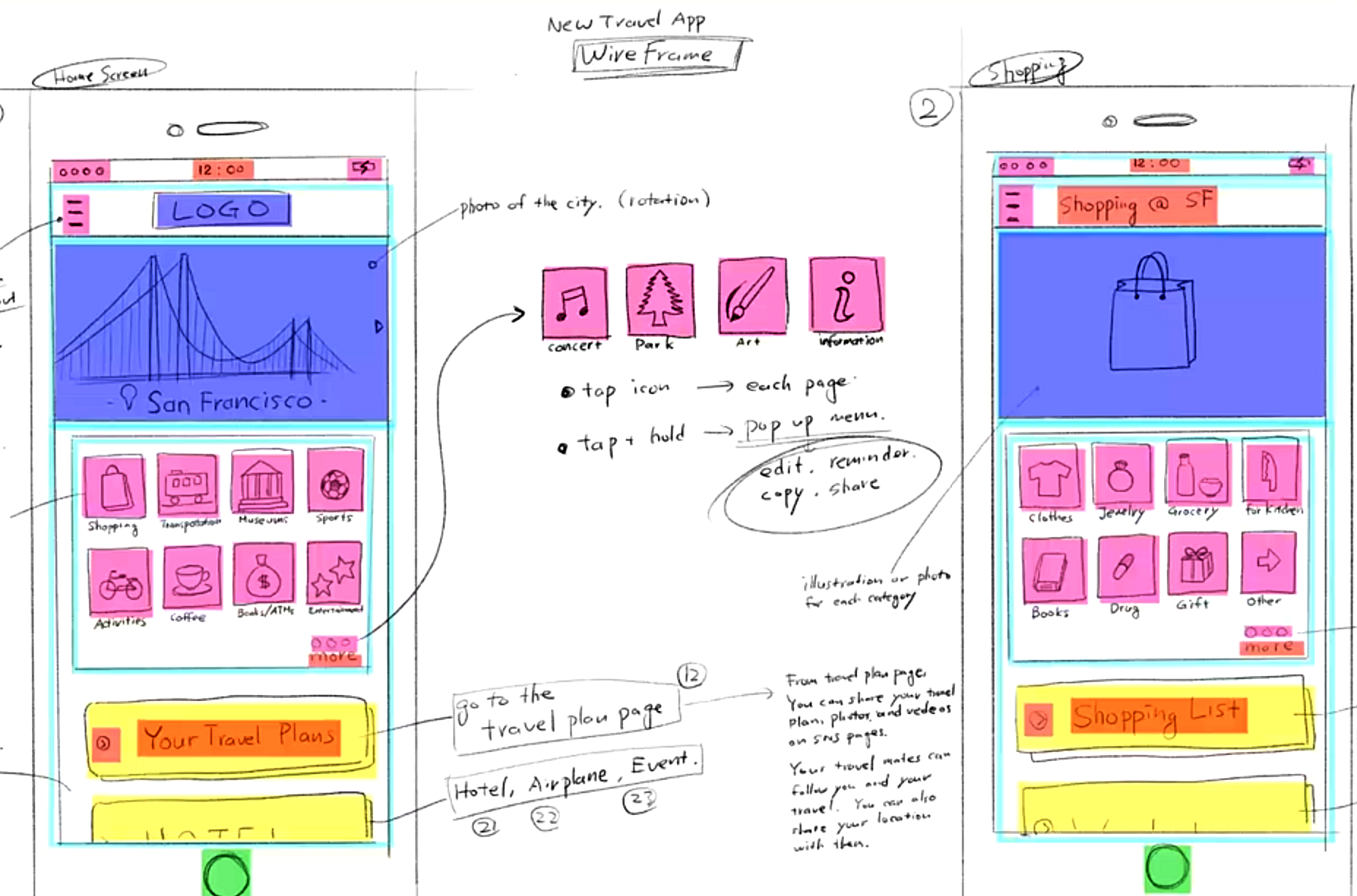}
    \caption{Magic Layouts parses UI layouts from sketched designs or app screenshots, exploiting learned prior knowledge of common component arrangements to improve recognition accuracy.  In this parsed example, colour  indicates different component classes.}
    \label{fig:teaser}
    \vspace{-5mm}
\end{figure}

We conduct experiments on two publicly available datasets of UI layouts; the RICO dataset of mobile app UX designs, and the DrawnUI dataset comprising hand-sketched UX wireframes.  Our proposed approach yields improvements in detection for modalities,  demonstrating that co-occurrences of UI components is a useful prior upon which to condition component detection and recognition when parsing UI layouts.  We incorporate our detection model into an interactive tool dubbed `Magic Layouts' capable of parsing UI layouts from mobile camera photographs of sketches (Fig.~\ref{fig:teaser}), or screenshots from mobile app stores.  Additionally, Magic Layouts incorporates sketch based image search to replace sketched graphics with higher fidelity artwork.

\section{Related Work}

Detection and recognition of objects within images is a long-standing computer vision problem. Classical detectors include sliding-window approaches \cite{Dollar2014},  super-pixel grouping \cite{Arbelaez2014,Uijlings2013} and  object proposal methods \cite{Alexe2012} often combined with sparse gradient features and dictionary learning for the recognition step.  With the advent of deep learning, simultaneous object localization and detection was initially explored via semantic segmentation \cite{long2015fully}, and region-based convolutional neural networks (R-CNN \cite{rcnn,faster-rcnn}) that classify a short-list of bounding boxes generated via selective search \cite{Uijlings2013}.  Region proposal networks (RPNs) were later fused with classifiers and trained end-to-end, to recognise candidates bounding boxes proposed with associated objectness scores in Faster-RCNN \cite{faster-rcnn}. Improvements upon Faster-RCNN included RetinaNet  mitigating foreground-background class imbalance \cite{retinanet2017}, and Mask-RCNN to detect and classify arbitrary shaped object regions \cite{maskrcnn2017} (unlike UI components). All these approaches make decisions locally, without consideration of neighbouring regions or image structure.  Recently, SGRN~\cite{xu2019spatial} aims to improve object proposal features by encoding spatially-related regions using Graph Neural Network (GNN). The SGRN graph encourages visual similarity and so spatially coherent labelling (\ie biased towards connecting similar objects of the same class). This differs from our goal of modelling frequently co-occurring arrangements of objects from different classes; common in UX designs. 

Layout has been studied from the perspective of synthesis, including automated reflow of banner adverts and graphic design \cite{Hurst2009}, steered by gaze-tracking \cite{Pang2016} or learned common design patterns \cite{ODonovan2014,ODonovan2015}.  Aesthetic score prediction for document layout has been modelled  \cite{Harrington2004} and used to drive automated  layout decisions  \cite{Geigel2001,Goldenbert2003}.  
UI Layouts specifically have been addressed through re-use of layouts via similarity search \cite{uist2018rico,Manandhar2020} leveraging Rico; a crowd-annotated dataset \cite{deka2017Rico} of mobile app screenshots.
Most closely related are works that learn design heuristics to parse screenshots for layout re-use \cite{Yang2017,Swearngin2018} or for code generation \cite{pix2code}.  All these techniques are bottom-up; driven by initial detection of individual UI components (\eg via Faster-RCNN or edge-grouping heuristics \cite{Moran2018}) which are post-processed and associated via learned (or designed) rules.  Whilst we also parse UI layouts, our technical contribution is to enhance accuracy of that detection step by integrating a prior for component co-occurence at the initial step \ie enhancing Faster-RCNN.
\section{Methodology}

\begin{figure*}
    \centering
    \includegraphics[width=0.75\textwidth]{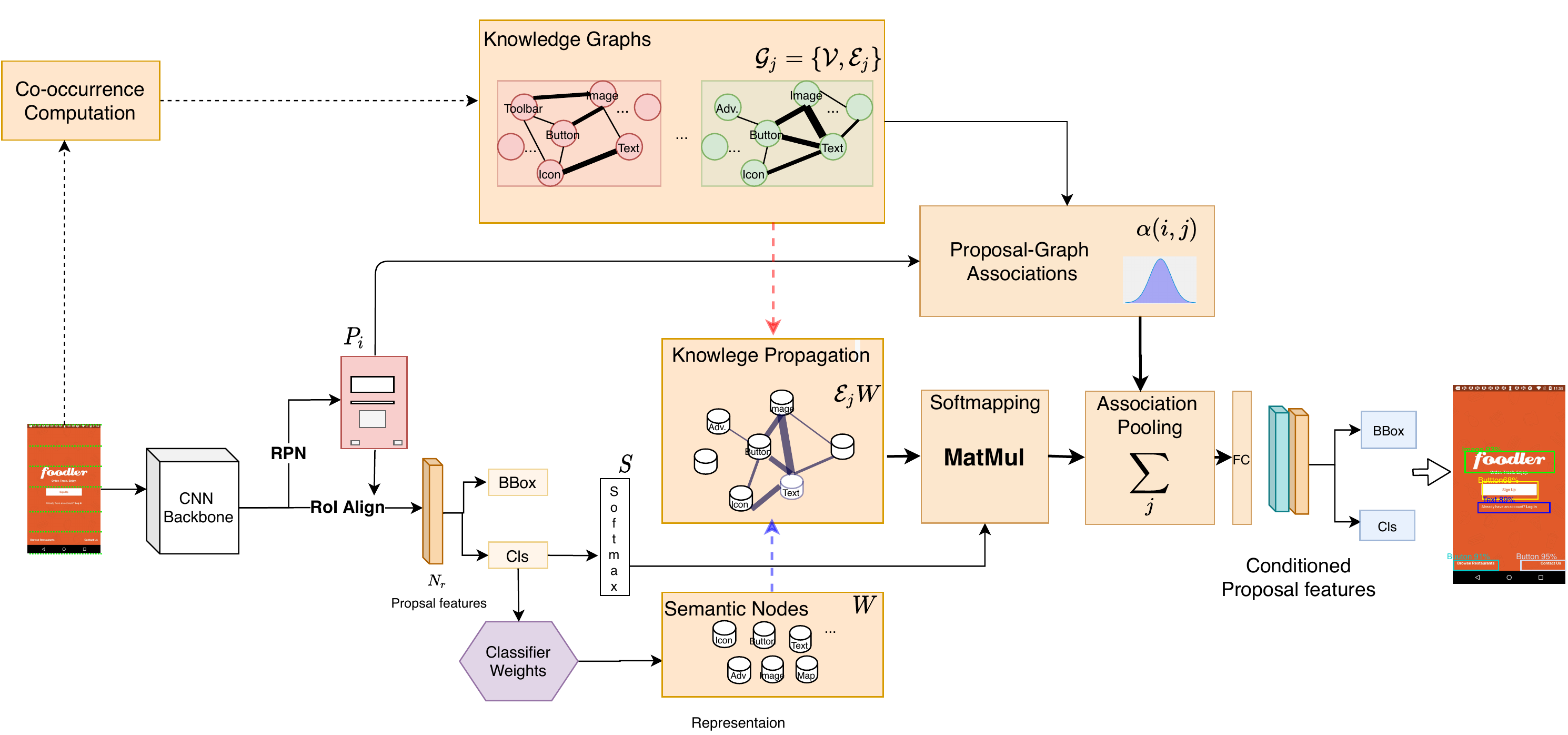}
    \caption{Magic Layout Architecture. Our framework exploits co-occurrence knowledge graphs (computed offline) as a prior to condition the proposal features. The semantic node representations obtained from the classifier are used to propagate features along the edges of knowledge graphs which are further soft-mapped to region features. Conditioned proposal features are then obtained by pooling the features based on proposal-graph associations eventually producing better UI detections.} 
    \label{fig:arch}
    
\end{figure*}

We introduce \emph{Magic Layouts} that exploits a learned structural prior for user interface (UI) layout parsing. Fig.~\ref{fig:arch} shows the architecture of the proposed framework.  We propose to condition region proposals  using a structural prior which essentially encodes common spatial co-occurrence among UI components distributed over various UX regions as knowledge graphs. To this end, we learn co-occurrence graphs from various UX regions and use high-level semantic representations that are readily available in the network to propagate them through the graphs. Representations from different regions are aggregated based on the proposal-graph associations. We show that such representations when integrated with original features offer more accurate UI parsing for both app screenshots as well as hand-drawn UI layouts (subsec. \ref{ssec:eval}). The following sections describe our approach to learning the prior and how this information is embeded into the network.

\subsection{Object Co-occurrence as Knowledge Graph}
\label{ssec:comat}

UX components usually co-exist together to form layout designs and have semantics  underpinning the co-occurrence. We propose to explore such  information and integrate this into detection frameworks to enhance  UI detection. Formally, let $\left\lbrace c_i\right\rbrace_{i=1}^{C}$ be set of UI components $C$ being the total number of component classes. We  aim to obtain knowledge graphs $\boldsymbol{\mathcal{G}} = \lbrace \boldsymbol{\mathcal{V}}, \boldsymbol{\mathcal{E}} \rbrace$  where the vertices $\boldsymbol{\mathcal{V}}$ represent  UI classes and edges $\boldsymbol{\mathcal{E}}$ encode common spatial occurrence information from a training set, and use that prior  for both  training and inference.

The co-occurrence statistics among UI components vary across different regions of UX especially along the vertical direction - for example, the distribution statistics of components in the top of UX (often consisting of \emph{Toolbar, Multi-tabs} etc.) may differ from that in the bottom which may consists of emph{Advertisement}or  \emph {Exit} controls. Moreover, UXs are usually scrolled vertically and component relations are often within local regions. In view of this, we propose to estimate co-occurrence information in local regions of layouts. To achieve this, we divide the UI into several horizontal bands and observe  frequency statistics of  co-occurring classes. Note that the bands can be designed to be disjoint or overlapping in a sliding-window fashion. Let $N_b$ be the number of bands that divide UX layout. We initialise  $C\times C$ graph for each band with  edge values  $e_{mn} \in \boldsymbol{\mathcal{E}} =0, \forall m,n \in \lbrace 1, \cdots, C \rbrace$.  A component is associated to a band if its center lies inside the upper and lower bounds of the band. We count a hit and increment the value of edge $e_{mn}$  by 1 if two components from class $m$ and class $n$ both lie in the corresponding band. Algorithm~\ref{algo:compute_comat} summarises co-occurrence graph computation. This process yields  graphs corresponding to the various bands in the UX. The obtained graphs are  row-column normalised  $e_{mn} := \frac{e_{mn}}{\sqrt{\sum_n e_{mn} \sum_m e_{mn}}}$. Finally, we obtain $N_g (=N_b)$ graphs that carry component co-occurrence information distributed across various regions in UX layout. We leverage these knowledge graphs to condition the proposal features for better UI detection as described in following sections.

\begin{algorithm}
\caption{Co-occurrence graph computation \label{algo:compute_comat}}
{\footnotesize	
\begin{algorithmic}[1]
\Statex \textbf{Input: } $N$ Bounding boxes $bb=\lbrace bb_i\rbrace_{i=1}^{N}$ and their labels $L=\lbrace l_i\rbrace_{i=1}^{N} $ for all training UXs; Width $W_{b}$, number of bands $N_{b}=N_{g}$ 
\Statex \textbf{Output: } Co-occurrence graphs $\lbrace\mathcal{G}_j = (\mathcal{V}_j, \mathcal{E}_j)\rbrace; j= \lbrace 1, \cdots, {N_g}\rbrace$
\ForAll {UX in training set}
	\State Get UX height $H$
	\State Compute Bands $\boldsymbol{S} = \lbrace S_j \rbrace_{j=1}^{N_b}$ with upper and lower bounds:
	\Statex	\hphantom{xxxxx}(a) $Upper = $ \code{range}$(0,H, H/N_{b})$
	\Statex	\hphantom{xxxxx}(b) $Lower = Upper + W_{b}$
	\State Compute matrix $M$ \textbf{s.t.} $M[i,j]=1$ \textbf{if} $bb_{i} $ in band $S_j$ \textbf{else} $0$
	\State $M := M [:, \code{sum}(M,0) > 1]$  \Comment{Bands with co-occurrences} 
	\ForAll {$bb_i$ in bb}
		\State $S(bb_i) = \left\lbrace S_j \right\rbrace \in \boldsymbol{S}$ \textbf{s.t.} $M[i,j] \neq 0 $ \Comment{Band where $bb_i$ lies}
		\State $coind = \left\lbrace i\right\rbrace$ \textbf{s.t.}  $M[i, S(bb_i)] \neq 0 $   \Comment{Co-occurrence index}
		\State $coind = \code{Unique}(coind)$   \Comment {Remove duplicates}
		\State Get class labels for $bb_i$ and $coind$: $L_i$ and $L_{coind}$
		\State $\mathcal{E}_j[L_i,L_{coind}] +=1$  \Comment{Update edges}
	\EndFor
\EndFor
\State $e^{j}_{mn} := \frac{e^{j}_{mn}} {\sqrt{\sum_n e^{j}_{mn} \sum_m e^{j}_{mn}}} ;\forall j$ \Comment{Normalisation}
\State $e^{j}_{mm} := 1 \;\forall m, j $
\end{algorithmic}
}
\end{algorithm}

\subsection{Semantic node representation}
Our aim is to enrich the proposal  representation with the learned co-occurrence knowledge graphs. We first need to define node features that would be propagated through the edges of the graphs. Regions and proposals are often represented using appearance features within an individual image \cite{dai2017detecting, chen2018iterative, marino2017more}. However, such representations may not be robust when there are overlapping and nested objects that lead to heavy occlusions which is often observed in UI layouts. Moreover, visual ambiguities among various components can lead to ineffective or even wrong propagation. Recently, few/zero-shot methods \cite{wang2018zero,gidaris2018dynamic} and object recognition \cite{xu2019spatial,xu2019reasoning} have used the classifier weights as a visual embedding for unseen classes and the proposal's latent representation to guide recognition. Motivated by this, we use classifier weights as semantic node feature of the graphs. In particular, to obtain this representation, we copy the weights of the previous classifier head of the base network including the bias \ie $\mathbf{W}\in \mathbb{R}^{C\times (D+1)}$ where $D$ is input dimension to the classifier head and $C$ is the total number of UI classes. The use of this representation comes with three main advantages: (i) the representation captures high-level semantics which acts as class embedding for each category, (ii) they are readily available without requiring computationally expensive  feature averaging or clustering over large samples \cite{lee2018cleannet}, and  (iii) the representations are dynamically updated during training thus they improve over time. 

\subsection{Knowledge graph-based proposal conditioning}
The co-occurrence knowledge graphs contain component relationship information across different regions of UX layouts. We associate each proposal to the learned knowledge graphs in order to propagate their representations through their respective edges. A natural rule of proposal-to-graph assignment can be associating each proposal to its nearest band (and hence the corresponding graph) or to the band that encloses the proposal. However, this single hard-assignment may be too strict and can be noisy as proposal boxes are only initial estimates of objects which are essentially regressed for the final predictions. Thus, we propose to assign proposals to multiple graphs in a weighted manner; as a Gaussian function of their spatial proximities. 

\begin{figure*}
    \centering
    \includegraphics [width=0.75\textwidth]{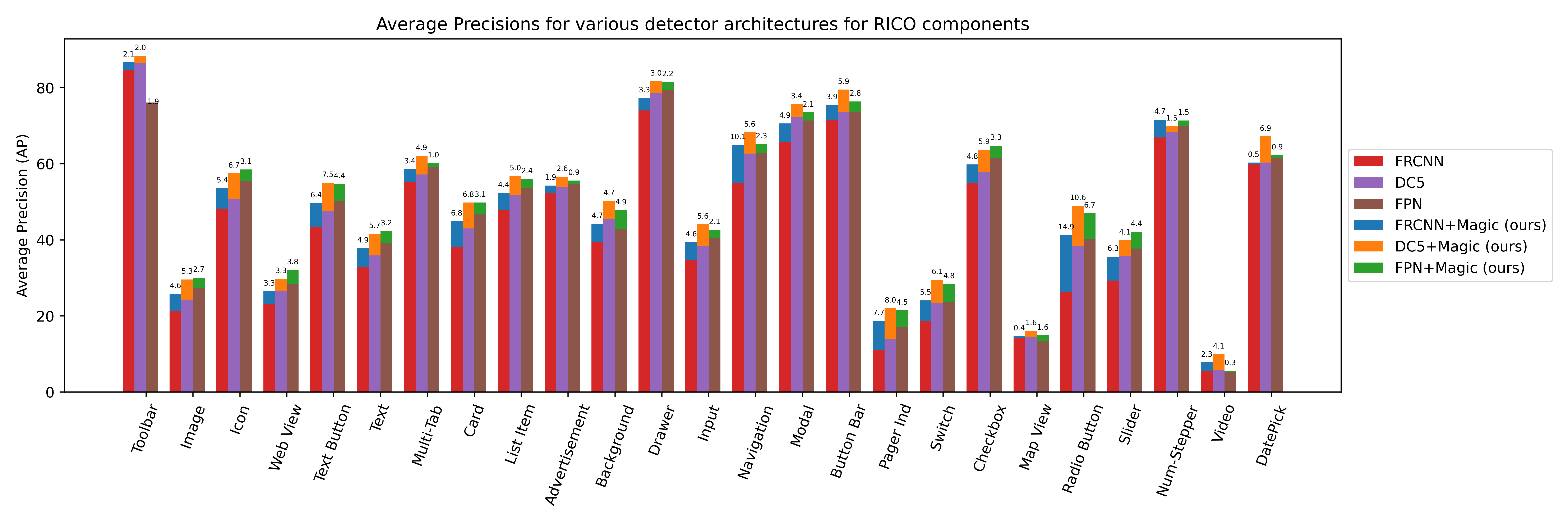}
    \caption{Average precision (AP) for RICO UI categories for baselines and the proposed method using various architectures: Faster-RCNN (FRCNN) \cite{faster-rcnn}, DC5 \cite{fpn2017} and FPN \cite{fpn2017}. For each network, the AP obtained using our proposed method are shown as stacked bars over their corresponding baselines where the figures on tops show the absolute improvements for each category (zoom-in for best view).}
    \label{fig:bars_rico}
\end{figure*}

\begin{table*} [!h]
\caption{\textbf{Performance comparison on RICO dataset}}
\label{tab:comparison_rico_complete} 
\centering
\scriptsize{
        \begin{tabular}{|l||  c c c||  ccc|| ccc|| ccc|| }
        \hline
        
		Method  			& AP 	&AP50	&AP75 	&APs 	&APm 	&APl 	&AR 	&AR 	 &AR   &ARs  &ARm   &ARl\\ 
		@IoU				&0.5:95 	&0.5 	&0.75   &0.5:95 &0.5:95 &0.5:95 &0.5:95 &0.5:95 &0.5:95   &0.5:95 &0.5:95 &0.5:95 \\
		maxDets				& 100	&100	&100	&100	&100	&100	&1		&10		&100  &100	&100	&100	\\	\hline \hline	 
		
		Faster-RCNN \cite{faster-rcnn}		&43.0	&52.8	&46.0	&2.3	&16.5	&43.2	&40.5	&58.1	&60.1   &5.7    &29.2   &60.0	\\
		RetinaNet \cite{retinanet2017}	 	&41.7	&52.8	&45.8	&\textbf{3.6}	&20.5	&42.0	&38.8	&57.0	&59.1   &7.2    &34.7   &58.8	\\ 
			
        FRCNN+SGRN \cite{xu2019spatial}  	&45.2	&54.2	&47.8	&2.4	&19.0	&45.8	&41.4	&60.0	&62.1   &6.4    &32.3   &62.0 \\
        \textbf{FRCNN+Magic}		    &\textbf{47.8}	&\textbf{57.4}	&\textbf{51.0}	&3.1	&\textbf{21.7}	&\textbf{48.4}	&\textbf{42.6}	&\textbf{61.9}	&\textbf{63.9 }  &\textbf{6.9 }   &\textbf{34.9 }  &\textbf{63.9}	\\	
		\hline 
		
		DC5 \cite{fpn2017}			&46.7  	&56.2	&49.8 & 2.9 &20.5 &47.2 &41.9 &60.7 &62.7 &6.7 &33.0 &62.7  \\ 
        DC5+SGRN \cite{xu2019spatial} 					&49.0	&58.0	&51.8	&\textbf{4.9}	&27.5	&49.6	&43.0	&61.8	&63.9   &\textbf{8.4} &40.5  &63.9\\
        \textbf{DC5+Magic} &\textbf{51.8}	&\textbf{61.1}	&\textbf{54.9}	&4.4	&\textbf{29.9}	&\textbf{52.2}	&\textbf{44.8}	&\textbf{64.8}	&\textbf{66.7 }  &8.1    &\textbf{41.6}   &\textbf{66.9} \\
		\hline 

		FPN	\cite{fpn2017}    				&47.6   &57.1	&50.4 	&4.6 	&30.6 	&47.7	&41.6	&61.0	 &63.1  &9.5    &44.6   &62.6\\

        FPN+SGRN \cite{xu2019spatial} 			        &49.9	&59.6	&52.7	&7.6	&33.5	&50.0 &42.6	&62.4	&64.5 &\textbf{13.5} &45.8 &64.0 \\	
        \textbf{FPN+Magic}		            &\textbf{50.3}	&\textbf{60.1}	&\textbf{53.4}	&\textbf{8.4}	&\textbf{34.7}	&\textbf{50.2}	&\textbf{43.0}	&\textbf{63.0}	 &\textbf{65.0 } &13.2   &\textbf{46.4}   &\textbf{64.5} \\ 		\hline

        \end{tabular}
}        
\end{table*}

Formally let $\left\lbrace P_i\right\rbrace_{i=1}^{N_r}$ be $N_r$ region proposals with features $f$ and bounding boxes  $\left\lbrace x_{i1},y_{i1},x_{i2},y_{i2}\right\rbrace_{i=1}^{N_r}$. Similarly, we have $\left\lbrace\boldsymbol{\mathcal{G}}_j\right\rbrace_{j=1}^{N_g}$ knowledge graphs related to $N_b(=N_g)$ bands from different regions of UX (sec.~\ref{ssec:comat}). 
We compute the association $\alpha(i,j)$ between proposal $P_i$ to graph $\boldsymbol{\mathcal{G}}_j$ using the following equations
\begin{align}
    \alpha(i,j) &= \frac{1}{\sqrt{2\pi\sigma^2}} \exp^{-\frac{1}{2} \left( \frac{\delta_i^j - \mu}{\sigma} \right)^2},\\
    \delta_i^j &= \frac{yc_i - yb_j}{H} ;
\end{align}
where $\delta_i^j$ is  the vertical displacement between the proposal $P_i$ and band $j$; and $\sigma$ and $\mu$ are the parameters of Gaussian distribution. Similarly,  $yc_i$ and $yb_j$ are y-components of centriods of the proposal and the band given by  $yc_i = (y_{i1} + y_{i2})/2$ and $yb_j = (y_{j,{uppper}} + y_{j,{lower}})/2$ respectively; and $H$ is the height of the UX which normalises the displacement taking care of varying UX dimensions. We further normalise the association values $\alpha(i,j) := \frac{\alpha(i,j)}{\sum_j \alpha(i,j)}$. 
\begin{figure*}[t!]
    \centering
    \begin{tabular}{ccccc}
      \includegraphics[width=2.75cm]{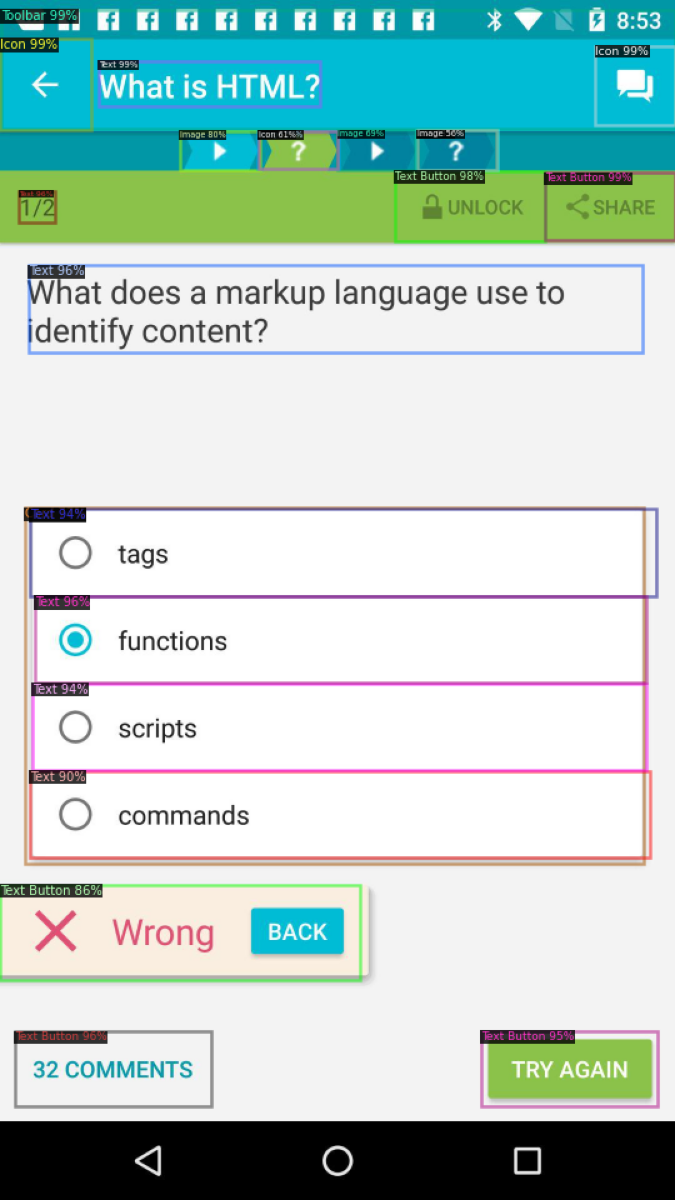} &  \includegraphics[width=2.75cm]{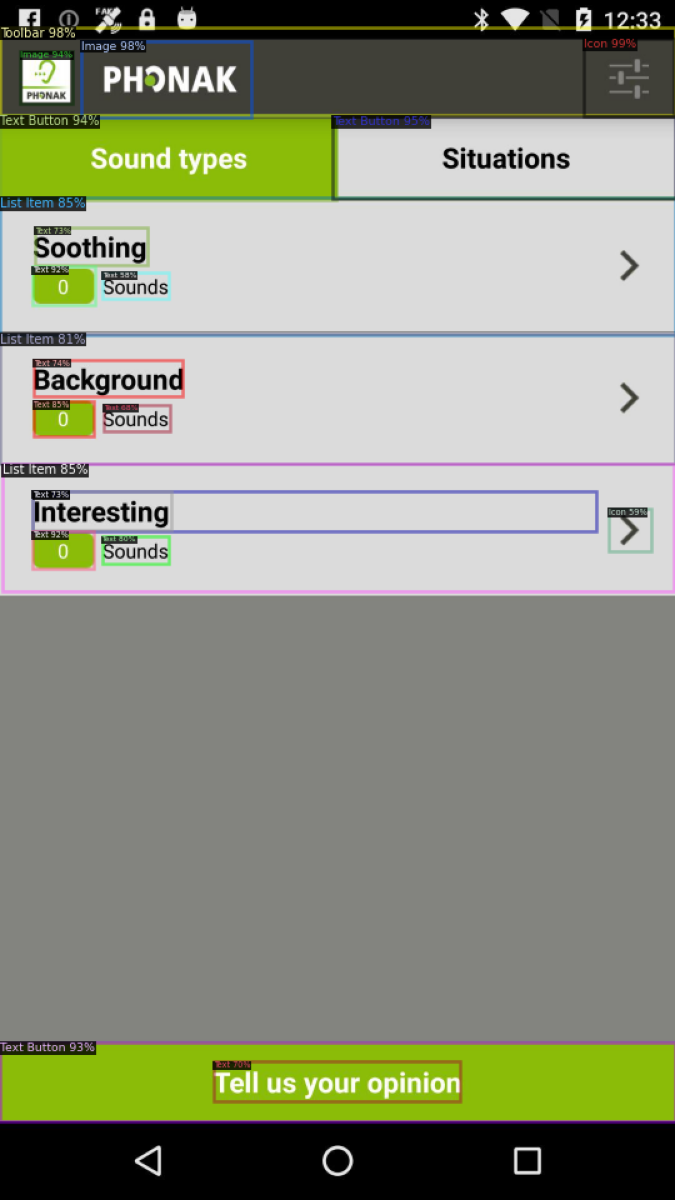} &
      \includegraphics[width=2.75cm]{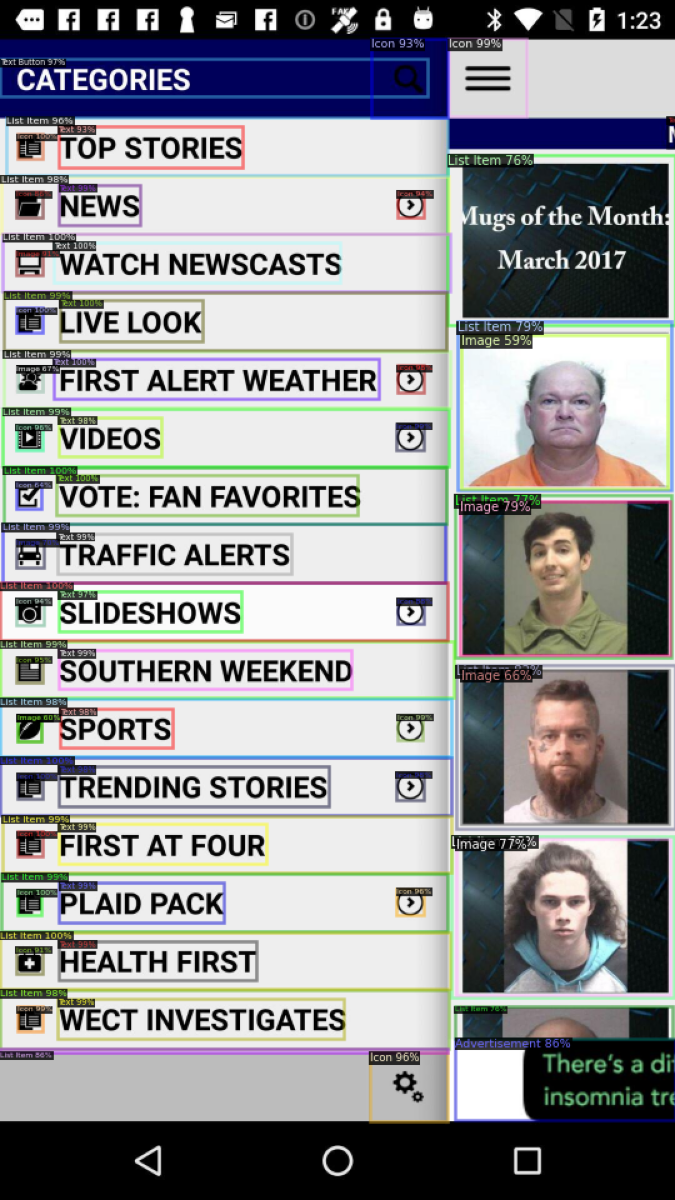} &       \includegraphics[width=2.75cm]{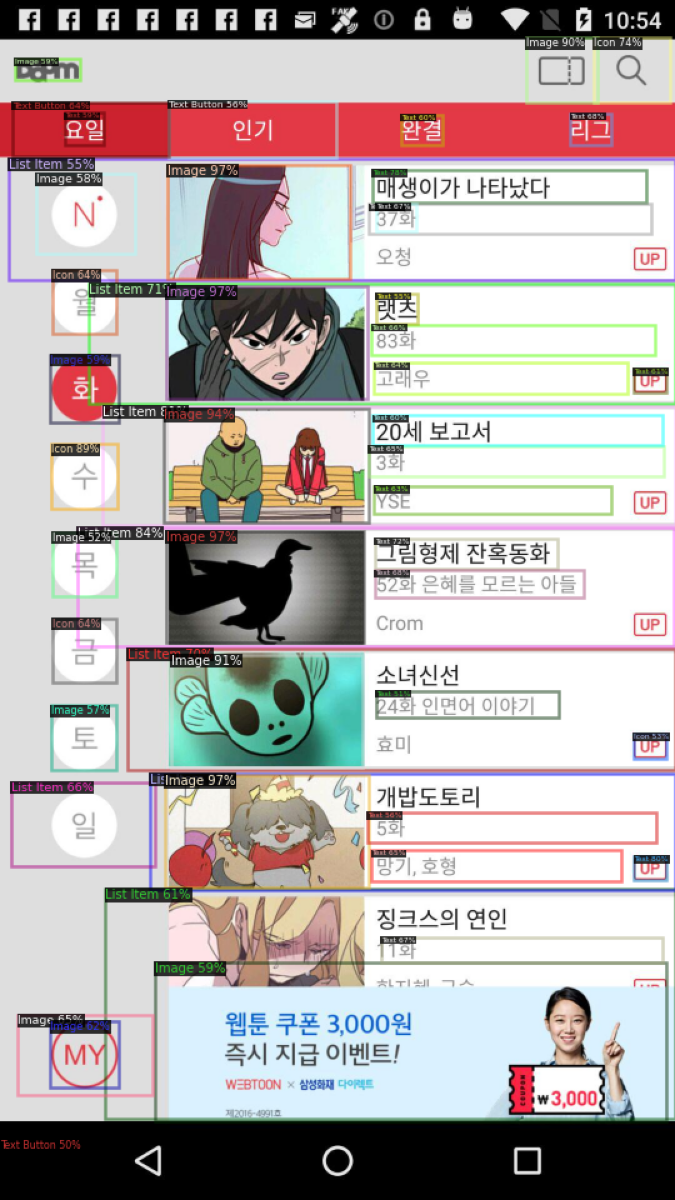} &
      \includegraphics[width=2.75cm]{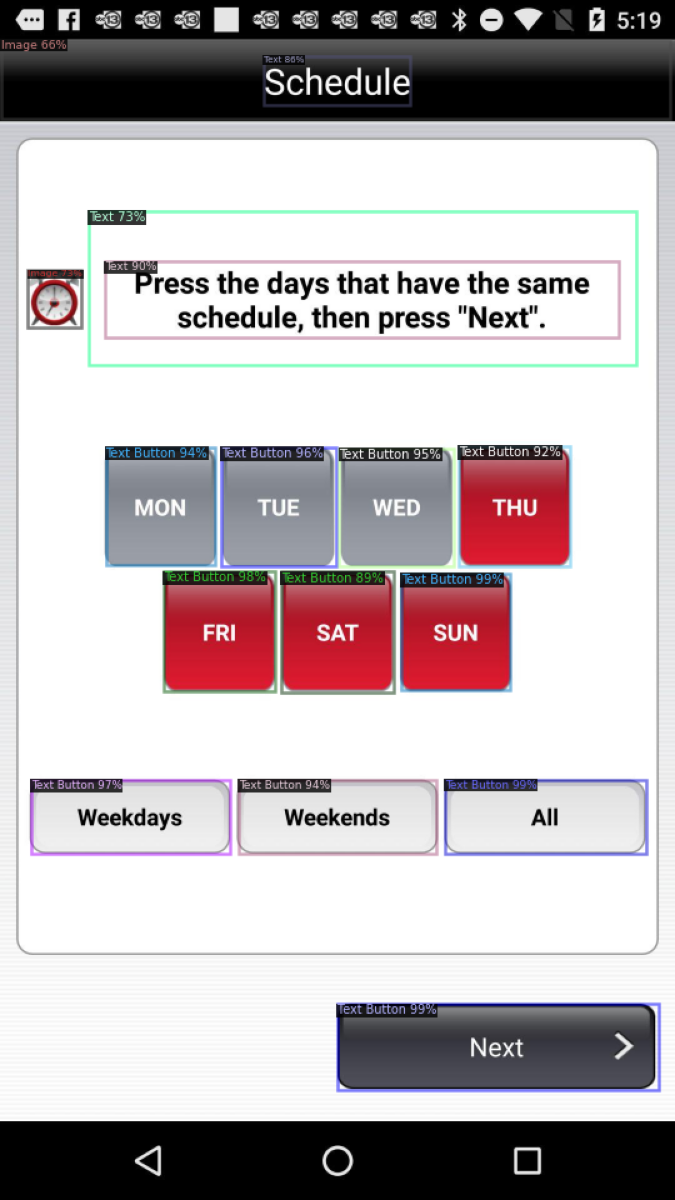}  \\
      \multicolumn{5}{c}{\small{(a) }} \\
      \includegraphics[width=2.75cm]{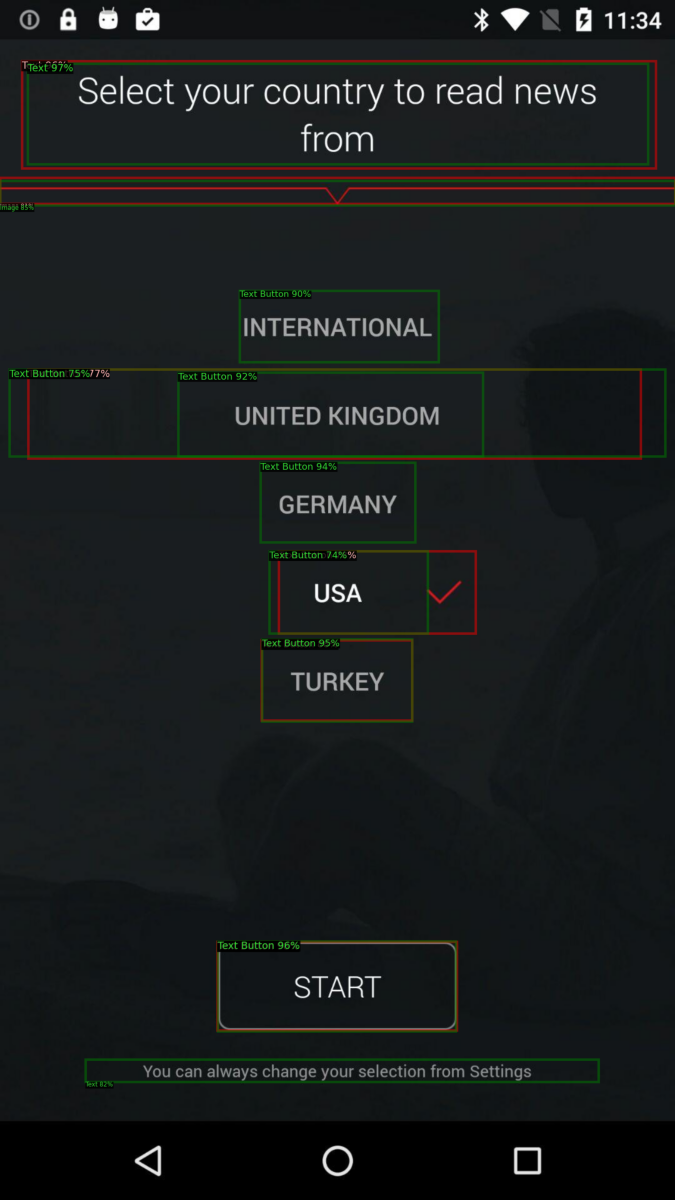} &  
      \includegraphics[width=2.75cm]{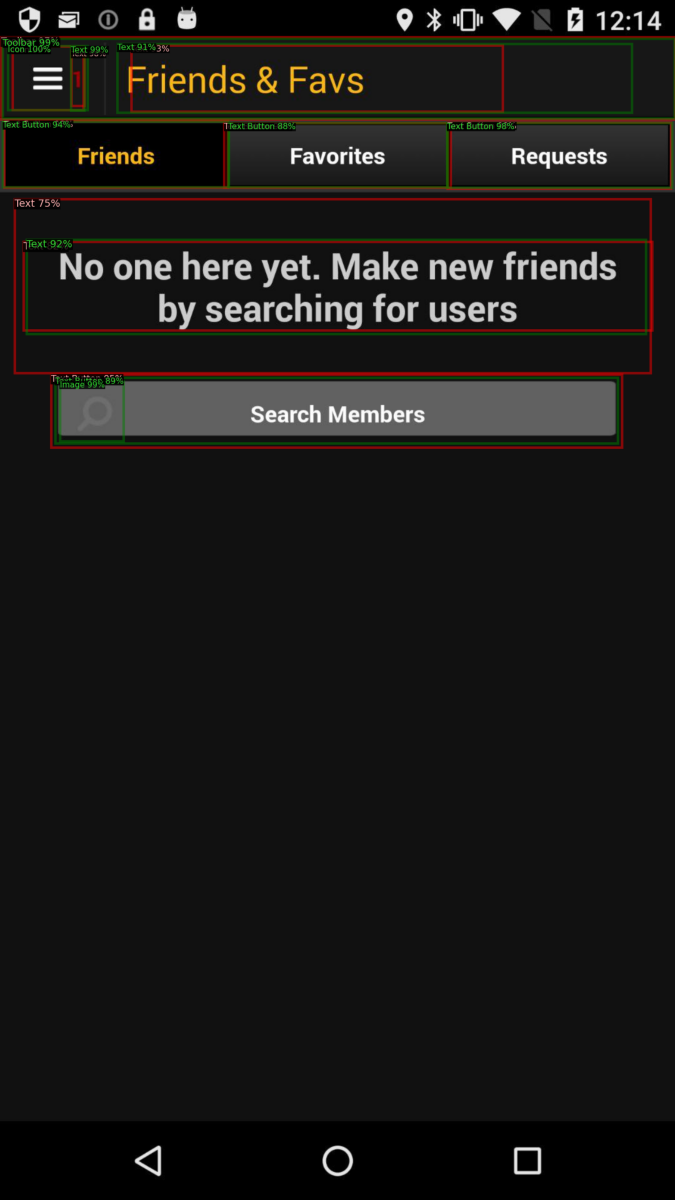} &
      \includegraphics[width=2.75cm]{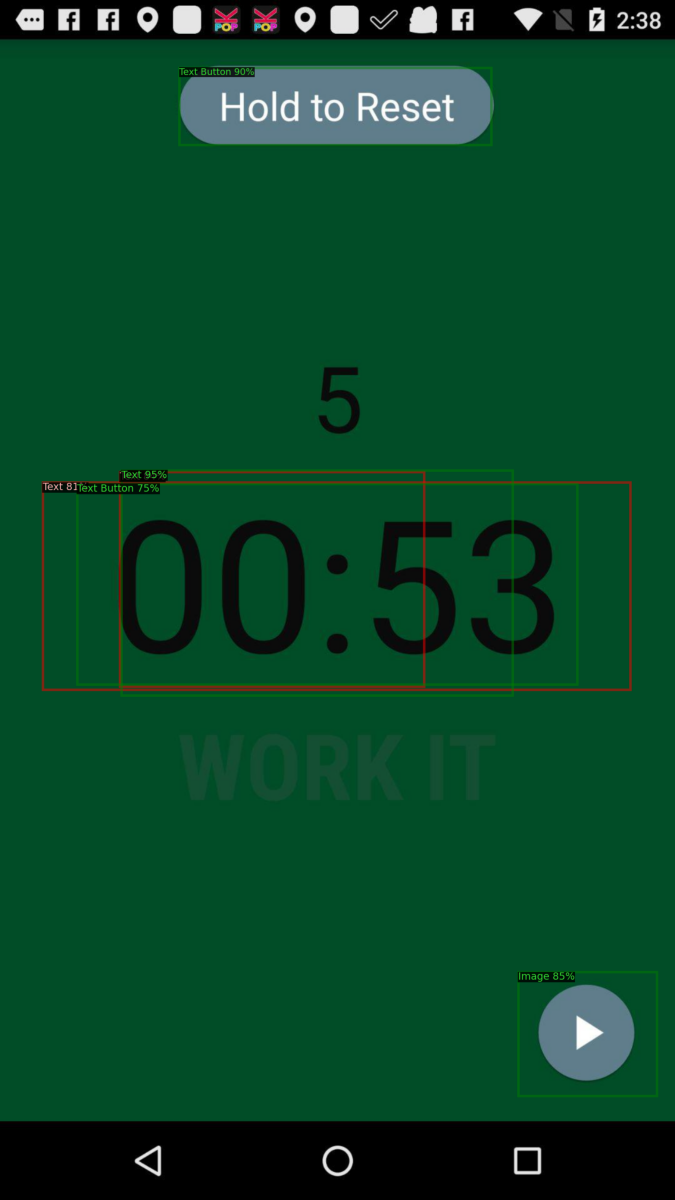} &
      \includegraphics[width=2.75cm]{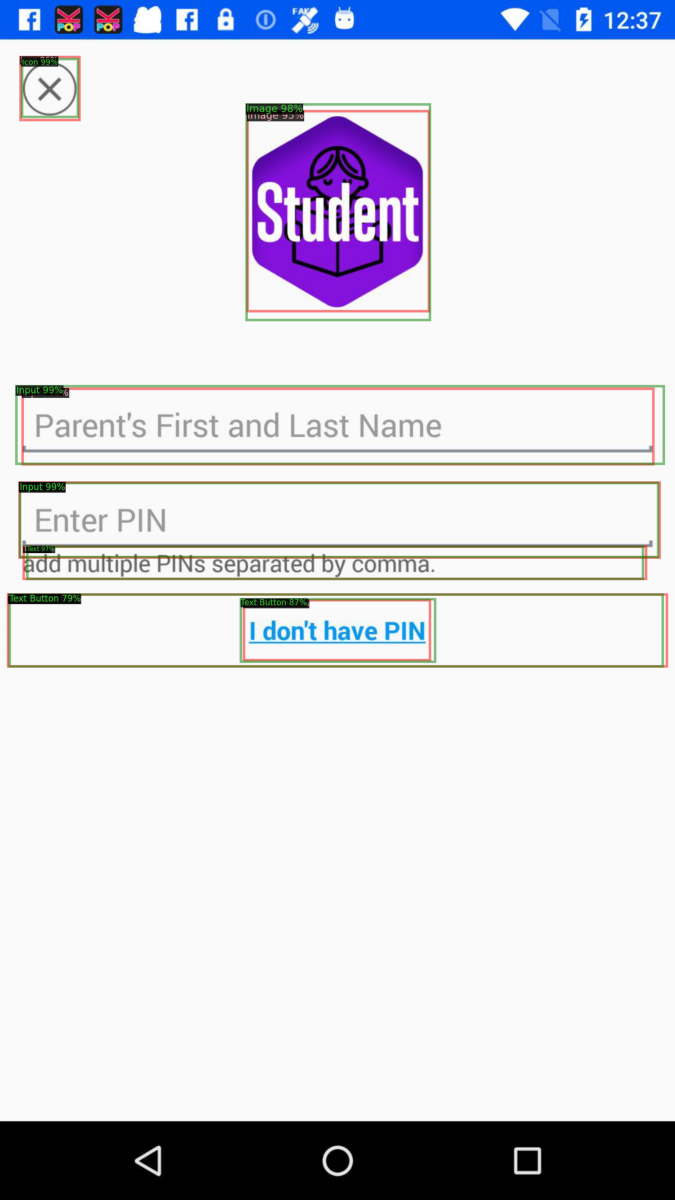} &  
      \includegraphics[width=2.75cm]{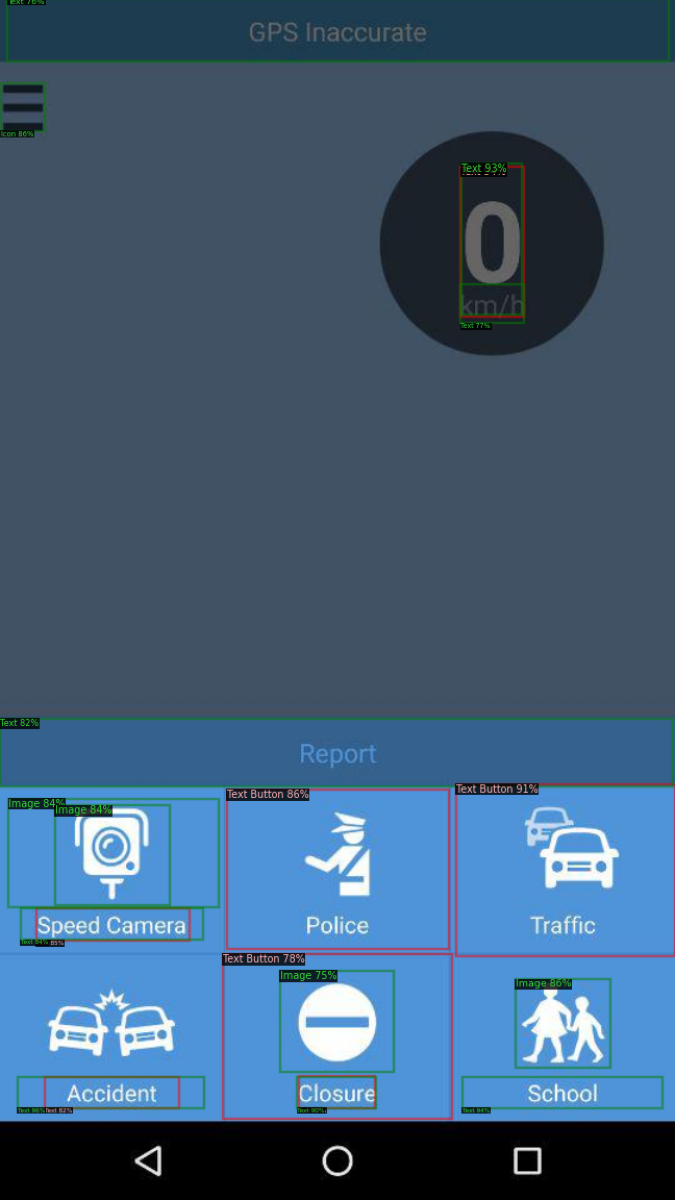} \\
      \multicolumn{5}{c}{\small{(b)}}
    \end{tabular}
    \caption{\small{ UX parsing on  RICO mobile screenshots \cite{deka2017Rico}. (a) Examples of UX parsing using MagicLayouts (Magic+FPN) (b) Comparison: Standard FPN (shown in red boxes) vs. MagicLayouts (green boxes). Our method is able to recognise components with higher confidence with lower false detections.  Zoom-in for better view.}}
    \label{fig:rico_det}
    \vspace{-3mm}
\end{figure*}

We enhance proposal features using priors from different layout regions taking their associations into account. We propagate the node representations $\mathbf{W} \in \mathbb{R}^{C\times (D+1)}$  via the graphs $\mathcal{G}_j$  using prior knowledge in edges $\boldsymbol{\mathcal{E}}_j$ given by $  \boldsymbol{\mathcal{E}}_j\mathbf{W}$. This allows to share  high-level semantics between related UI categories according to the graph knowledge. Next we map this semantic level information into individual proposal by a category-to-region mapping to condition them. A direct mapping for each region to a class can be obtained based on prediction of the previous classifier. However, such one-to-one mapping can be harsh as it is prone to  noise due to false predictions. In this paper, we instead propose to use a soft-mapping strategy which operates on probability distributions of proposals over all classes. Concretely we compute a mapping matrix $\mathbf{S} \in \mathbb{R}^{Nr\times C}$ given by $s_{ij} = \frac{\exp{p_{ij}}}{\sum_j \exp{p_{ij}}} $ where $s_{ij}$ is the probability that proposal $P_i$ belong to class $c_j$. Using these definitions, we obtain conditioned features as follows.
\begin{align}
\label{eq:enc_feat}
    f'_i = \sum_j^{Nb} \alpha(i,j) \odot \mathbf{S} \boldsymbol{\mathcal{E}}_j\mathbf{W} \mathbf{Z}_e ,
\end{align}
where $\odot$ is scalar element-wise multiplication, and  $\mathbf{Z}_e \in \mathbb{R} ^{(D+1) \times D'}$ is weight of the final embedding layer that $D'-$ dimension proposal features conditioned on the structural priors.  Note these proposal features are computed by aggregating all common spatial co-occurrence information from various regions from UX layout. We concatenate this conditioned representation with original feature $f = [f, f']$ and pass them through a final classifier head and a bounding-box regression head to obtain better detection results based on the conditioned proposal features.

\section{Experiments and Discussion}

We evaluate the performance of Magic Layouts and contrast  state-of-the-art baselines such as (Faster-RCNN \cite{faster-rcnn} and popular variants \cite{fpn2017}, and RetinaNet \cite{retinanet2017}) as well as the recent spatial-aware graph network (SGRN) \cite{xu2019spatial}.

\subsection{Datasets}
We evaluate on images from two input modalities: (i) mobile app screenshots and (ii) hand-sketched UI designs. 

\noindent \textbf{RICO dataset} \cite{deka2017Rico} is the largest publicly available dataset of UX designs containing 66K screenshots of mobile apps curated by crowd-sourcing and mining 9.3K free Android apps. The screenshots are annotated using bounding boxes to create semantic view hierarchies which are each assigned to one of $C=25$  classes of user interface (UI) component.  We partition the dataset into 53K training/validation samples $\mathcal{T}$ and a test set of 13K layouts for inference.

\noindent \textbf{DrawnUI dataset} \cite{drawnUI2020} contains 2,363 images of hand-drawn sketches released as development set of ImageCLEF 2020 drawnUI recognition task. The main motivation of this dataset is to enable designers to build UX layout by drawing them on whiteboard or on paper. The idea is to develop automatic UI parsing algorithm that can be further leveraged to convert them into UX codes. Each image is annotated for UI components with their bounding boxes and  class labels from a set of  $C=21$ predefined UI classes. We partition the dataset into a training set of 2000 images and perform evaluation on the remaining 263 images.

\subsection{Evaluation Metrics}
For both datasets, we report performance metrics used in COCO detection evaluation criterion \cite{lin2014microsoft} and provide mean Average Precision (AP) across various IoU thresholds \ie IoU = $\left\lbrace0.50:0.95, 0.5,  0.75\right\rbrace$ and various scales: $\lbrace \text{small, medium and large} \rbrace$. We also report Average Recall (AR) with different number of detection - $\left\lbrace1,10,100 \right\rbrace$ and scales: $\lbrace \text{small, medium and large} \rbrace$. Unless specified, we refer mAP@[0.50:0.95] to as mAP (primary metric) and AR@[0.50:0.95] as AR for conciseness.

\begin{figure*}[t!]
    \centering
    \includegraphics [width=0.75\textwidth]{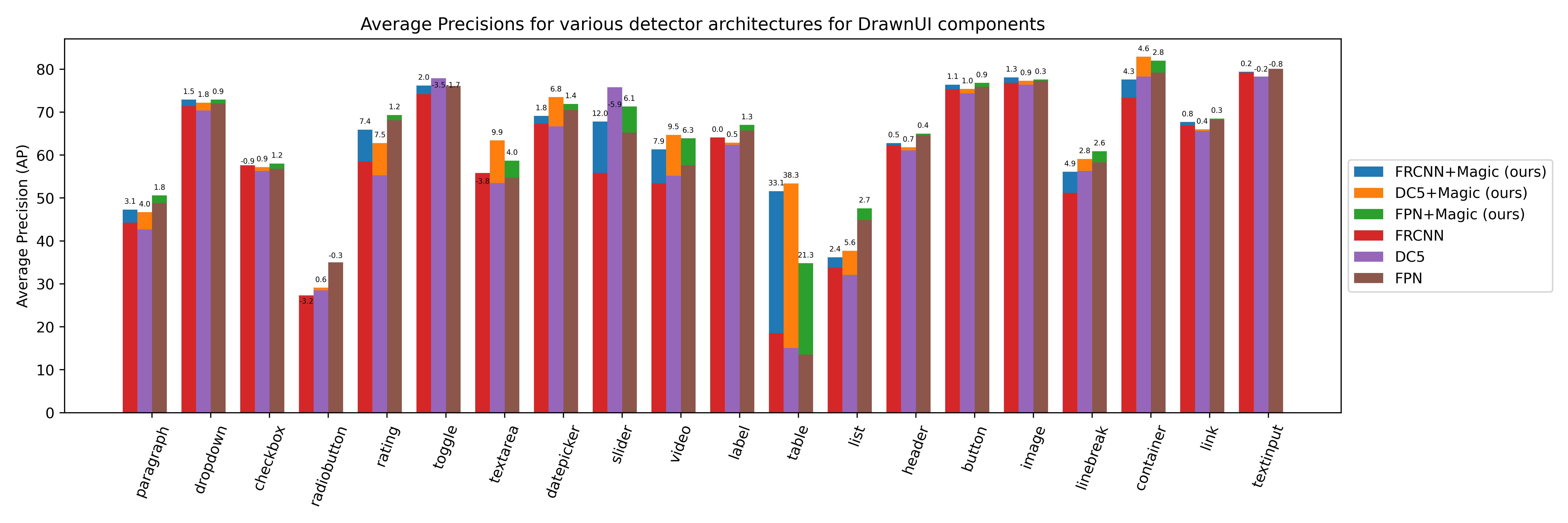}
    \caption{Average precision (AP) for DrawnUI categories for baselines and the proposed method using Faster-RCNN (FRCNN) \cite{faster-rcnn}, DC5 \cite{fpn2017} and FPN \cite{fpn2017}. See Fig.~\ref{fig:bars_rico} for details. Zoom-in for best view}
    \label{fig:bars_dui}
\end{figure*}

\begin{table*} [t!]
\caption{\textbf{Performance comparison on DrawnUI dataset.}}
\label{tab:comparison_dui_complete} 
\centering
\scriptsize{
        \begin{tabular}{|l||  c c c||  ccc|| ccc||   ccc|| }
        \hline
		Method  			& AP 	&AP50	&AP75 	&APs 	&APm 	&APl 	&AR 	&AR 	 &AR   &ARs  &ARm   &ARl\\ 
		@IoU				&0.5:95 	&0.5 	&0.75   &0.5:95 &0.5:95 &0.5:95 &0.5:95 &0.5:95 &0.5:95 &0.5:95 &0.5:95 &0.5:95\\
		maxDets				& 100	&100	&100	&100	&100	&100	&1		&10		&100  &100	&100	&100	\\	\hline \hline	 
		
		Faster-RCNN \cite{faster-rcnn}				&58.6	&86.3	&65.7	&21.7	&51.5	&60.8	&27.7	&61.7	&65.5   &35.2   &58.3   &67.0\\
		RetinaNet  \cite{retinanet2017}  	            	&58.6	&85.5	&66.1	&19.0	&52.9	&60.6	&29.9	&63.9	&68.1   &23.6   &60.1   &69.0\\ 

		FRCNN+SGRN \cite{xu2019spatial} 		&61.4	&87.9	&71.5	&27.0	&\textbf{56.9}	&63.5	&29.0	&64.7	&68.4 &34.7 &\textbf{63.6}   &69.6   \\ 

		\textbf{FRCNN+Magic}		&\textbf{62.2}	&\textbf{88.5}	&\textbf{72.8}	&\textbf{27.2}	&55.5	&\textbf{64.9}	&\textbf{30.4}	&\textbf{66.4}	&\textbf{70.1} &\textbf{33.6} &62.2 &\textbf{71.5}\\ \hline
		
		DC5 \cite{fpn2017}			&59.1	&85.4	&69.3	&23.6	&51.3	&61.2	&28.1	&62.6	&66.3 &29.9 &58.9 &67.1	 \\ 

		DC5+SGRN \cite{xu2019spatial}   &62.5	&\textbf{90.2}	&70.7	&26.4	&55.1	&65.0	&30.0	&65.8	&69.5  &\textbf{31.6} &61.4 &70.5\\ 
        \textbf{DC5+Magic}  &\textbf{63.4}	&89.9	&\textbf{72.9}	&\textbf{26.9}	&\textbf{56.9}	&\textbf{65.9}	&\textbf{30.7}	&\textbf{66.8}	&\textbf{70.6} &31.1 &\textbf{63.1} &\textbf{71.6}	\\\hline 
		
		FPN	\cite{fpn2017} 			&61.6	&87.3	&70.6	&32.1	&57.3	&63.5	&28.9	&64.5	&68.6 &36.9 &64.3   &69.3 \\	
		FPN+SGRN   \cite{xu2019spatial}          &63.3       &88.6   &73.7   &\textbf{34.6}   &\textbf{58.1}   &65.8   &30.1   &66.3   &70.5  &\textbf{39.1}    &\textbf{64.6}   &71.4 \\
		\textbf{FPN+Magic}   &\textbf{64.3}	&\textbf{89.5}	&\textbf{74.4}	&32.2	&54.4	&\textbf{66.5}	&\textbf{30.3}	&\textbf{66.7}	&\textbf{71.0} &37.1 &64.0   &\textbf{71.7}\\\hline	 
\end{tabular}
}
 \vspace{-5mm}
\end{table*}

\subsection{Experimental Settings}
\subsubsection{Architectures} 
We conduct experiments with widely adopted backbone network (ResNet \cite{resnet2016}) and best-performing detectors to demonstrate the effectiveness and generality of the proposed method. In particular, we build our  method using three popular variants of the Faster-RCNN architecture: \textbf{(i) Faster-RCNN} \cite{faster-rcnn},   \textbf{(ii) Dilated Convolutional Network (DC5) }\cite{fpn2017}, and  \textbf{ (iii)  Feature Pyramid Network (FPN) }\cite{fpn2017}.   
For Faster-RCNN \cite{faster-rcnn}, following the standard practise \cite{resnet2016}, we compute  region proposals on top of {\em conv4}, and all layers of {\em conv5} are adopted as predictor head with two sibling layers for classification and regression. The DC5 architecture uses dilated convolution in {\em conv5} layers and compute region proposals and perform RoI pooling over {\em conv5} features. As the prediction head, DC5 uses 2-\textit{fc} MLP followed by the two siblings layers which is lighter weight and faster than the {\em conv5} head \cite{fpn2017}. FPN \cite{fpn2017} has an alternate backbone where top-down and lateral connections are used to build a pyramid of features. Proposals are computed from all the pyramid scales and RoI pooling is performed on the most appropriate scale based on size of each proposal. FPN achieves the best speed and accuracy trade-off when compared to Faster-RCNN and DC5 \cite{fpn2017}. We use regional proposal network (RPN) to generate proposals and RoIAlign \cite{maskrcnn2017} is used for pooling the region features from feature maps. We show that all three architectures benefit from  our spatial prior for object co-occurrence (subsec. \ref{ssec:eval}).

\subsubsection{Implementation Details} 
We implement our framework using Pytorch \cite{pytorch} with detectron2 \cite{detectron2} codebase. We use ResNet50 \cite{resnet2016} pretrained on ImageNet as our backbone network. Images are resized such that shorter side has maximum of 800 pixels and larger side has 1333 pixels. For all settings, we sample $Nr = 256$ proposals from each image after non-maximal suppression (NMS) which are assigned as positive if the proposal and a ground-truth box has $\text{IoU}>0.7$ or as negative if the $\text{IoU}<0.3$. We follow other standard settings as in \cite{fpn2017}.

We use SGD  optimizer with a momentum update of 0.9 and a weight decay of 0.0001; and set the initial learning of 0.02 and decay it by a factor of 0.1 twice during training. We use 3 images per GPU and a mini-batch of 9 for training. We train all three network architectures for 21 epochs and 45 epochs for RICO and DrawnUI dataset respectively. We use more epochs for DrawnUI as it has fewer UXs compared to RICO. We observe that the performances saturates on validation data after 16 and 31 epoch for RICO and DrawnUI respectively; further training does not improve the performance. To obtain the conditioned proposal features in our framework, we initialise our model using pre-trained networks from their respective architectures.

\begin{figure*}[t!]
    \centering
    \begin{tabular}{ccccc}
      \includegraphics[width=2.75cm]{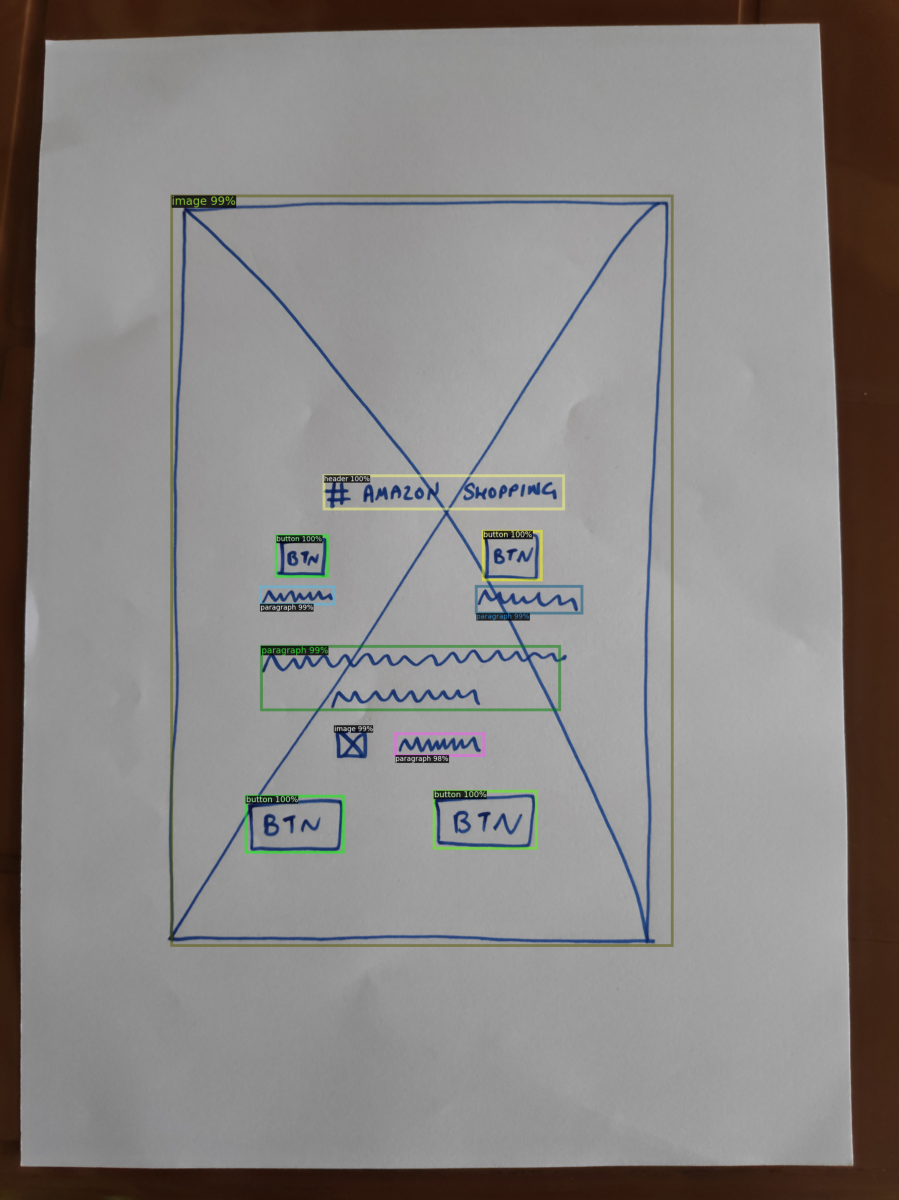} &  \includegraphics[width=2.75cm]{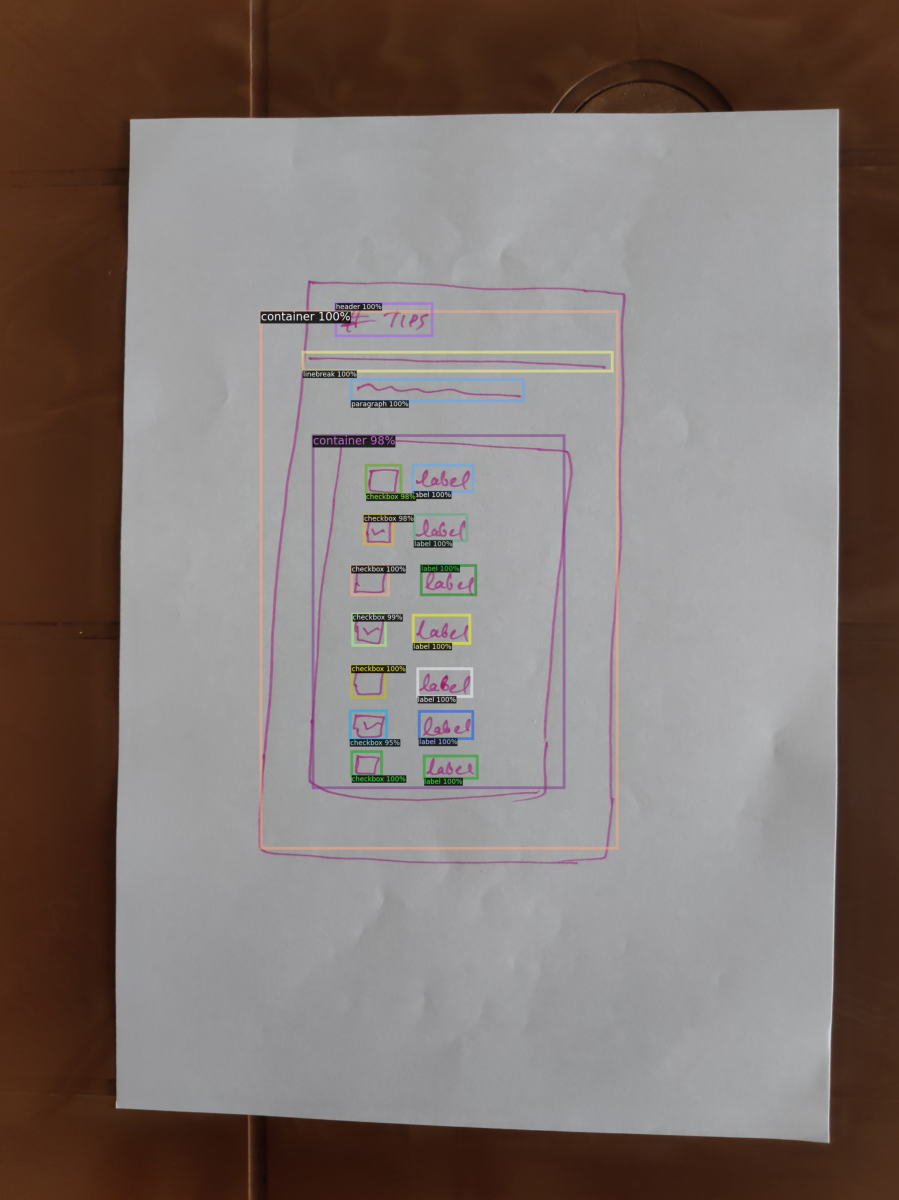} &
      \includegraphics[width=2.75cm]{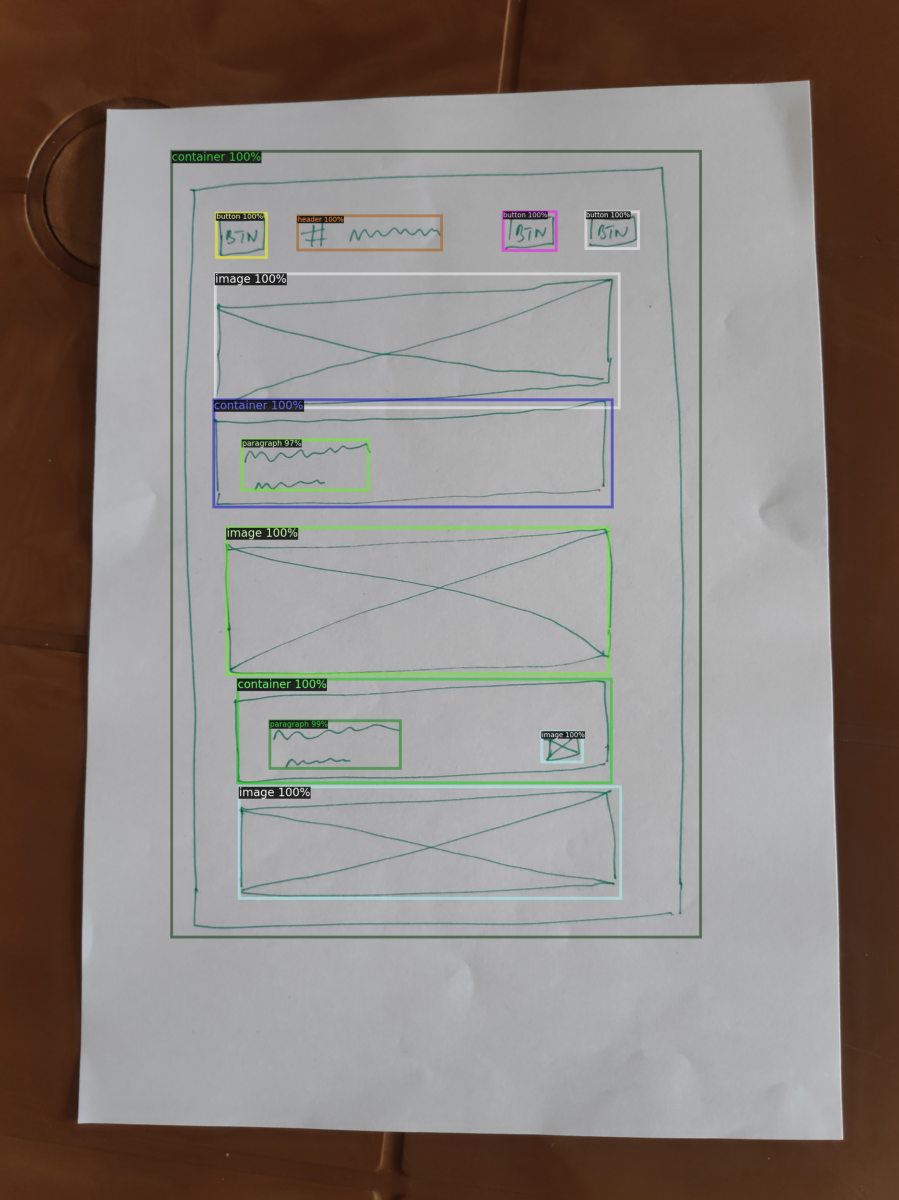} &       \includegraphics[width=2.75cm]{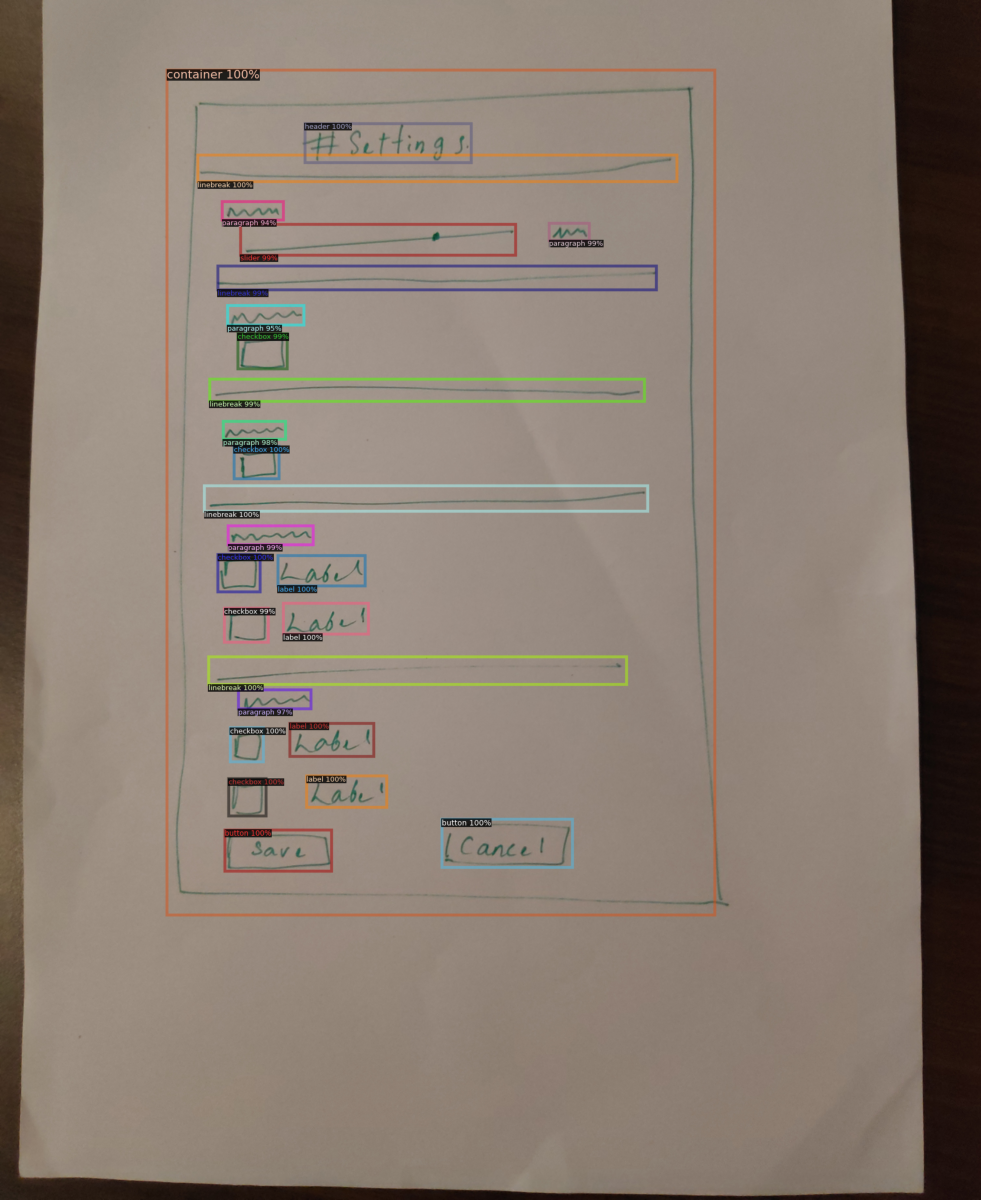} &
      \includegraphics[width=2.75cm]{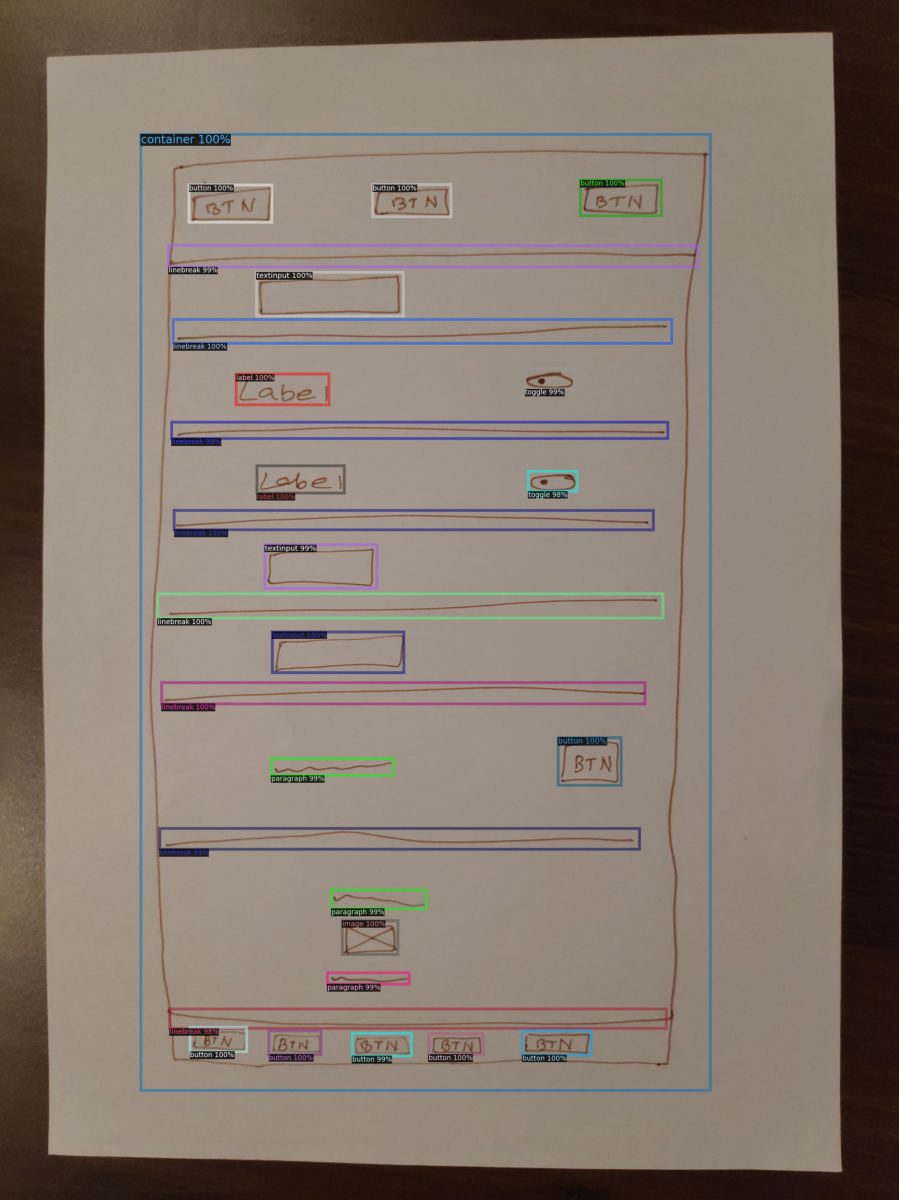} \\
       \multicolumn{5}{c}{\small{(a) }} \\  
      \includegraphics[width=2.75cm]{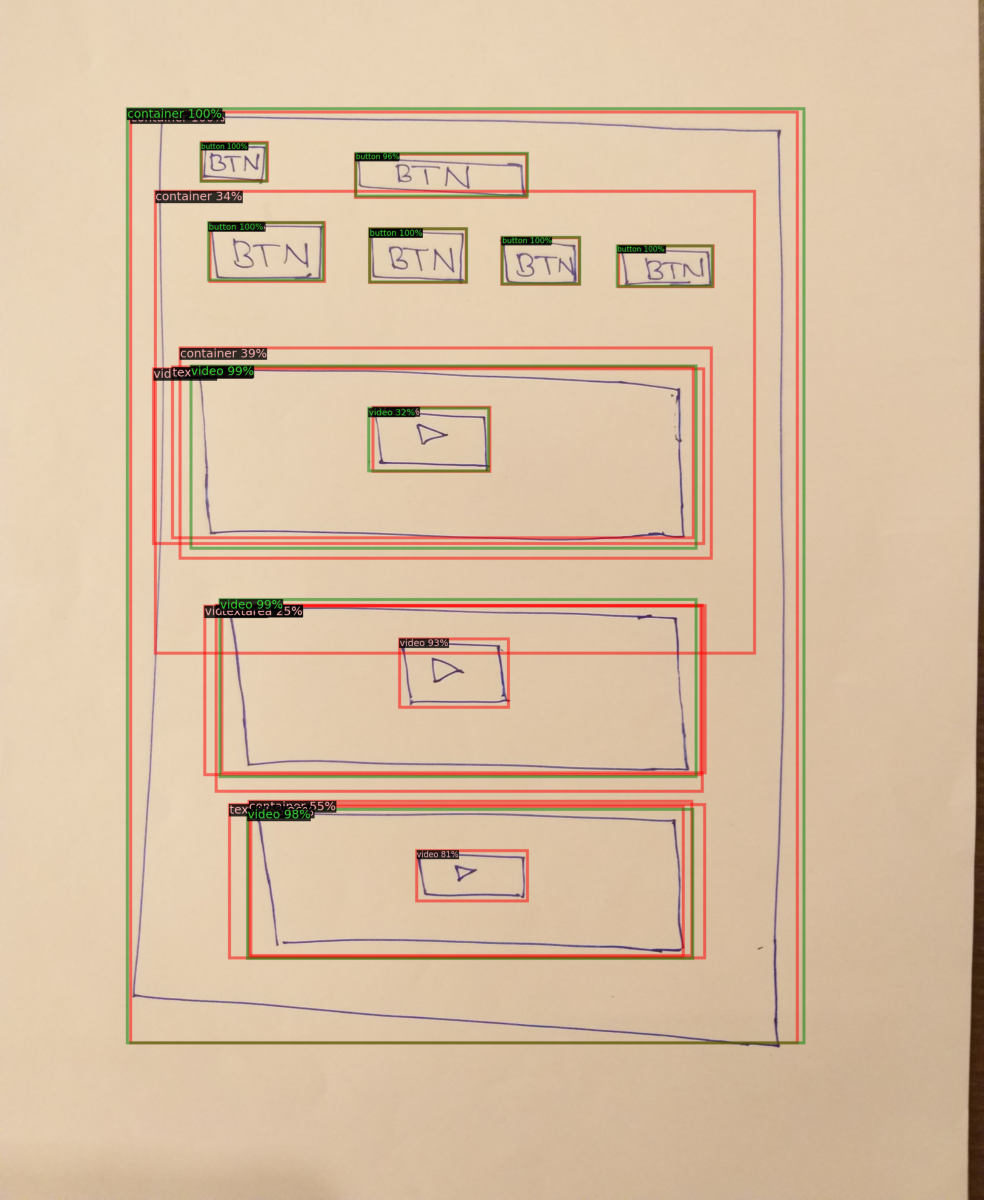} &  \includegraphics[width=2.75cm]{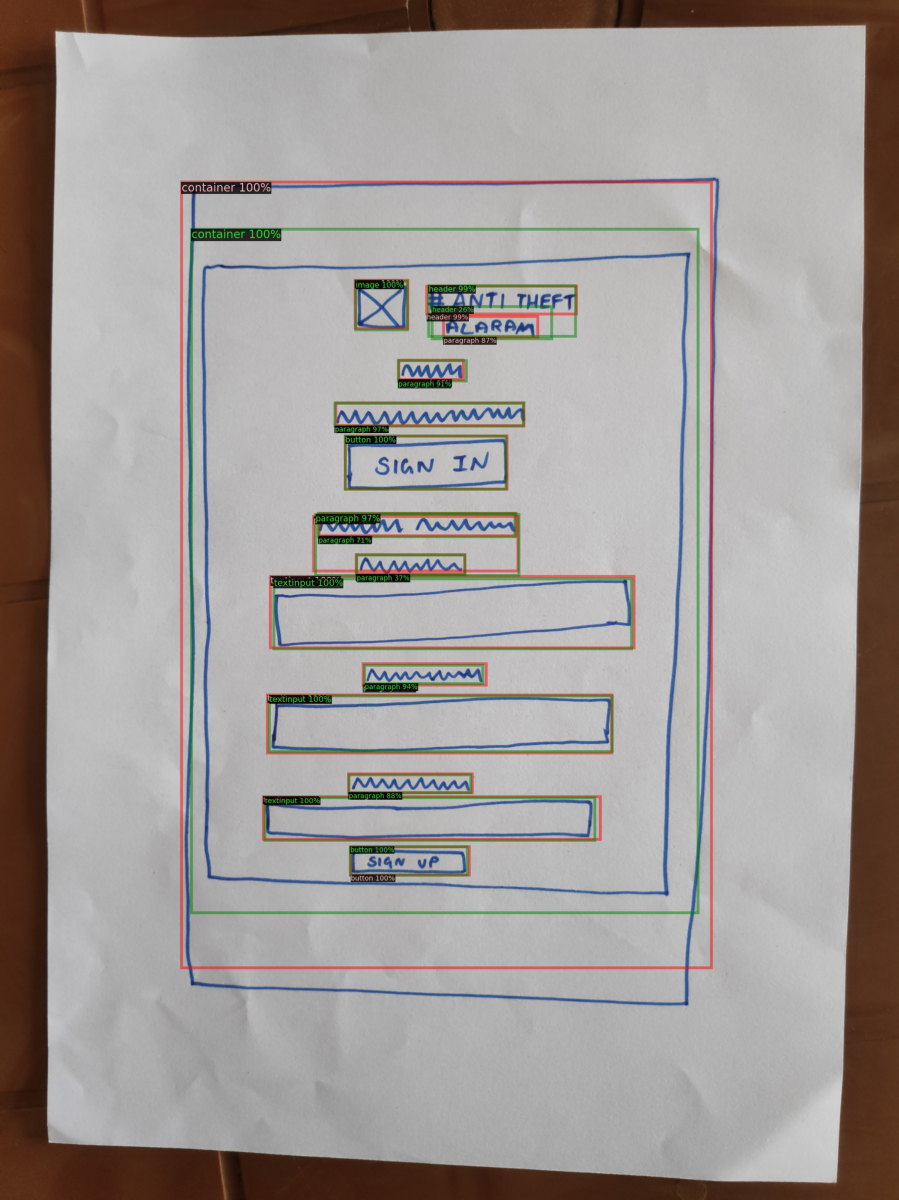} &
      \includegraphics[width=2.75cm]{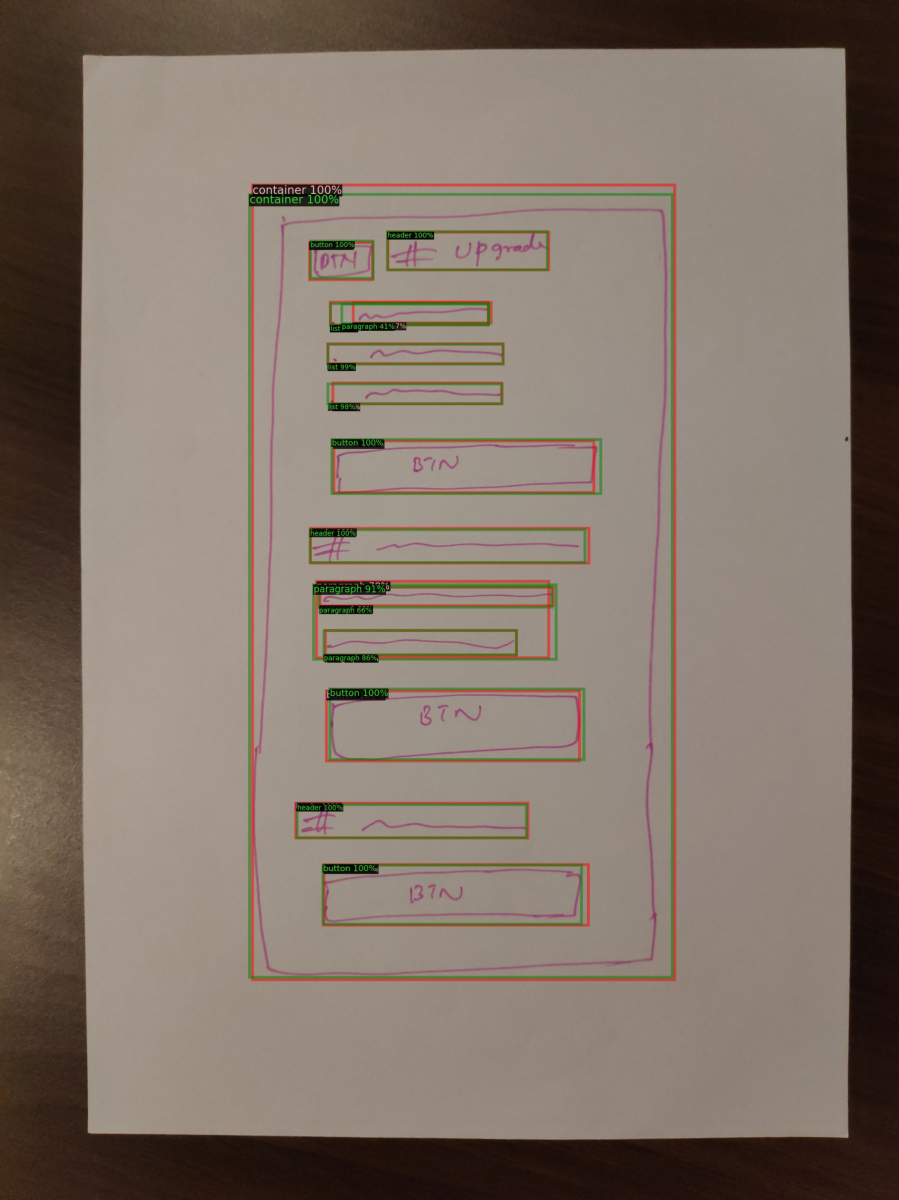} &
      \includegraphics[width=2.75cm]{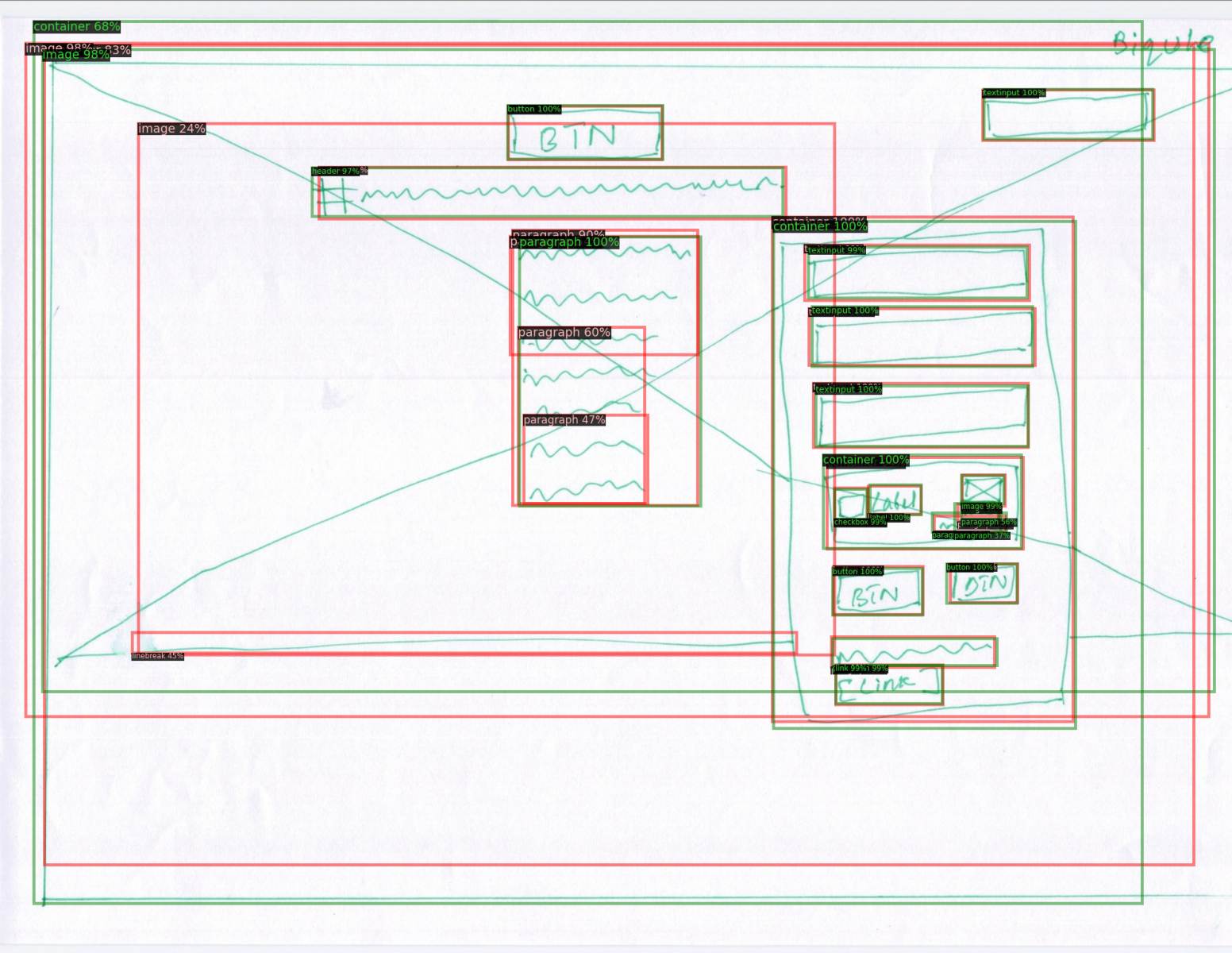} &       \includegraphics[width=2.75cm]{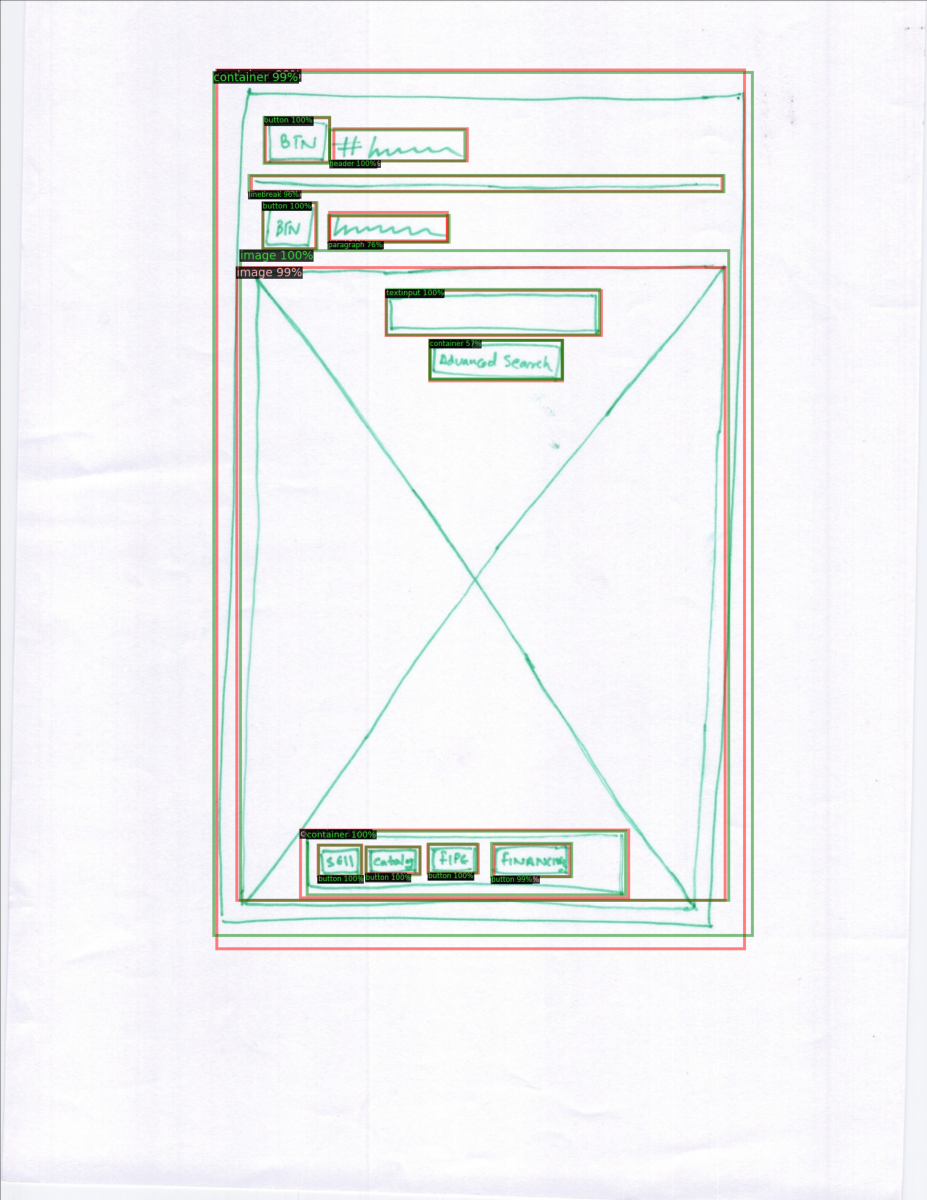} \\
      \multicolumn{5}{c}{\small{(b)}}
    \end{tabular}   
    \caption{UX parsing on hand-sketched DrawnUI layouts \cite{drawnUI2020}. (a) Examples of UX parsing with MagicLayouts (Magic+FPN). (b) Comparison: Standard FPN (shown in red boxes) vs. MagicLayouts (green boxes). See caption of Fig~\ref{fig:rico_det}}
    \label{fig:dui_det}
    \vspace{-3mm}
\end{figure*}
We conducted experiments using both non-overlapping as well as overlapping bands and observed that structural priors estimated from both achieve similar performances. Hence we opt for simplicity and divide UXs into $N_b (=N_g) = 10$ non-overlapping bands for all experiment unless stated. We also study impact of $Ng$ on performance (subsec. \ref{sssec:ablations}). We used a zero-mean Gaussian distribution with a standard deviation empirically set to $\sigma =0.3$ in order to compute the associations of proposals to the bands. The dimension of conditioned feature $D'$ is set to 512.

\subsection{Results}
\label{ssec:eval}
\subsubsection{Parsing UX screenshots on RICO}
Fig.~\ref{fig:bars_rico} shows average precision (AP) using the proposed method and compares  with baselines with various network architectures illustrating the improvements brought by Magic Layouts over various RICO UI categories. From the figure it can be observed that for all the architectures, our proposed method is able to boost the performance (numbers on top of each stacked bar, Fig.~\ref{fig:bars_rico})  of nearly all UI categories. For example, our method is able to deliver improvements of +7.7\%, +8.0\% and +4.5\% AP for \emph{Page Indicator} class over FasterRCNN, DC5 and FPN method respectively. This clearly demonstrates the advantage of incorporating structural priors of layout in UI recognition.

Table~\ref{tab:comparison_rico_complete} summarizes recognition performance in terms of  mean Average Precision (mAP) at different IoU thresholds and scales and Average Recall (AR) at different number of detections and scales. Faster-RCNN \cite{faster-rcnn} achieves an mAP of 43.0\% whilst our proposed framework achieves an mAP of 47.8\% - an absolute increase of +4.8\% indicating the effectiveness of the proposed method and the use of  structural prior. Our method also outperforms competing methods such as RetinaNet \cite{retinanet2017} and SGRN-FRCNN \cite{xu2019spatial} by mAPs of  6.1\% and 2.6\% respectively. The recent SGRN method aims to enhance proposal representation, however, it assumes homogeneous object relations across various regions of image and creates edges based on visual similarity of proposals which can potentially limit the exploration of inter-class occurrences. Compared to this, our method aggregates structural priors from various regions explicitly considering the variability in UI distributions and conditions the proposals at a semantic level, improving accuracy.

Magic Layouts with DC5 architecture achieves an mAP of 51.8\%  outperforming its counterpart by a margin of 5.1\% and SGRN-DC5 by 2.7\%. Similarly, Magic-Layout with FPN architecture achieves 50.3\% AP /ie +2.7\% over the standard FPN \cite{fpn2017} and +0.4\% over SGRN \cite{xu2019spatial}. From Table~\ref{tab:comparison_rico_complete}, we can observe similar improvements for Average Recall and related metrics; for example MagiLayouts with Faster-RCNN achieves 63.9\% AR@100 outperforming its counterpart by +3.8\% and SGRN by +1.8\%. In a nutshell, the proposed method offers benefits for various detector architectures and outperforms existing approaches that also incorporate relations among components. 

Fig.~\ref{fig:rico_det} shows sample qualitative UX parsed on RICO using our Magic-Layout (with FPN arch). In Fig.~\ref{fig:rico_det}(a) we observe that Magic-Layout is able to detect and recognize various UI components at different scales; in (b) we compare MagicLayouts with baseline and show our method is more effective when compared to baselines.   

\subsubsection{Parsing hand-sketched UXs}
We present evaluations on the DrawnUI dataset \cite{drawnUI2020} and show that Magic-Layouts can effectively detect components on hand-drawn wire-frames while outperforming all baselines.  Fig.~\ref{fig:bars_dui} shows improvements in average precision (AP) obtained by the proposed method for DrawnUI components using the three architectures. Our method provides consistent improvements for almost all component classes providing an average boost of ~3.5\% AP over the three architectures which clearly shows the advantage of incorporating co-occurrence prior into the framework. We also observed that categories which are comparatively rarer are largely benefited by the structural prior; \eg table component  has less than 50 instances in DrawnUI dataset; and hence detectors may perform poorly for such rare classes. Faster-RCNN, DC5 and FPN achieve APs of 18.5\%, 15.1\% and 13.5\% respectively for this class. Magic-Layouts boost these performance to 51.6\%, 53.4\% and 34.8\%  respectively for the three detectors demonstrating its effectiveness.

Table \ref{tab:comparison_dui_complete} presents mAPs and Average Recall for DrawnUI dataset obtained using the proposed method and compares with different architectures and existing methods. Magic-Layout with Faster-RCNN network achieves an mAP of 62.2\% which outperforms its counterpart \cite{faster-rcnn} by 3.6\%. It also outperforms RetinaNet and spatially-aware SGRN by 3.6\% and 0.8\% respectively. MagicLayouts with FPN achieves the best mAP of 64.3\% on this dataset.

In terms of recall, our method achieves the best AR@100 of 63.9\%, 66.7\% and 65\% for FRCNN, DC5 and FPN respectively. Overall, the proposed Magic Layouts provides improvements over all baseline Faster-RCNN variants \cite{faster-rcnn, fpn2017} and also outperforms other baseline methods using spatial relations \cite{xu2019spatial,retinanet2017}. Fig.~\ref{fig:dui_det} shows sample example of parsed UXs from DrawnUI dataset.Our method is able to better detect and recognize UIs  at different scales despite potential variations in hand-sketches and illuminations. 
\begin{figure}
    \centering
    \setlength{\tabcolsep}{0pt}
    \begin{tabular}{c c}
       \includegraphics[width= 0.20\textwidth]{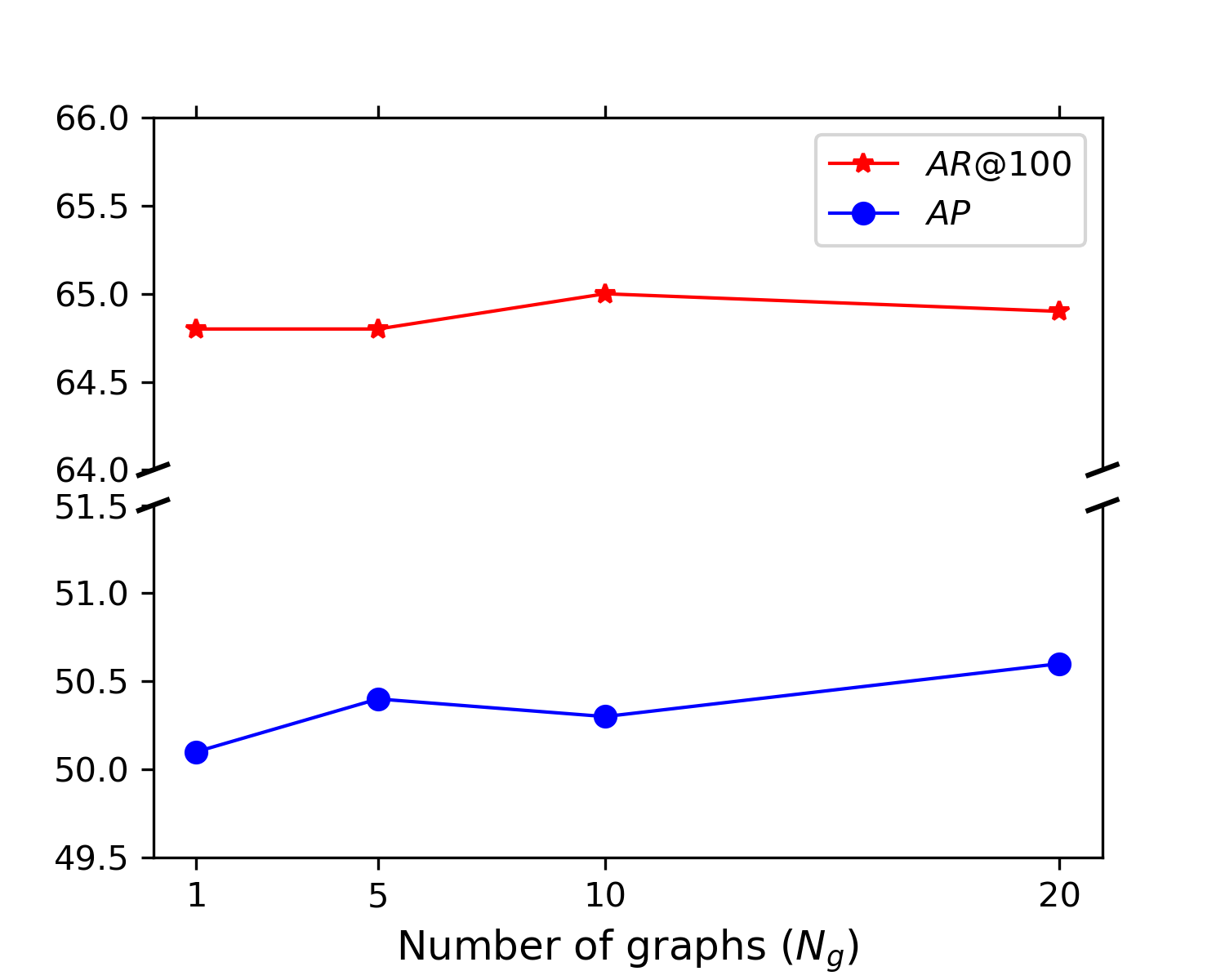}  &   \includegraphics[width= 0.20\textwidth]{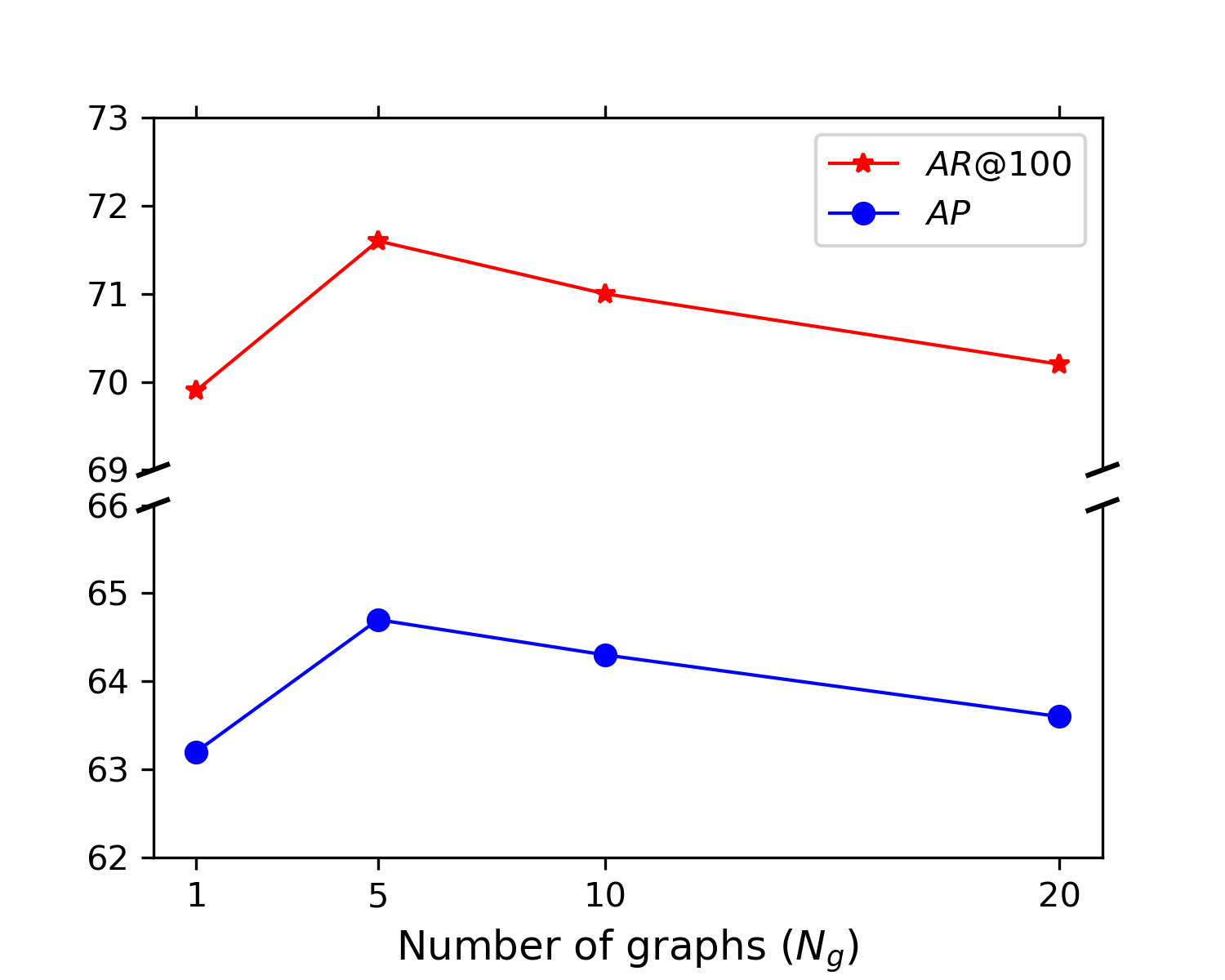}\\
         \small{(a) RICO} & \small{(b) DrawnUI}
    \end{tabular}
    \caption{Performance of FPN+Magic on  (a) RICO \cite{deka2017Rico} and (b) DrawnUI \cite{drawnUI2020} at various values of Ng = $\lbrace1,5,10,20\rbrace$.}
    \label{fig:ng_curve}
\end{figure}

\begin{figure}
    \centering
    \includegraphics [width=0.8\linewidth,height=5.5cm]{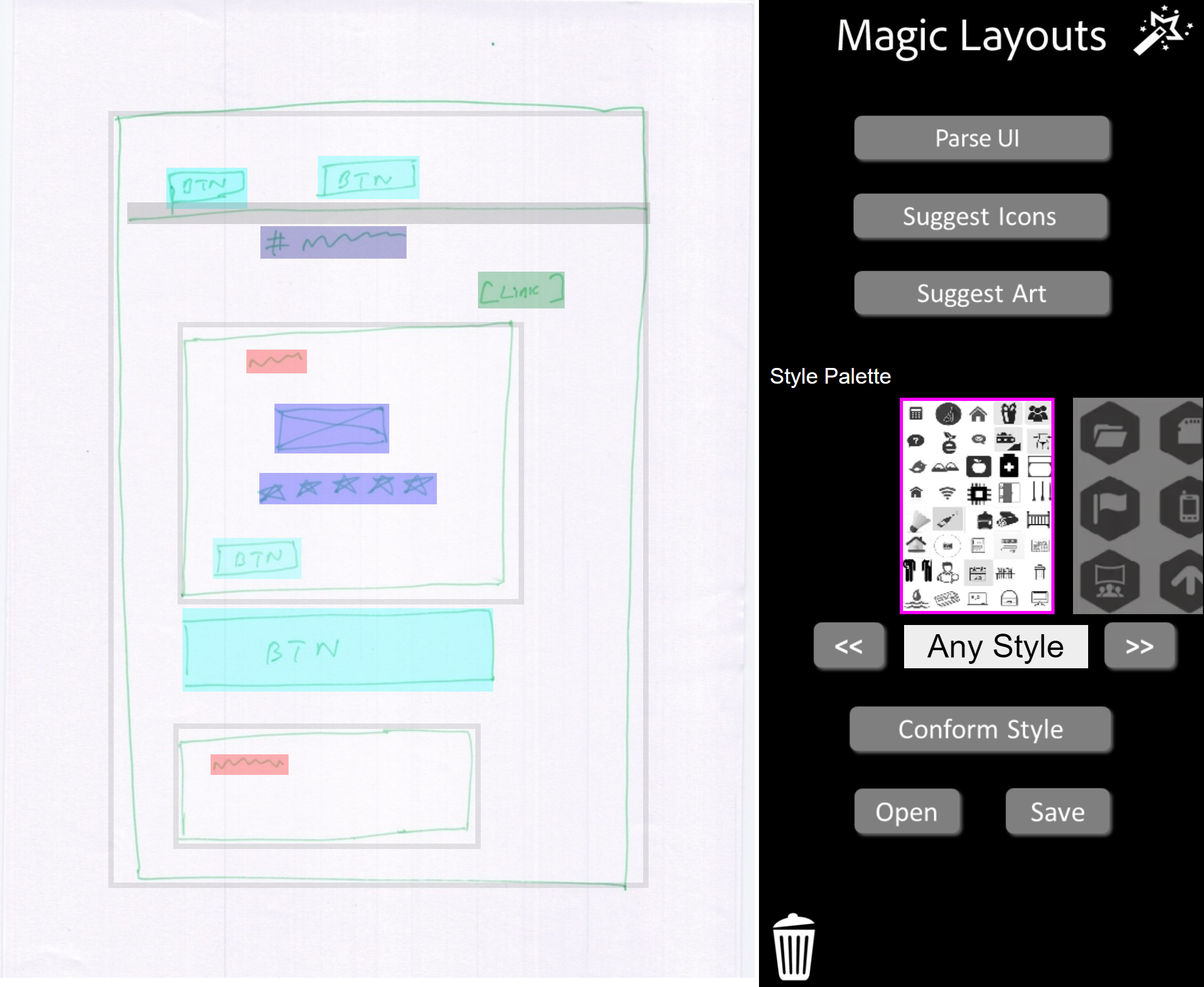}
    \caption{The Magic Layouts web app parsing a UI wireframe sketch from the DrawnUI dataset \cite{drawnUI2020}; coloured regions indicate different classes of recognised UI component. UX sketches are rapidly converted into higher digital prototypes.  The tool also incorporates a sketch based visual search to replace sketched artwork with higher fidelity graphics.}
    \label{fig:mlayout}
\end{figure}

\subsubsection{Ablation Studies}
\label{sssec:ablations}
\noindent\textbf{Impact of $N_g$:} We conduct experiments using different number of graphs ($N_g$) during co-occurrence computations. Fig~\ref{fig:ng_curve} shows the performance of the proposed method in term of AR@100 and AP[0.5:0.95] for different values of $N_g$. Our method achieves  similar performances for different $N_g$ values with deviations about 0.5\% and 1\% for RICO and DrawnUI respectively indicating that our method is fairly insensitive to the parameter. $N_g=10$ provides good trade-off between AP and AR for both datasets.

\noindent\textbf{Design choice study:} We conduct experiments with various strategies for graph-proposal association and category-to-region mapping. In particular, we run experiment with following setups: A. Baseline, B. graph-proposal association: Single-Assignment vs. Equal vs. Gaussian Weighting, C. Category-to-region mapping: Soft vs. Hard/1-to-1,  D. Graph node: classifier's weight vs. proposals features). Tables \ref{tab:ablation_dui} and \ref{tab:ablations_rico} summarise the performances for DrawnUI and RICO respectively; both datasets follow the same trend. For graph-proposal association, multiple assignment using Gaussian weighting performs the best outperforming single graph as well equal weighting scheme. Assigning proposals appropriately to their corresponding bands also provides on-par performance for some metrics. For category-to-region mapping, the proposed soft-mapping outperforms one-to-one hard mapping for both single and Gaussian-weighted assignments. We further conduct an extra study, substituting $\mathbf{W}$ with representations computed from proposal features ($\mathbf{P} = \mathbf{FS} \in \mathbb{R}^{C\times D} $); we achieve on-par performances  (Table \ref{tab:ablation_dui}, \ref{tab:ablations_rico}-D \& E) indicating that our method works well with alternative choices of node features.

\subsection{Practical Use Case: Magic Layouts}

We deployed our proposed detection model into an interactive `Magic Layouts'  web app capable of parsing UI layouts from mobile camera photographs of sketches (Fig.~\ref{fig:mlayout}), or screenshots from mobile app stores. Magic Layouts incorporates sketch based image search (\cite{Collomosse2017}) to replace sketched graphics and icons with higher fidelity artwork. Please see supplementary video demo. 
\begin{table} [!h]
\caption{\textbf{Ablation studies DrawnUI}}
\label{tab:ablation_dui} 
\centering
\scriptsize{
        \begin{tabular}{|l|l| c| c| c|c| cc|   cc| }
        \hline
        
		\multicolumn{2}{|l|}{\textbf{Config/}} & $Nb$ & Assig- & Map- & Node  & \textbf{AP} 	&\textbf{AP50} 	&\textbf{AR} 	&\textbf{AR} \\
		\multicolumn{2}{|l|}{\textbf{Method}}  &    & nment  &  ping & Feat &  	& 	&\textbf{@1} 	&\textbf{@10} \\ \hline \hline
        \textbf{A}      & FPN         & -     & -             & -         &-              &61.6	  &87.3      &28.9   &64.5   \\ \hline \hline
        \textbf{B}     & Magic       & 1     & Single        & Soft      & $\mathbf{W}$  &63.2   &88.8       &29.0   &65.6     \\ 
                & Magic       & 10    & Equal         & Soft      & $\mathbf{W}$  &62.7   &87.6      &28.2   &66.0    \\
                & Magic       & 10    & Single        & Soft      & $\mathbf{W}$  &64.2   &89.4      &30.5   &\textbf{67.2}    \\ \hline\hline
        \textbf{C}      & Magic       & 10    & Single        & Hard      & $\mathbf{W}$  &63.5   &89.8      &29.4   &66.6    \\
                & Magic       & 10    & Gauss      & Hard      & $\mathbf{W}$  &63.9   &88.7      &30.0   &66.6   \\ \hline \hline
        \textbf{D}      & Magic       & 10    & Gauss      & Soft      & $\mathbf{P}$  &63.6   &\textbf{90.4}      &\textbf{30.5}   &66.9   \\ \hline  \hline  
        \textbf{E}      & Magic       & 10    & Gauss      & Soft      & $\mathbf{W}$  &\textbf{64.3}   &89.5      &30.3	 &66.7  \\ \hline
       \end{tabular}
}        
\end{table}
\begin{table} [!h]
\caption{\textbf{Ablation studies on RICO}}
\label{tab:ablations_rico} 
\centering
\scriptsize{
        \begin{tabular}{|l|l| c| c| c|c| cc|   cc| }
        \hline
		\multicolumn{2}{|l|}{\textbf{Config/}} & $Nb$ & \textbf{Assig-} & \textbf{Map-} & \textbf{Node}  & \textbf{AP} 	&\textbf{AP50} 	&\textbf{AR} 	&\textbf{AR} \\
		\multicolumn{2}{|l|}{\textbf{Method}}  &    & \textbf{nment}  &  \textbf{ping} & \textbf{Feat} &  	& 	&\textbf{@1} 	&\textbf{@10} \\ \hline \hline
        \textbf{A}      & FPN         & -     & -             & -         &-             	&47.6   &57.1	 		&41.6	&61.0	     \\ \hline \hline
        \textbf{B}     & Magic       & 1     & Single        & Soft     & $\mathbf{W}$       & 50.1 &\textbf{60.1}    &42.8 &62.8  \\ 
                & Magic       & 10    & Equal         & Soft      & $\mathbf{W}$     & 50.1 &59.8  &42.6 &62.5  \\
                & Magic       & 10    & Single        & Soft      & $\mathbf{W}$     &50.2 &60.0 &42.7 &62.5   \\ \hline\hline
        \textbf{C}      & Magic       & 10    & Single        & Hard      & $\mathbf{W}$    &50.3 &\textbf{60.1} &\textbf{43.0} &\textbf{63.0} \\
                & Magic       & 10    & Gauss      & Hard      & $\mathbf{W}$    & 49.9  &59.9    &42.8  &62.6     \\ \hline \hline
        \textbf{D}      & Magic       & 10    & Gauss      & Soft      & $\mathbf{P}$    &\textbf{50.4} &\textbf{60.1}  &42.9 &62.9 \\ \hline  \hline  
        \textbf{E}      & Magic       & 10    & Gauss      & Soft      & $\mathbf{W}$    &50.3	&\textbf{60.1}	&\textbf{43.0}	&\textbf{63.0}	   \\ \hline
       
       \end{tabular}
}        
\end{table}

\section{Conclusion}

We reported Magic Layouts; a technique for incorporating structural layout (common spatial object co-occurrences) as a prior to guide object detection and localization, investigating this in the context of UI layout parsing.  We extended the Faster-RCNN backbone to incorporate a learned prior based on spatial distribution of UI components, showing performance improvements over several Faster-RCNN variants \cite{faster-rcnn,fastrcnn}, RetinaNet \cite{retinanet2017} and including an existing graph based approach to learning spatial prior for detection \cite{xu2019spatial}.   We demonstrated the utility of our model within  `Magic Layouts' -- a UX parsing tool capable of automatically parsing UI layouts  in two domains: app screenshots and free-hand sketched prototypes.  Future work could explore not only object co-proximities but also hierarchical relationships which particularly for UI layouts could offer further structural cues.

{\small
\bibliographystyle{ieee_fullname}
\bibliography{uibib}

\begin{thebibliography}{10}\itemsep=-1pt

\bibitem{Alexe2012}
B. Alexe, T. Deselaers, and V. Ferrari.
\newblock Measuring the objectness of image windows.
\newblock {\em IEEE Trans. Pattern Analysis and Machine Intelligence TPAMI)},
  2012.

\bibitem{Arbelaez2014}
P. Arbelaez, J. Pont-Tuset, J.~T. Barron, F. Marques, and J. Malik.
\newblock Multiscale combinatorial grouping.
\newblock In {\em Proc. CVPR}, 2014.

\bibitem{pix2code}
T. Beltramelli.
\newblock pix2code: Generating code from a graphical user interface screenshot.
\newblock {\em arXiV 1705.07962v2}, 2017.

\bibitem{chen2018iterative}
Xinlei Chen, Li-Jia Li, Li Fei-Fei, and Abhinav Gupta.
\newblock Iterative visual reasoning beyond convolutions.
\newblock In {\em Proceedings of the IEEE Conference on Computer Vision and
  Pattern Recognition}, pages 7239--7248, 2018.

\bibitem{Collomosse2017}
J. Collomosse, T. Bui, M. Wilber, C. Fang, and H. Jin.
\newblock Sketching with style: Visual search with sketches and aesthetic
  context.
\newblock In {\em Proc. ICCV}, 2017.

\bibitem{dai2017detecting}
Bo Dai, Yuqi Zhang, and Dahua Lin.
\newblock Detecting visual relationships with deep relational networks.
\newblock In {\em Proceedings of the IEEE conference on computer vision and
  Pattern recognition}, pages 3076--3086, 2017.

\bibitem{deka2017Rico}
Biplab Deka, Zifeng Huang, Chad Franzen, Joshua Hibschman, Daniel Afergan, Yang
  Li, Jeffrey Nichols, and Ranjitha Kumar.
\newblock Rico: A mobile app dataset for building data-driven design
  applications.
\newblock In {\em Proceedings of the 30th Annual Symposium on User Interface
  Software and Technology}, UIST '17, 2017.

\bibitem{drawnUI2020}
Dimitri Fichou, Raul Berari, Paul Brie, Mihai Dogariu, Liviu~Daniel \c{S}tefan,
  Mihai~Gabriel Constantin, and Bogdan Ionescu.
\newblock Overview of {ImageCLEFdrawnUI} 2020: The detection and recognition of
  hand drawn website uis task.
\newblock In {\em CLEF2020 Working Notes}, {CEUR} Workshop Proceedings,
  Thessaloniki, Greece, September 22-25 2020. CEUR-WS.org
  $<$http://ceur-ws.org$>$.

\bibitem{Geigel2001}
J. Geigel and A. Loui.
\newblock Automatic page layout using genetic algorithms for electronic
  albuming.
\newblock In {\em Proc. Electronic Imaging}, 2001.

\bibitem{gidaris2018dynamic}
Spyros Gidaris and Nikos Komodakis.
\newblock Dynamic few-shot visual learning without forgetting.
\newblock In {\em Proceedings of the IEEE Conference on Computer Vision and
  Pattern Recognition}, pages 4367--4375, 2018.

\bibitem{fastrcnn}
R. Girshick.
\newblock Fast r-cnn.
\newblock In {\em Proceedings of the IEEE international conference on computer
  vision}, 2015.

\bibitem{rcnn}
R. Girshick, J. Donahue, T. Darrell, and J. Malik.
\newblock Rich feature hierarchies for accurate object detection and semantic
  segmentation.
\newblock In {\em Proc. CVPR}, 2014.

\bibitem{Goldenbert2003}
E. Goldenbert.
\newblock {Automatic layout of variable-content print data}.
\newblock Master's thesis, School of Cognitive \& Computing Sciences,
  University of Sussex, UK, 2000.

\bibitem{Harrington2004}
S. Harrington, J. Naveda, R. Jones, P. Roetling, and N. Thakkar.
\newblock Aesthetic measures for automated document layout.
\newblock In {\em Proc. ACM Document Eng.}, 2004.

\bibitem{maskrcnn2017}
Kaiming He, Georgia Gkioxari, Piotr Doll{\'a}r, and Ross Girshick.
\newblock Mask r-cnn.
\newblock In {\em Proceedings of the IEEE international conference on computer
  vision}, pages 2961--2969, 2017.

\bibitem{resnet2016}
Kaiming He, Xiangyu Zhang, Shaoqing Ren, and Jian Sun.
\newblock Deep residual learning for image recognition.
\newblock In {\em Proceedings of the IEEE conference on computer vision and
  pattern recognition}, pages 770--778, 2016.

\bibitem{Hurst2009}
N. Hurst, W. Li, and K. Marriott.
\newblock Review of automatic document formatting.
\newblock In {\em Proc. ACM Document Eng.}, 2009.

\bibitem{lee2018cleannet}
Kuang-Huei Lee, Xiaodong He, Lei Zhang, and Linjun Yang.
\newblock Cleannet: Transfer learning for scalable image classifier training
  with label noise.
\newblock In {\em Proceedings of the IEEE Conference on Computer Vision and
  Pattern Recognition}, pages 5447--5456, 2018.

\bibitem{fpn2017}
Tsung-Yi Lin, Piotr Doll{\'a}r, Ross Girshick, Kaiming He, Bharath Hariharan,
  and Serge Belongie.
\newblock Feature pyramid networks for object detection.
\newblock In {\em Proceedings of the IEEE conference on computer vision and
  pattern recognition}, pages 2117--2125, 2017.

\bibitem{retinanet2017}
Tsung-Yi Lin, Priya Goyal, Ross Girshick, Kaiming He, and Piotr Doll{\'a}r.
\newblock Focal loss for dense object detection.
\newblock In {\em Proceedings of the IEEE international conference on computer
  vision}, pages 2980--2988, 2017.

\bibitem{lin2014microsoft}
Tsung-Yi Lin, Michael Maire, Serge Belongie, James Hays, Pietro Perona, Deva
  Ramanan, Piotr Doll{\'a}r, and C~Lawrence Zitnick.
\newblock Microsoft coco: Common objects in context.
\newblock In {\em European conference on computer vision}, pages 740--755.
  Springer, 2014.

\bibitem{uist2018rico}
Thomas~F. Liu, Mark Craft, Jason Situ, Ersin Yumer, Radomir Mech, and Ranjitha
  Kumar.
\newblock Learning design semantics for mobile apps.
\newblock In {\em The 31st Annual ACM Symposium on User Interface Software and
  Technology}, UIST '18, pages 569--579, New York, NY, USA, 2018. ACM.

\bibitem{long2015fully}
Jonathan Long, Evan Shelhamer, and Trevor Darrell.
\newblock Fully convolutional networks for semantic segmentation.
\newblock In {\em Proc. CVPR}, 2015.

\bibitem{Manandhar2020}
D. Manandhar, D. Ruta, and J. Collomosse.
\newblock Learning structural similarity of user interface layouts using graph
  networks.
\newblock In {\em Proc. ECCV}, 2020.

\bibitem{marino2017more}
Kenneth Marino, Ruslan Salakhutdinov, and Abhinav Gupta.
\newblock The more you know: Using knowledge graphs for image classification.
\newblock In {\em Proceedings of the IEEE Conference on Computer Vision and
  Pattern Recognition}, pages 2673--2681, 2017.

\bibitem{Moran2018}
K. Moran, C. Bernal-Cardenas, M. Curcio, R. Bonett, and D. Poshyvanyk.
\newblock Machine learning-based prototyping of graphical user interfaces for
  mobile apps.
\newblock {\em IEEE Trans. Soft. Eng.}, 2018.

\bibitem{ODonovan2014}
P. O'Donovan, A. Agarwala, and A. Hertzmann.
\newblock Learning layouts for single-page graphic designs.
\newblock {\em IEEE Transactions on Visualization and Computer Graphics}, 2014.

\bibitem{ODonovan2015}
P. O'Donovan, A. Agarwala, and A. Hertzmann.
\newblock Designscape: Design with interactive layout suggestions.
\newblock In {\em Proc. ACM Conf. Human Factors in Comp. Sys.}, pages
  1221--1224, 2015.

\bibitem{pytorch}
Adam Paszke, Sam Gross, Soumith Chintala, Gregory Chanan, Edward Yang, Zachary
  DeVito, Zeming Lin, Alban Desmaison, Luca Antiga, and Adam Lerer.
\newblock Automatic differentiation in pytorch.
\newblock 2017.

\bibitem{faster-rcnn}
Shaoqing Ren, Kaiming He, Ross Girshick, and Jian Sun.
\newblock Faster r-cnn: Towards real-time object detection with region proposal
  networks.
\newblock In {\em Advances in neural information processing systems}, pages
  91--99, 2015.

\bibitem{Swearngin2018}
A. Swearngin, M. Dontcheva, W. Li, J. Brandt, M. Dixon, and A. Ko.
\newblock Rewire: Interface design assistance from examples.
\newblock In {\em Proc. ACM CHI}, 2018.

\bibitem{Uijlings2013}
J.~R. Uijlings, K.~E. Sande, T. Gevers, and A.~W. Smeulders.
\newblock Selective search for object recognition.
\newblock {\em Intl. Journal Computer Vision (IJCV)}, 2013.

\bibitem{wang2018zero}
Xiaolong Wang, Yufei Ye, and Abhinav Gupta.
\newblock Zero-shot recognition via semantic embeddings and knowledge graphs.
\newblock In {\em Proceedings of the IEEE conference on computer vision and
  pattern recognition}, pages 6857--6866, 2018.

\bibitem{detectron2}
Yuxin Wu, Alexander Kirillov, Francisco Massa, Wan-Yen Lo, and Ross Girshick.
\newblock Detectron2.
\newblock \url{https://github.com/facebookresearch/detectron2}, 2019.

\bibitem{Pang2016}
R.~Lau X.~Pang, Y.~Cao and A. Chan.
\newblock Directing user attention via visual flow on web designs.
\newblock In {\em Proc. ACM SIGGRAPH}, 2016.

\bibitem{xu2019spatial}
Hang Xu, Chenhan Jiang, Xiaodan Liang, and Zhenguo Li.
\newblock Spatial-aware graph relation network for large-scale object
  detection.
\newblock In {\em Proceedings of the IEEE Conference on Computer Vision and
  Pattern Recognition}, pages 9298--9307, 2019.

\bibitem{xu2019reasoning}
Hang Xu, ChenHan Jiang, Xiaodan Liang, Liang Lin, and Zhenguo Li.
\newblock Reasoning-rcnn: Unifying adaptive global reasoning into large-scale
  object detection.
\newblock In {\em Proceedings of the IEEE Conference on Computer Vision and
  Pattern Recognition}, pages 6419--6428, 2019.

\bibitem{Yang2017}
X. Yang, E. Yumer, P. Asente, M. Kraley, D. Kifer, and C. Giles.
\newblock Learning to extract semantic structure from documents using
  multimodal fully convolutional neural networks.
\newblock In {\em Proc. CVPR}, pages 5315--5324, 2017.

\bibitem{Dollar2014}
C.~L. Zitnick and P. Dollar.
\newblock Edge boxes: Locating object proposals from edges.
\newblock In {\em Proc. ECCV}, 2014.

\end{thebibliography}
}

\clearpage

\setcounter{equation}{0}
\setcounter{figure}{0}
\setcounter{table}{0}
\setcounter{page}{1}
\setcounter{section}{0}
\makeatletter
\renewcommand{\thetable}{S\arabic{table}}
\renewcommand{\theequation}{S\arabic{equation}}
\renewcommand{\thefigure}{S\arabic{figure}}
\renewcommand{\thesection}{S\arabic{section}}

\renewcommand{\cite}[1]{S#1}

\begin{strip}
\begin{center}
\textbf{\huge Magic Layouts - Supplementary Materials}
\end{center}
\end{strip}

\begin{table*} [b]
\caption{\textbf{Performance of proposed method (FPN+Magic) on RICO dataset \cite{deka2017Rico} different $Nr$ values }}
\label{tab:nr_rico} 
\centering
\footnotesize{
        \begin{tabular}{|l||  c c c||  ccc|| ccc|| ccc|| }
        \hline
        
		Method  			& AP 	&AP50	&AP75 	&APs 	&APm 	&APl 	&AR 	&AR 	 &AR   &ARs  &ARm   &ARl\\ 
		@IoU				&0.5:95 	&0.5 	&0.75   &0.5:95 &0.5:95 &0.5:95 &0.5:95 &0.5:95 &0.5:95   &0.5:95 &0.5:95 &0.5:95 \\
		maxDets				& 100	&100	&100	&100	&100	&100	&1		&10		&100  &100	 	&100	&100	\\	\hline \hline 
		FPN+Magic Nr=64     & 48.5 &58.5 &51.4 &7.0 &31.4 &48.5 &42.2 &61.8 &63.9 &12.8 &45.0 &63.4 \\
        FPN+Magic Nr=128    & 49.5 &59.4 &52.3 &8.1 &33.8 &49.5 &42.8 &62.6 &64.7 &13.2 &46.4 &64.1 \\

		FPN+Magic Nr=256    &50.3	&60.1	&53.4	&8.4	&34.7	&50.2	&43.0	&63.0	 &65.0   &13.2   &46.4   &64.5 \\
		FPN+Magic Nr=512    &50.8   &60.5   &53.7   &9.1    &36.6   &50.7   &42.5   &62.4    &64.4   &14.2   &48.3   &63.7 \\
        \hline
        \end{tabular}
}        
\end{table*}

\begin{table*} [b]
\caption{\textbf{Performance of proposed method (FPN+Magic) on DrawnUI dataset \cite{drawnUI2020} different $Nr$ values }}
\label{tab:nr_dui} 
\centering
\footnotesize{
        \begin{tabular}{|l||  c c c||  ccc|| ccc|| ccc|| }
        \hline
        
		Method  			& AP 	&AP50	&AP75 	&APs 	&APm 	&APl 	&AR 	&AR 	 &AR   &ARs  &ARm   &ARl\\ 
		@IoU				&0.5:95 	&0.5 	&0.75   &0.5:95 &0.5:95 &0.5:95 &0.5:95 &0.5:95 &0.5:95   &0.5:95 &0.5:95 &0.5:95 \\
		maxDets				& 100	&100	&100	&100	&100	&100	&1		&10		&100  &100	&100	&100	\\	\hline \hline	 
        FPN+Magic Nr=64       &61.8 &87.8   &71.5   &34.2   &58.8   &63.2   &28.9   &65.2   &69.4 &38.9 &66.3   &69.6 \\       
        FPN+Magic Nr=128      &63.3 &89.0   &73.8   &32.6   &59.0   &65.1   &30.5   &66.8   &71.2 &37.6 &66.4   &71.5 \\
        FPN+Magic Nr=256	  &64.3	&89.5	&74.4	&32.2	&54.4	&66.5	&30.3	&66.7	&71.0 &37.1 &64.0   &71.7 \\
        FPN+Magic Nr=512      &64.1 &89.7   &75.3   &32.8   &59.9   &65.9   &30.2   &67.0   &71.2 &37.4 &65.7   &71.7 \\ 
        \hline
        \end{tabular}
}        
\end{table*}

\section{Magic Layouts Tool}
We have included a demo video for Magic Layouts showing how UXs are parsed and rapidly converted to digital prototypes  demonstrating its practical use. Please watch the demo video.

\section{Impact of $Nr$}
We conduct experiments to evaluate the impact of number of proposal sampled from each image ($N_r$) during training. Table \ref{tab:nr_rico} and Table \ref{tab:nr_dui} show the performance of the proposed Magic Layout with FPN architecture for various $Nr= \lbrace64, 128,256,512\rbrace$ on RICO and DrawnUI dataset respectively. Both mean Average Precision (AP) and Average Recall (AR) slightly decreases (1-2\%) while using lower $Nr$ value \eg 64, 128. Overall for both RICO and DrawnUI datasets, Nr=256 is sufficient for obtaining good precision and recall metrics.       

\section {Additional Results}
Fig. \ref{fig:rico-1} \& \ref{fig:rico-2} and Fig. \ref{fig:dui-1} \& \ref{fig:dui-2} provide additional examples of UX parsed by the proposed Magic Layout for RICO \cite{deka2017Rico} and DrawnUI \cite{drawnUI2020} respectively.

\newcommand{\imgw}{4.2cm}

\begin{figure*}[!b]
    \centering
    \begin{tabular}{ccc}
    \multicolumn{3}{l}{\textbf{ \hspace{-0.5cm} A. RICO Screenshots}} \\
      \includegraphics[width=\imgw]{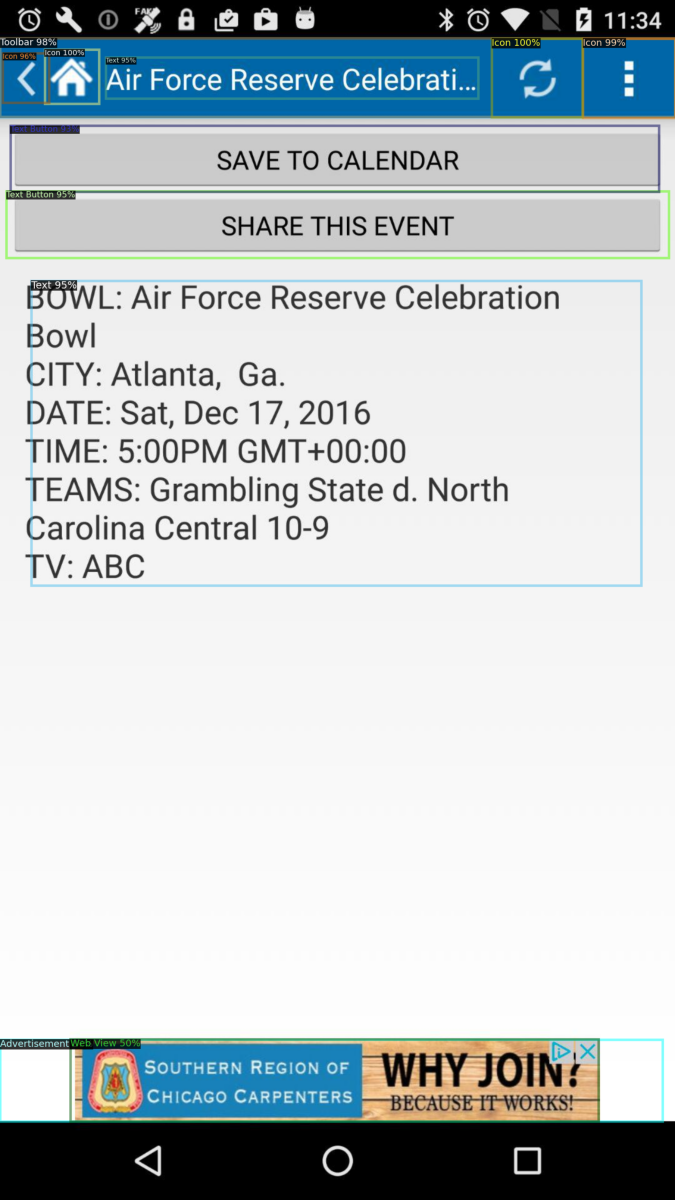} &        \includegraphics[width=\imgw]{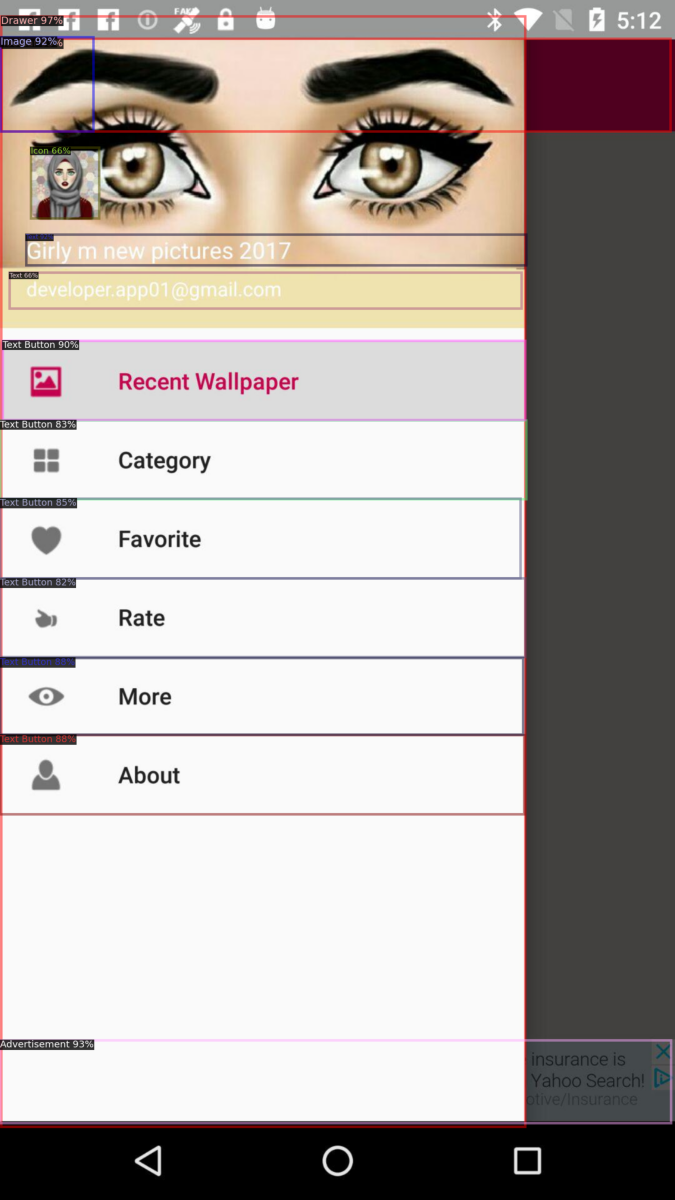} & \includegraphics[width=\imgw]{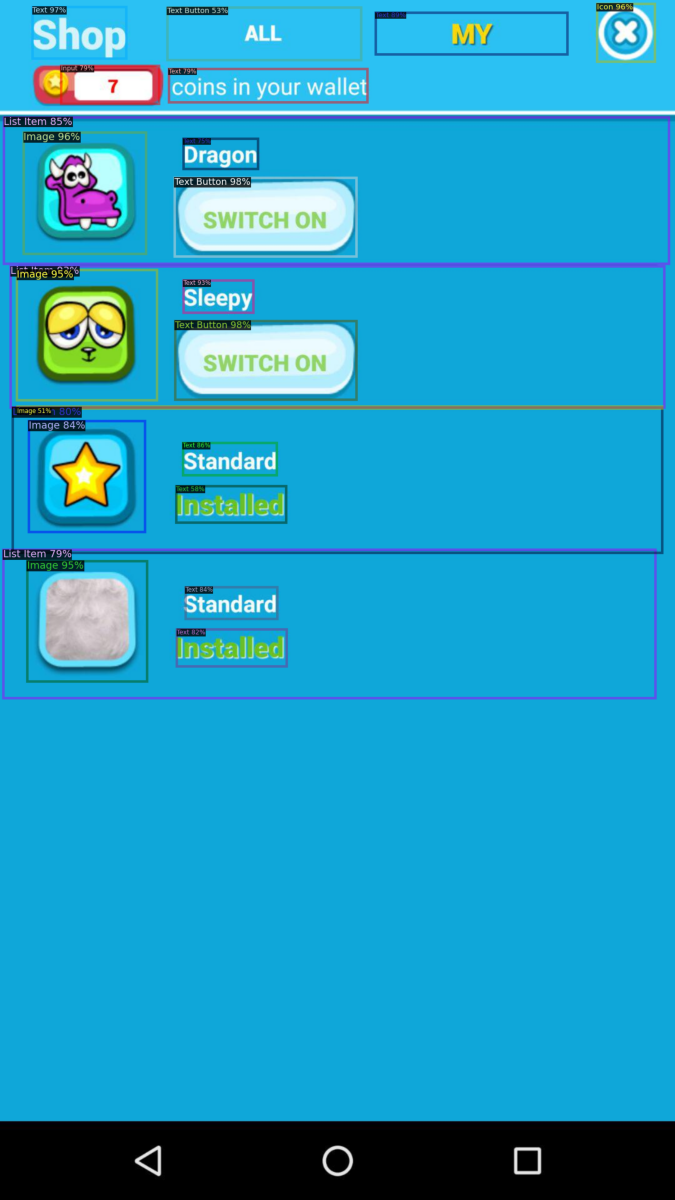} \\
      \includegraphics[width=\imgw]{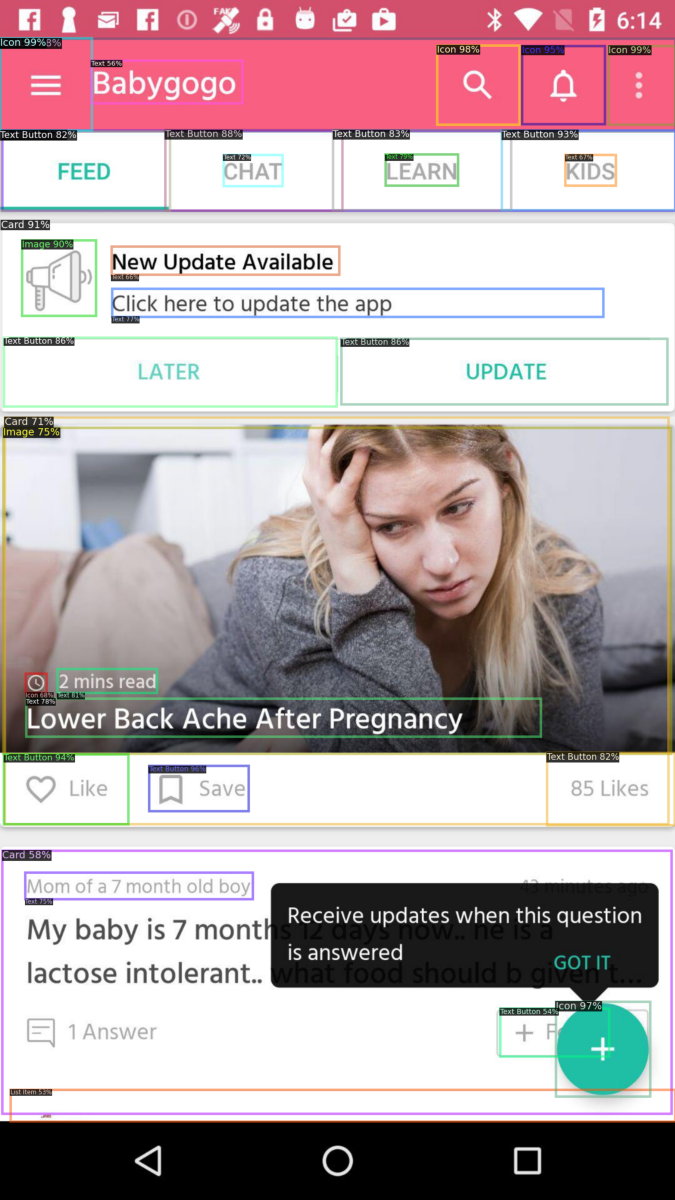} &       \includegraphics[width=\imgw]{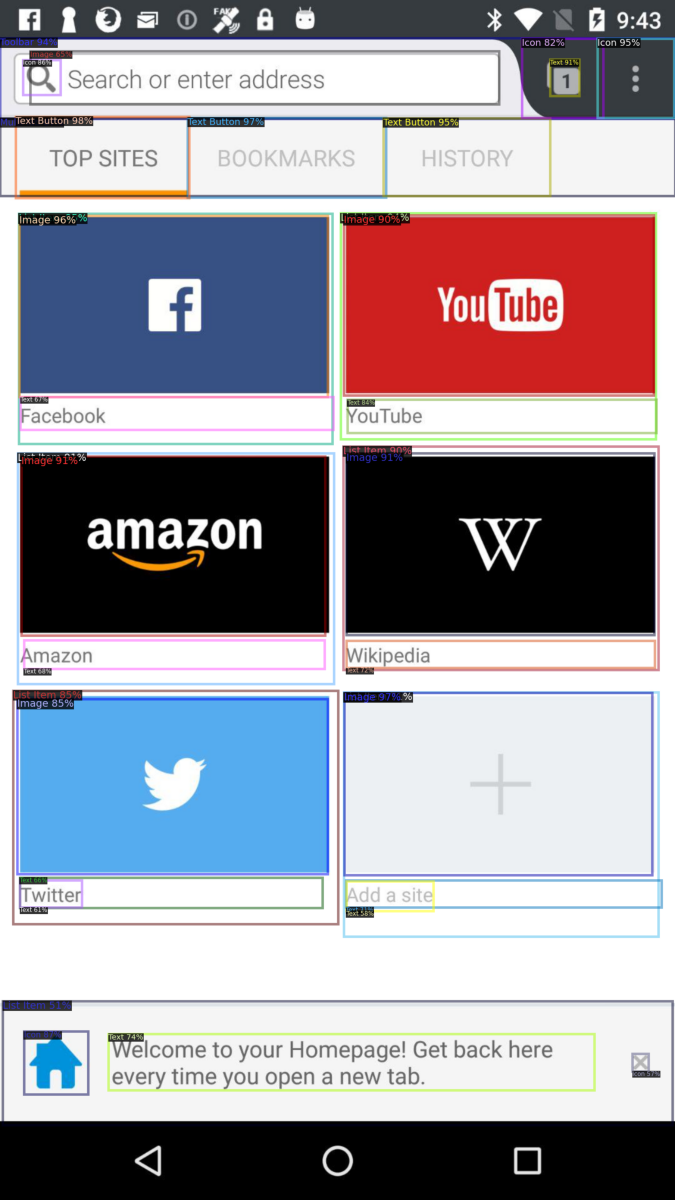} & \includegraphics[width=\imgw]{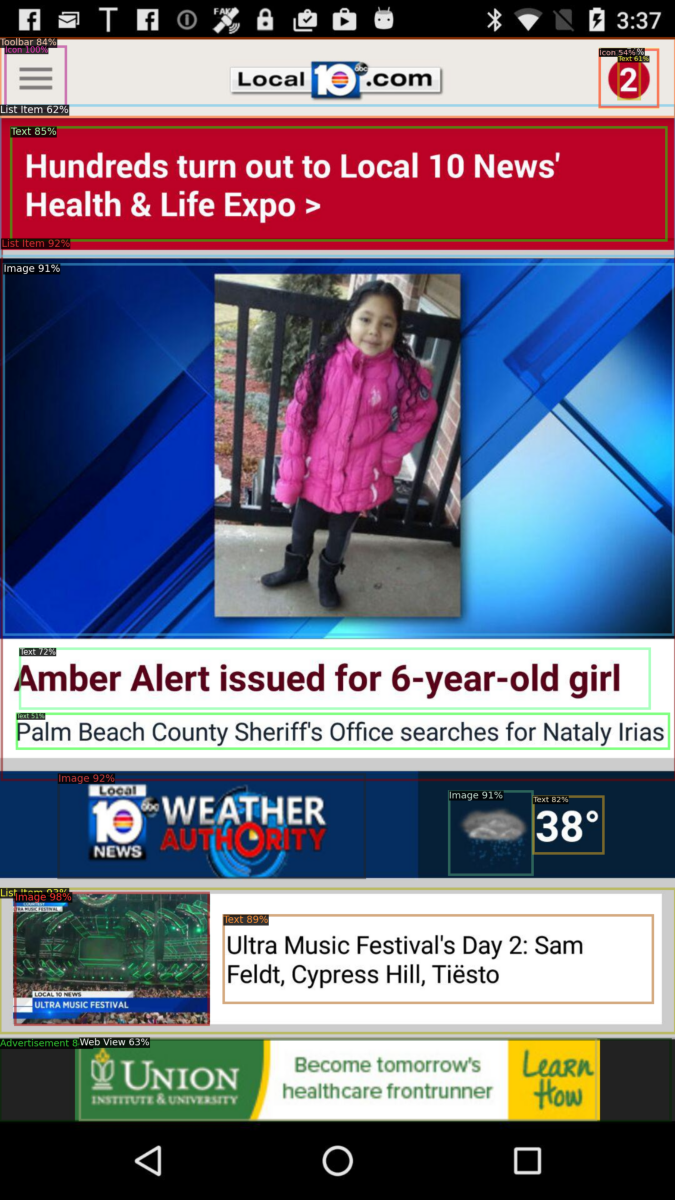} \\
      \includegraphics[width=\imgw]{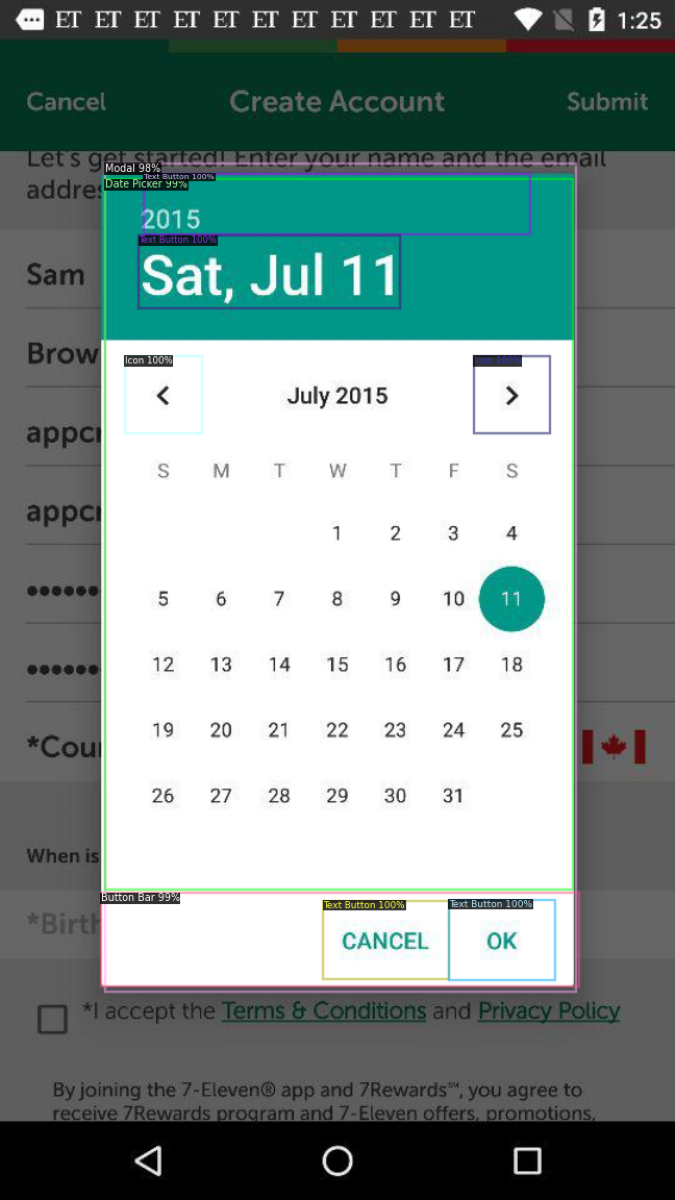} &       \includegraphics[width=\imgw]{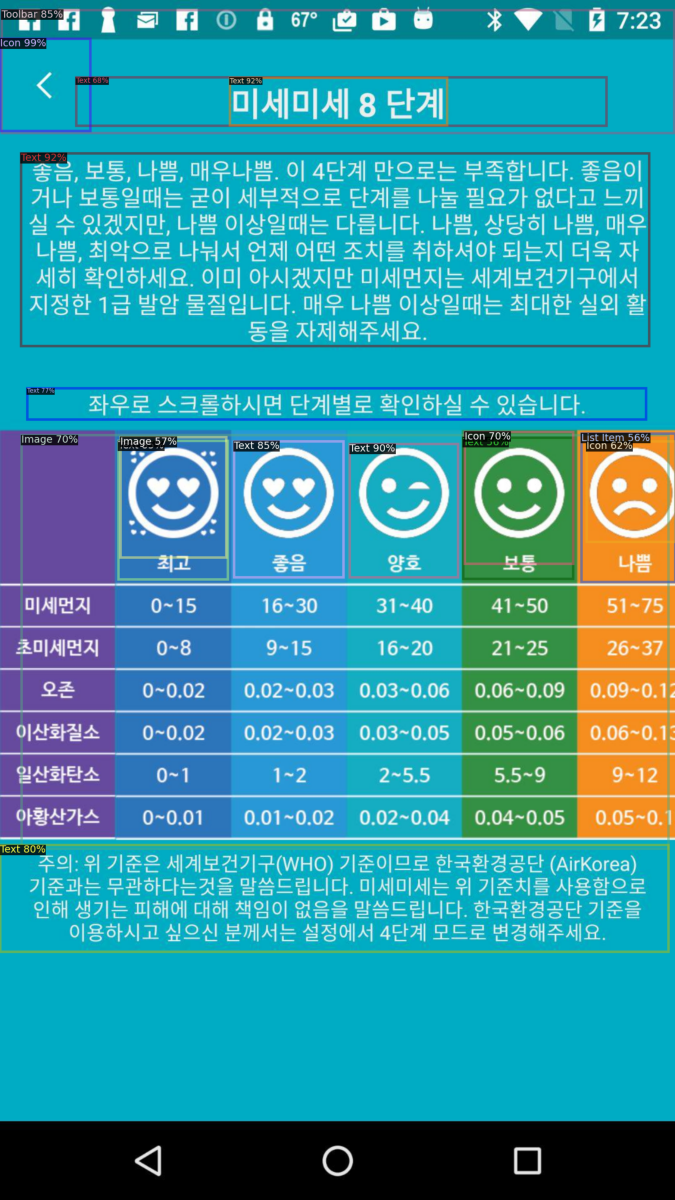} &  \includegraphics[width=\imgw]{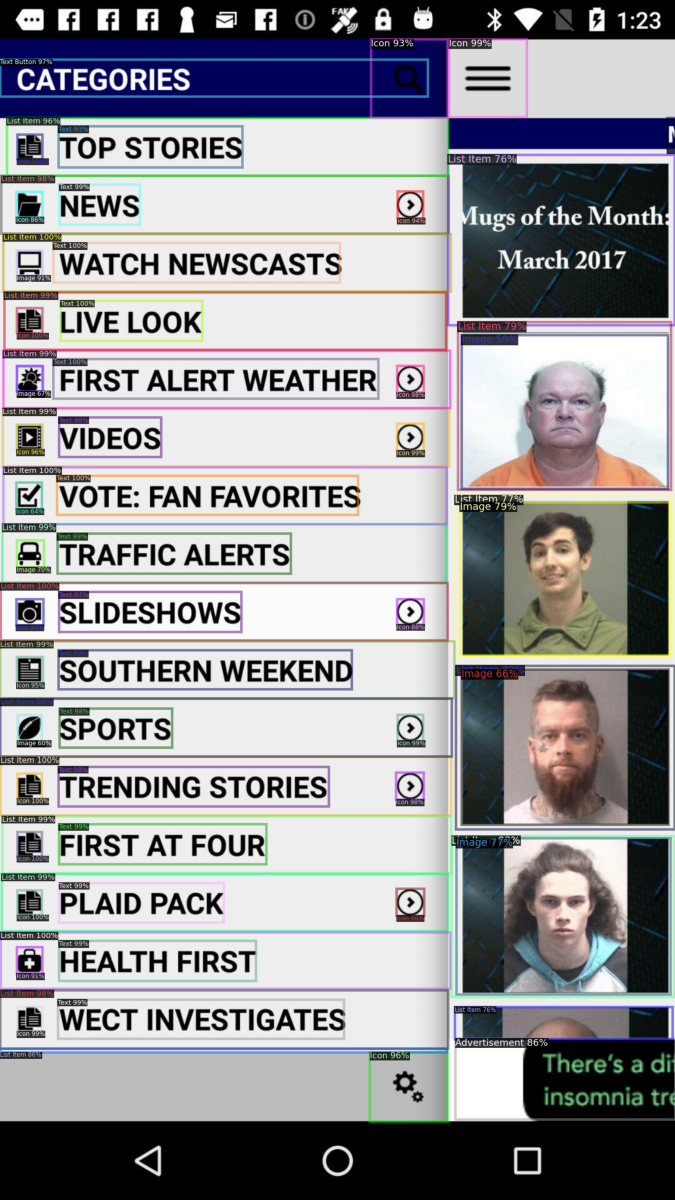}\\
    \end{tabular}   
    \caption{RICO screenshots parsed by Magic Layouts - 1.}
    \label{fig:rico-1}
    \vspace{-3mm}
\end{figure*}

\begin{figure*}[!b]
    \centering
    \begin{tabular}{ccc}
      \includegraphics[width=\imgw]{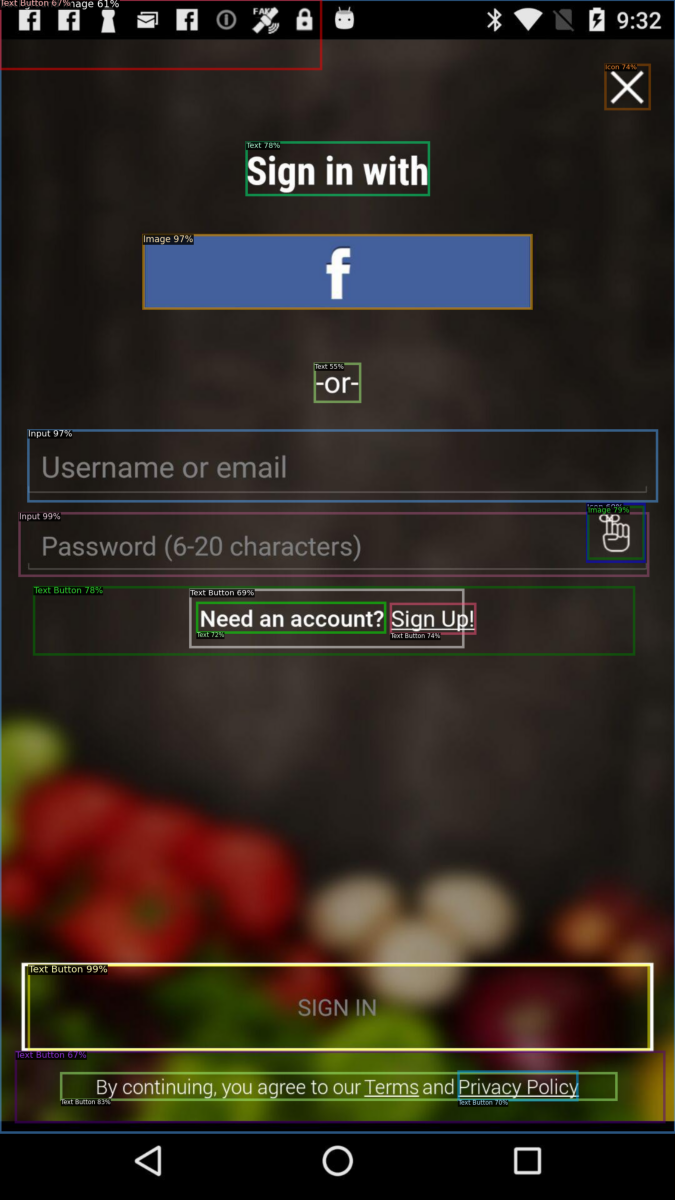} &  \includegraphics[width=\imgw]{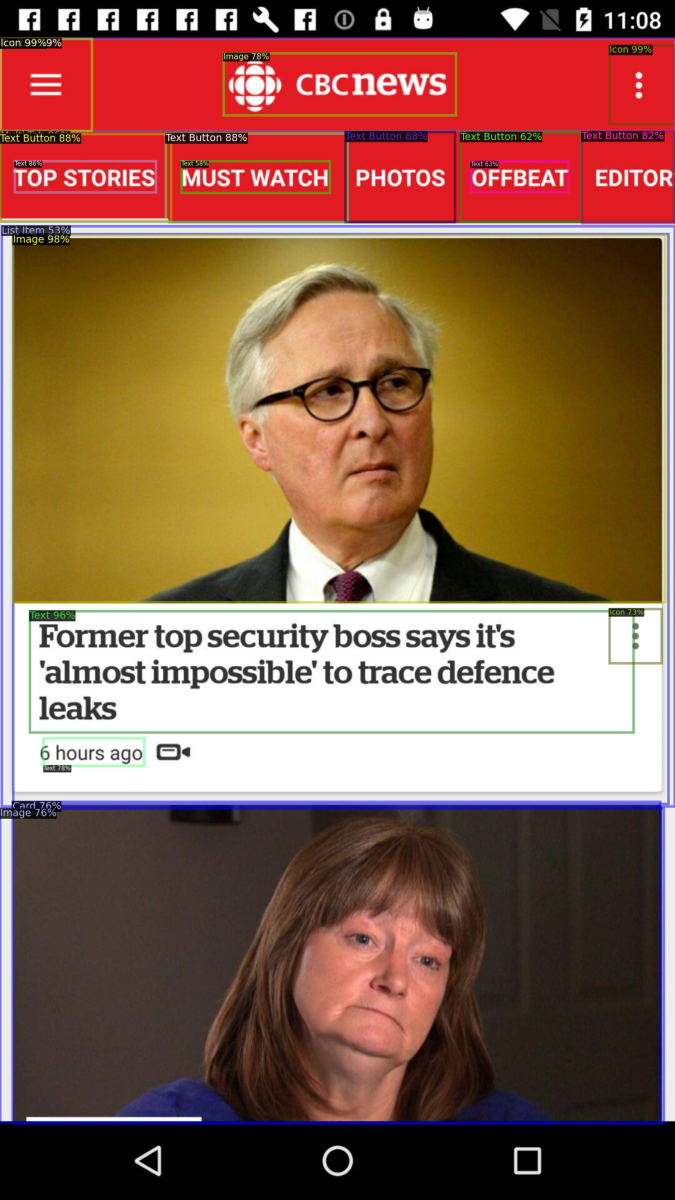}  &  \includegraphics[width=\imgw]{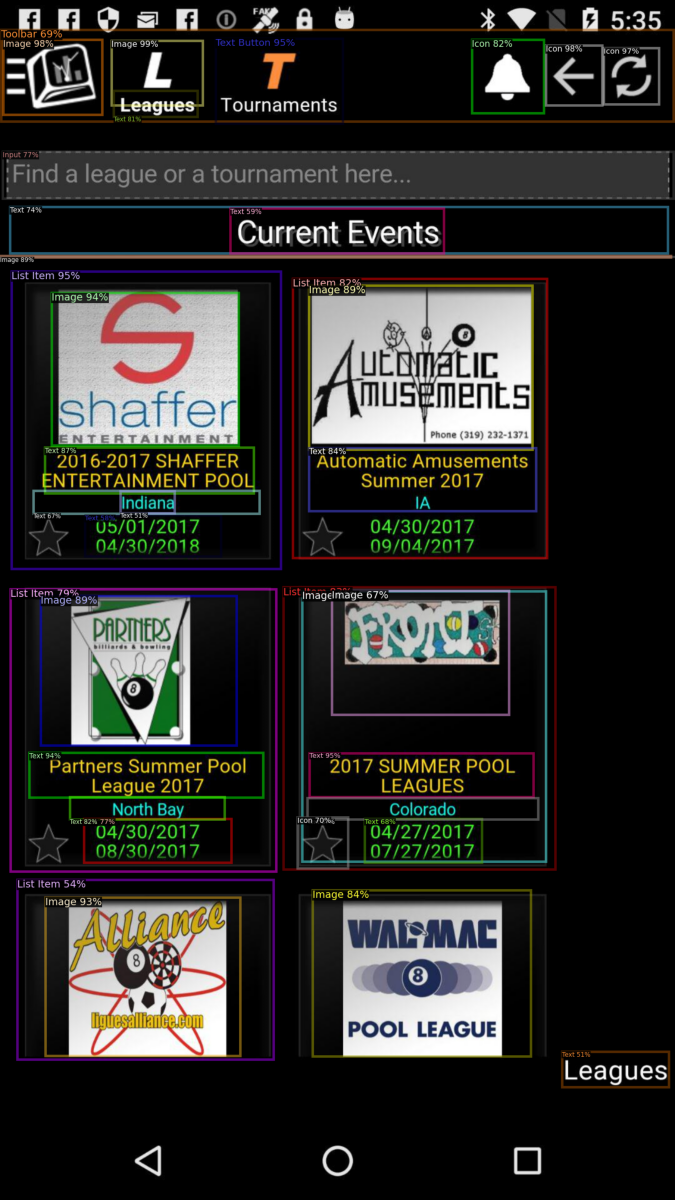} \\
      \includegraphics[width=\imgw]{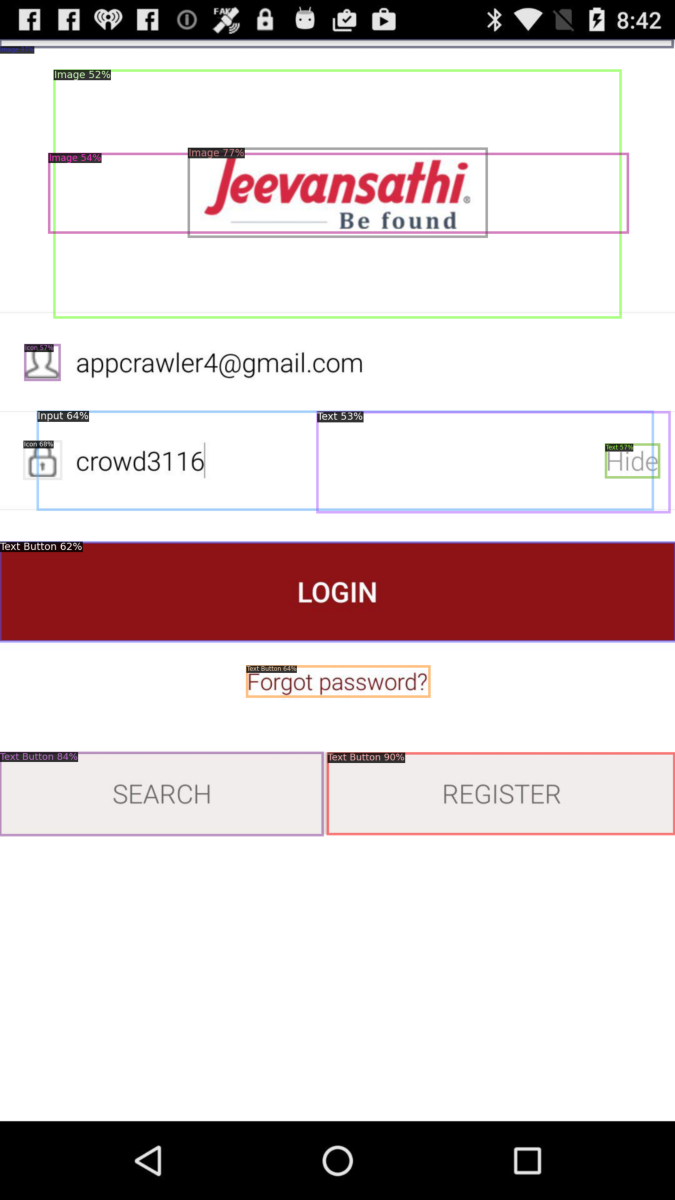} &       \includegraphics[width=\imgw]{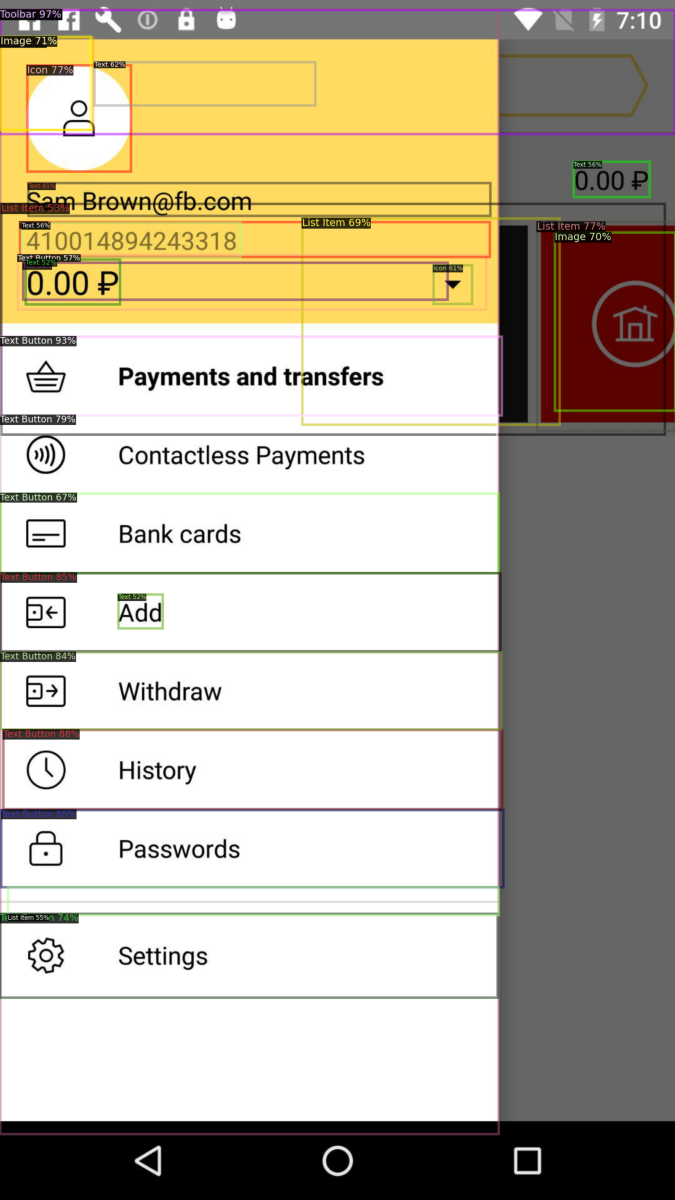}  &  \includegraphics[width=\imgw]{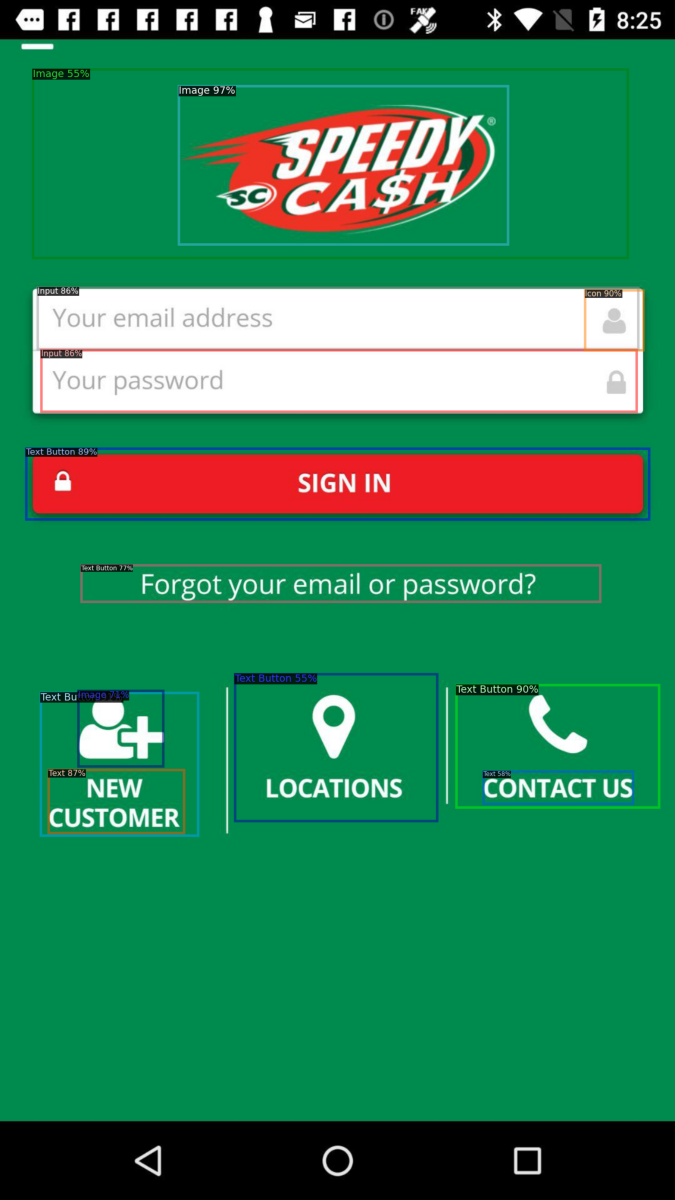} \\
      \includegraphics[width=\imgw]{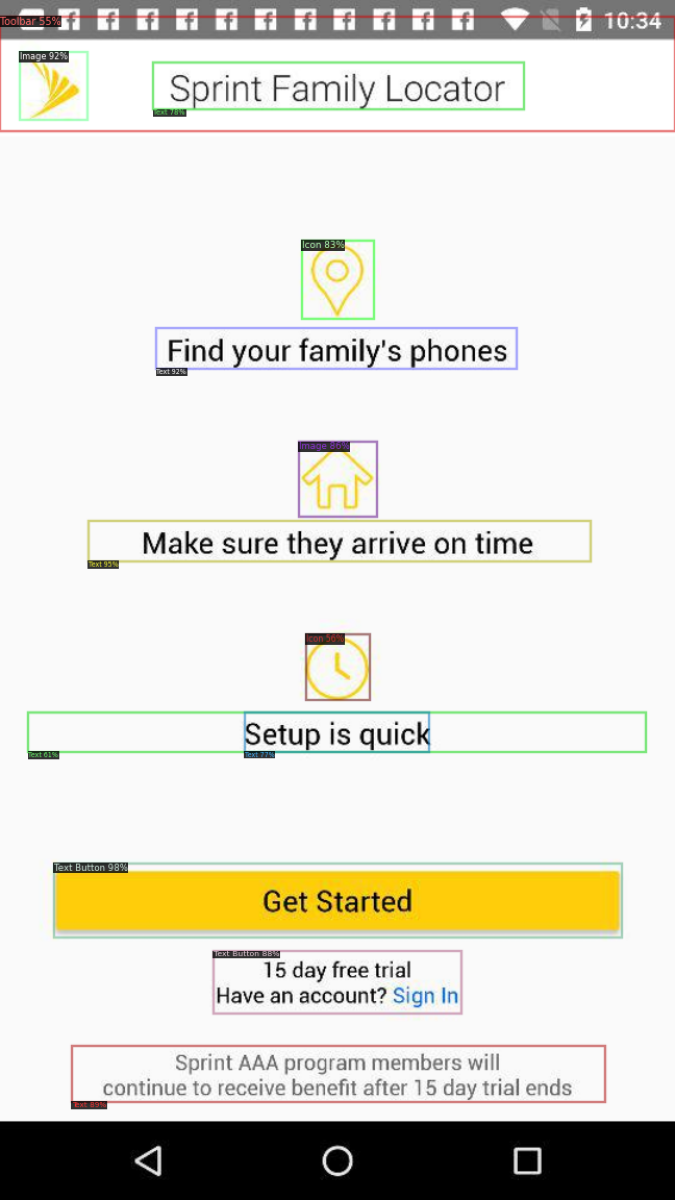} &       \includegraphics[width=\imgw]{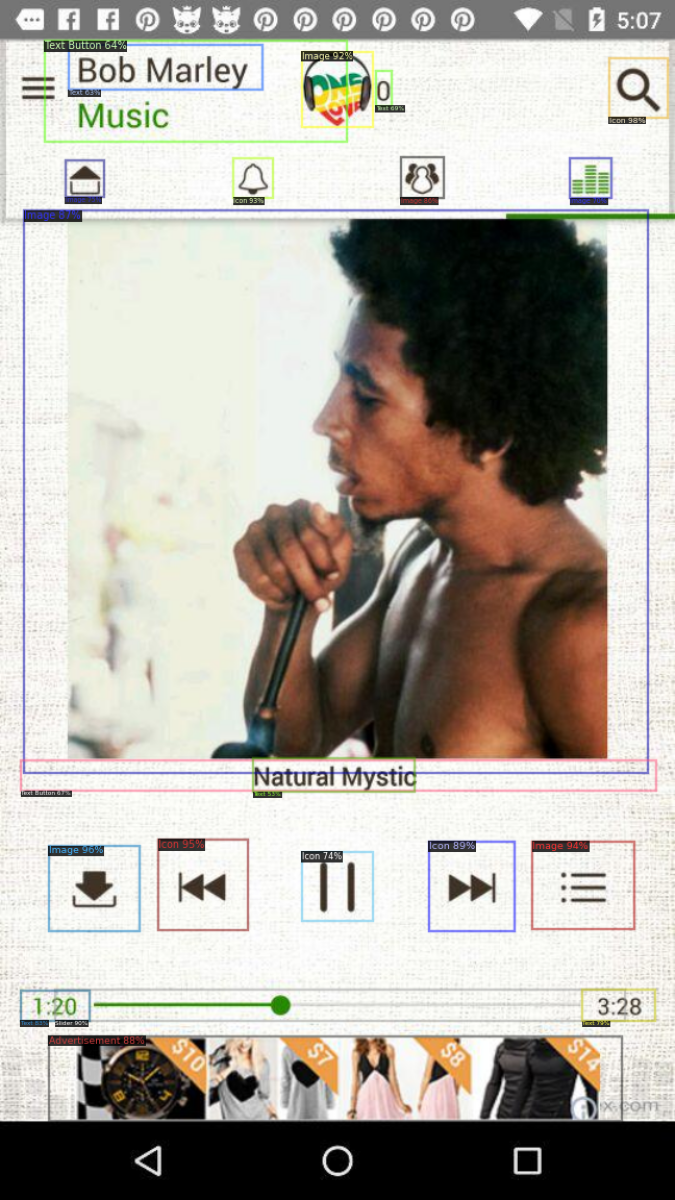}  &  \includegraphics[width=\imgw]{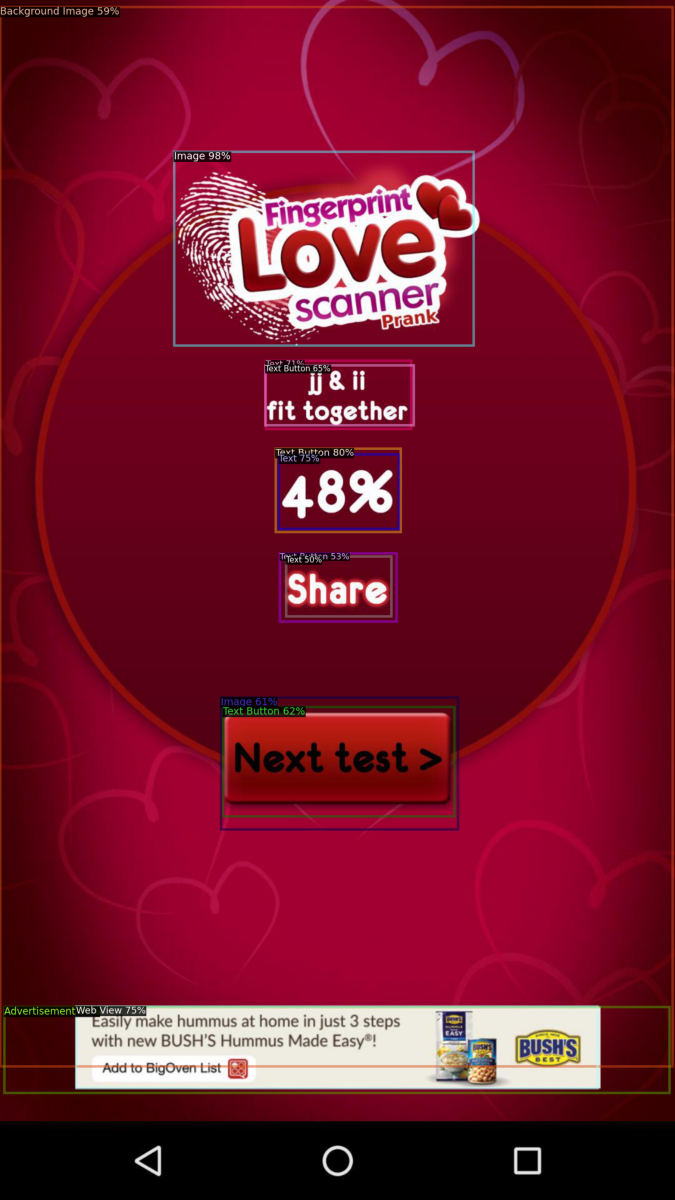} \\
    \end{tabular}   
    \caption{RICO screenshots parsed by Magic Layouts - 2.}
    \label{fig:rico-2}
    \vspace{-3mm}
\end{figure*}

\newcommand{\imgwd}{5.3cm}

\begin{figure*}[!h]
    \centering
    \begin{tabular}{ccc} 
    \multicolumn{3}{l} {\textbf{\hspace{-0.5cm} B. DrawnUI hand-sketched UXs}} \\ \vspace{2mm}
      \includegraphics[width=\imgwd]{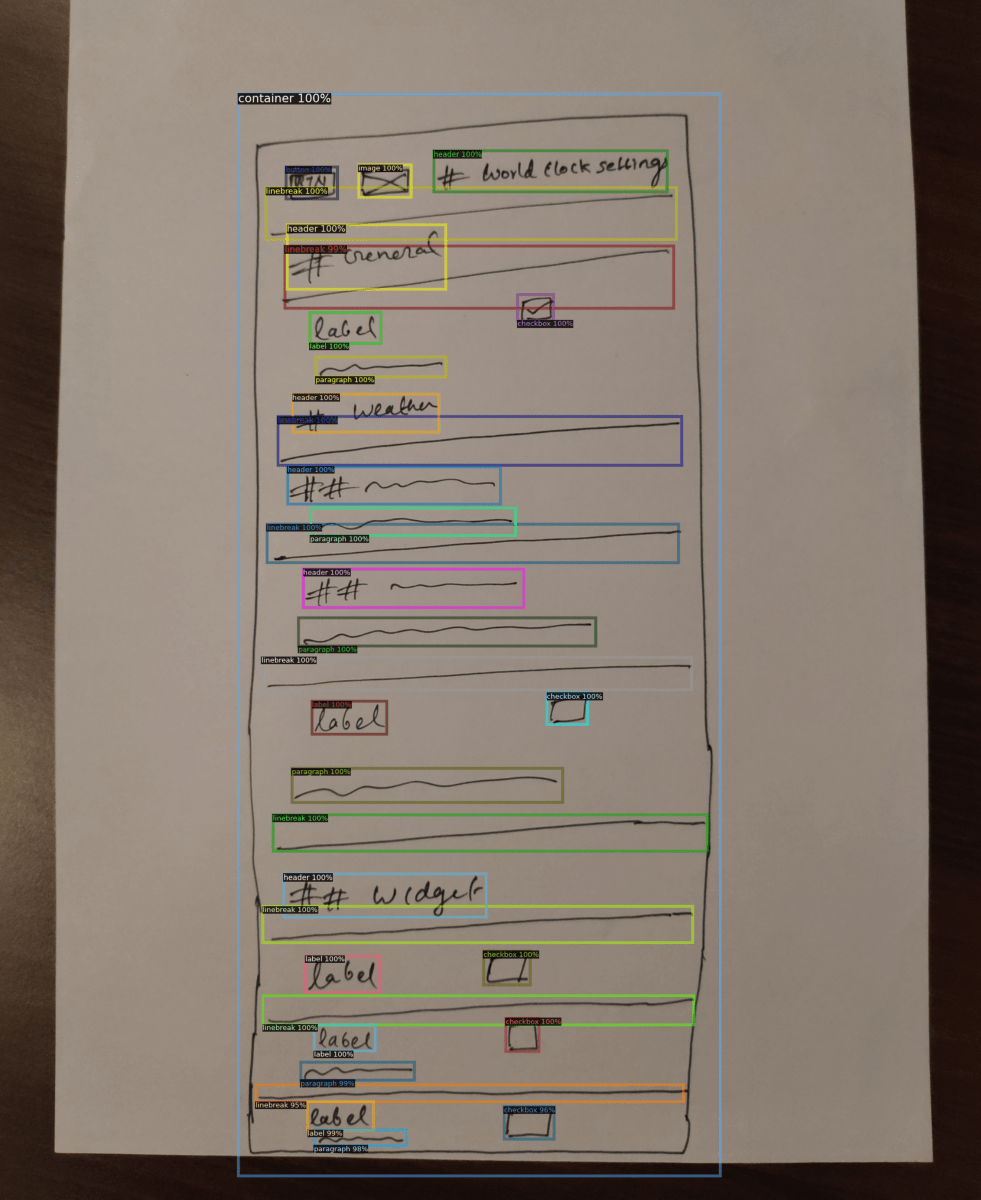} &  \includegraphics[width=\imgwd]{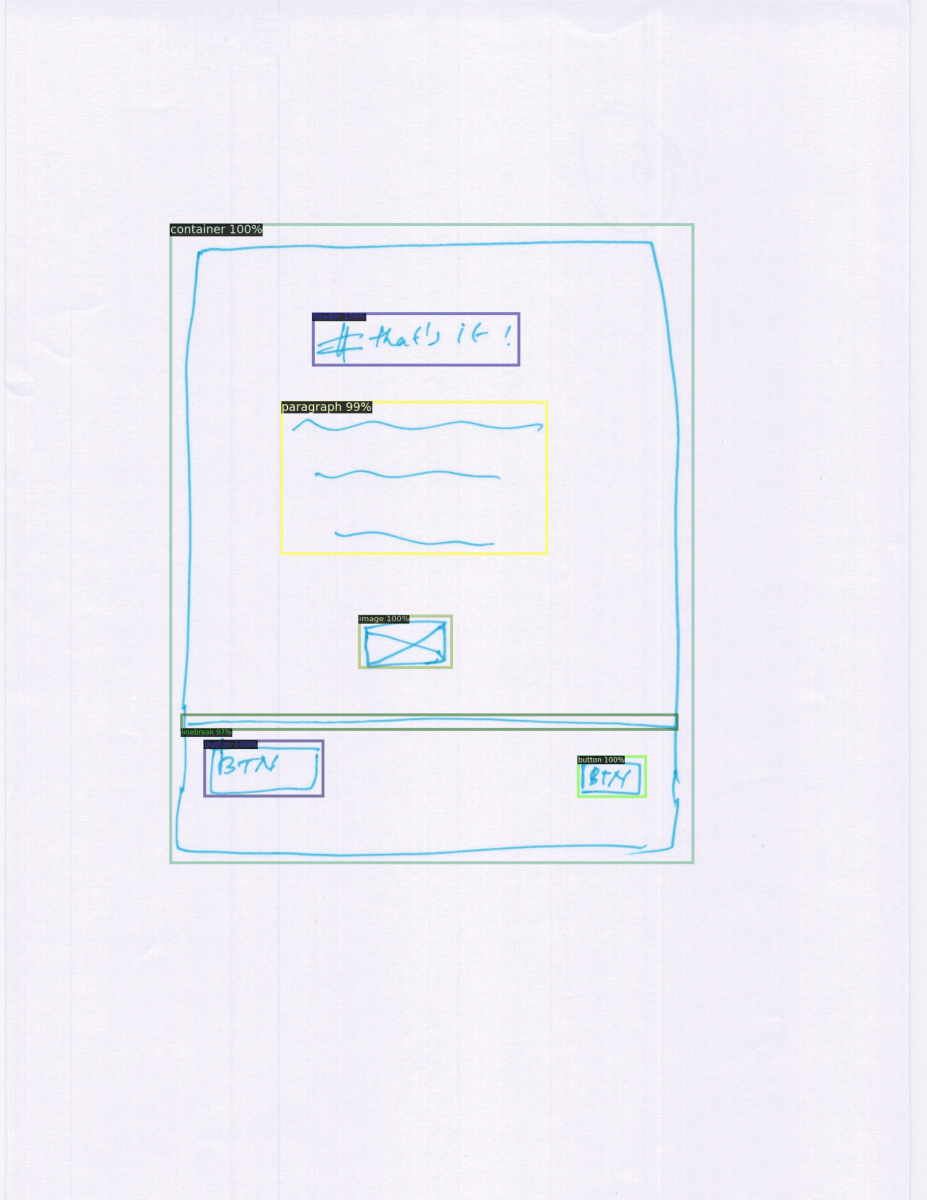} &  \includegraphics[width=\imgwd]{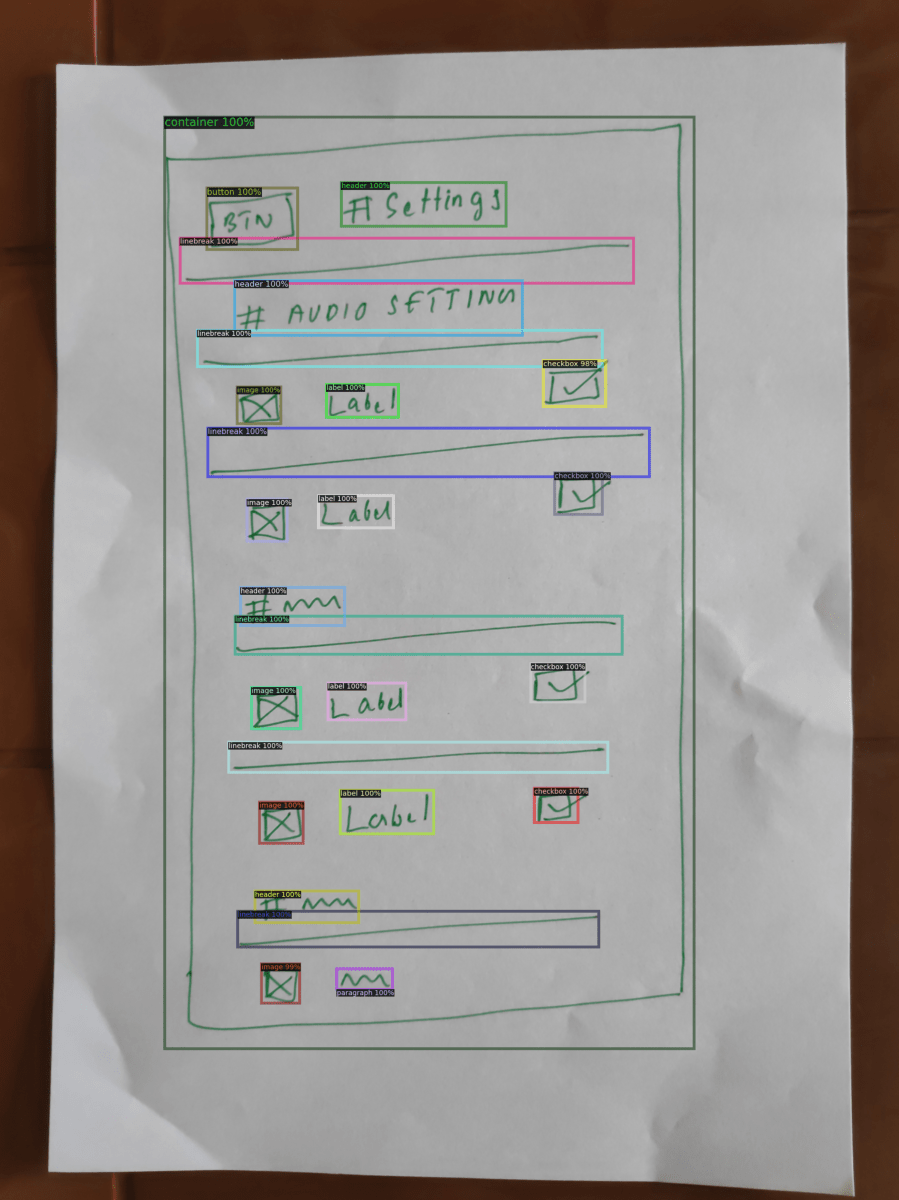} \\
      \includegraphics[width=\imgwd] {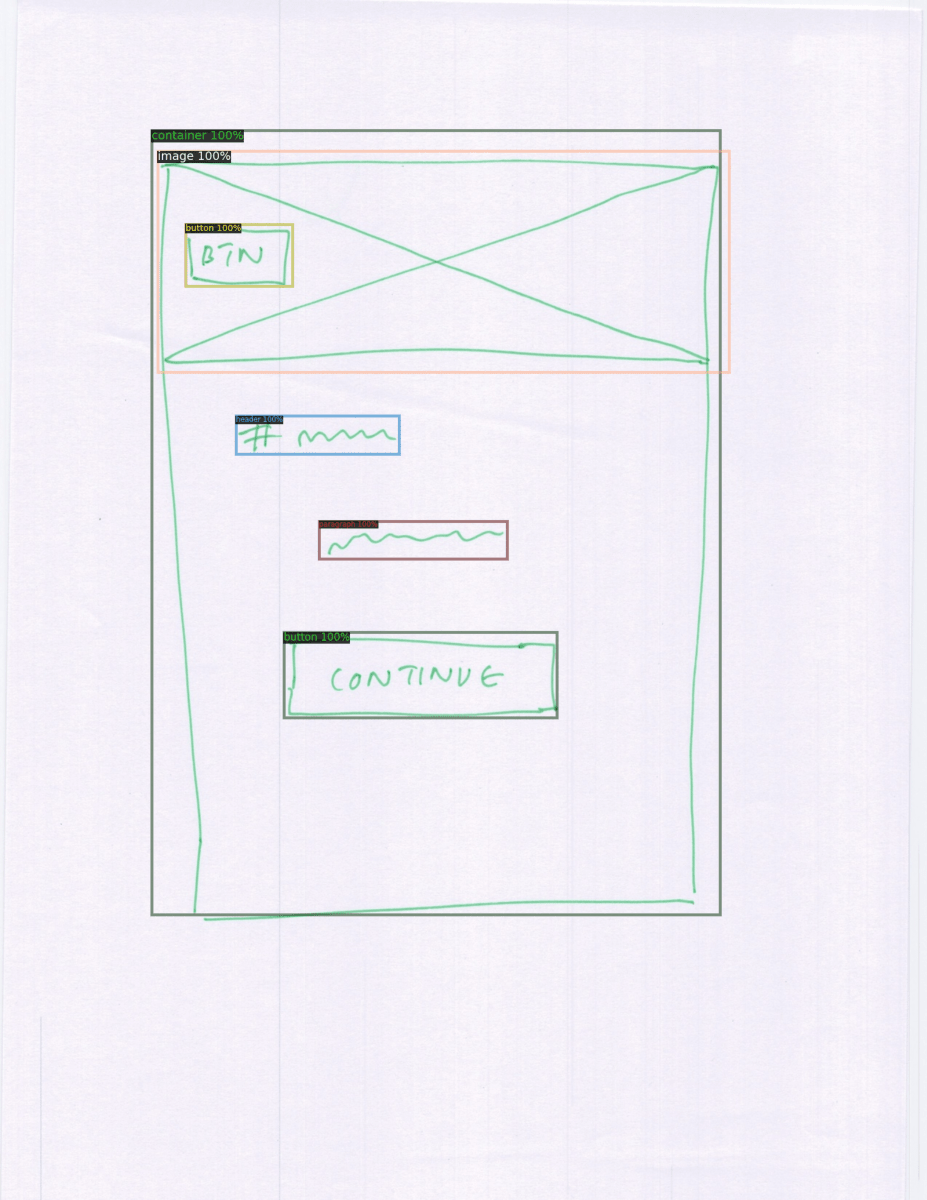}&       \includegraphics[width=\imgwd]{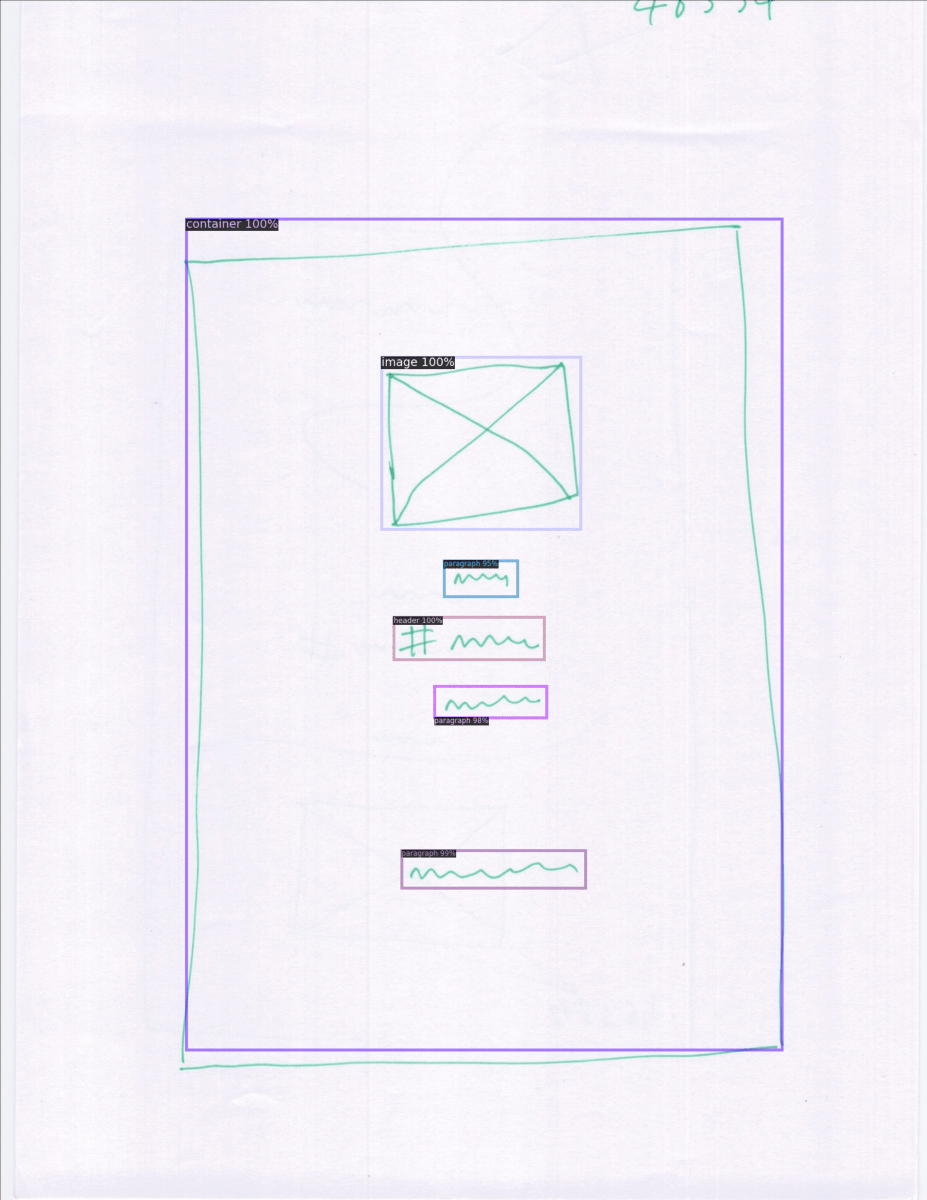} &  \includegraphics[width=\imgwd]{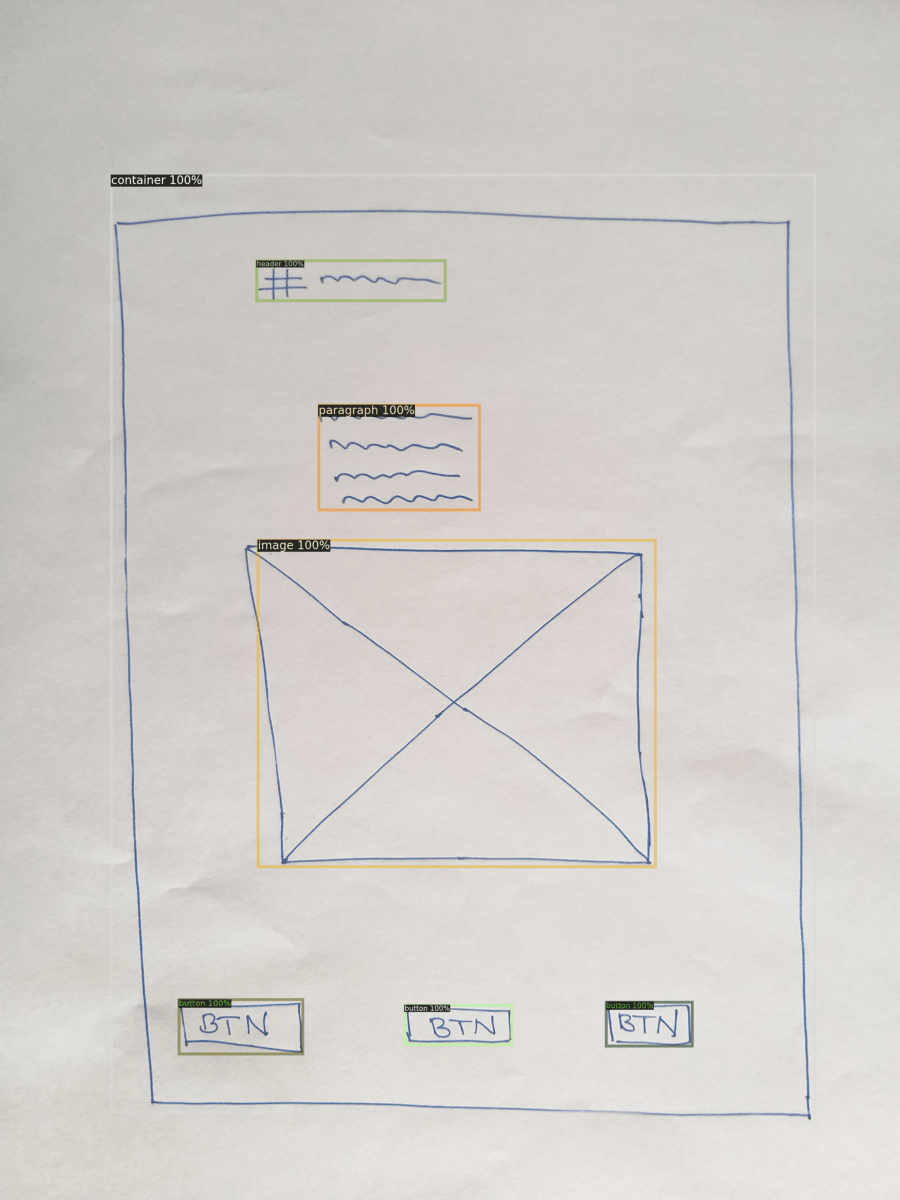} \\
      \includegraphics[width=\imgwd]{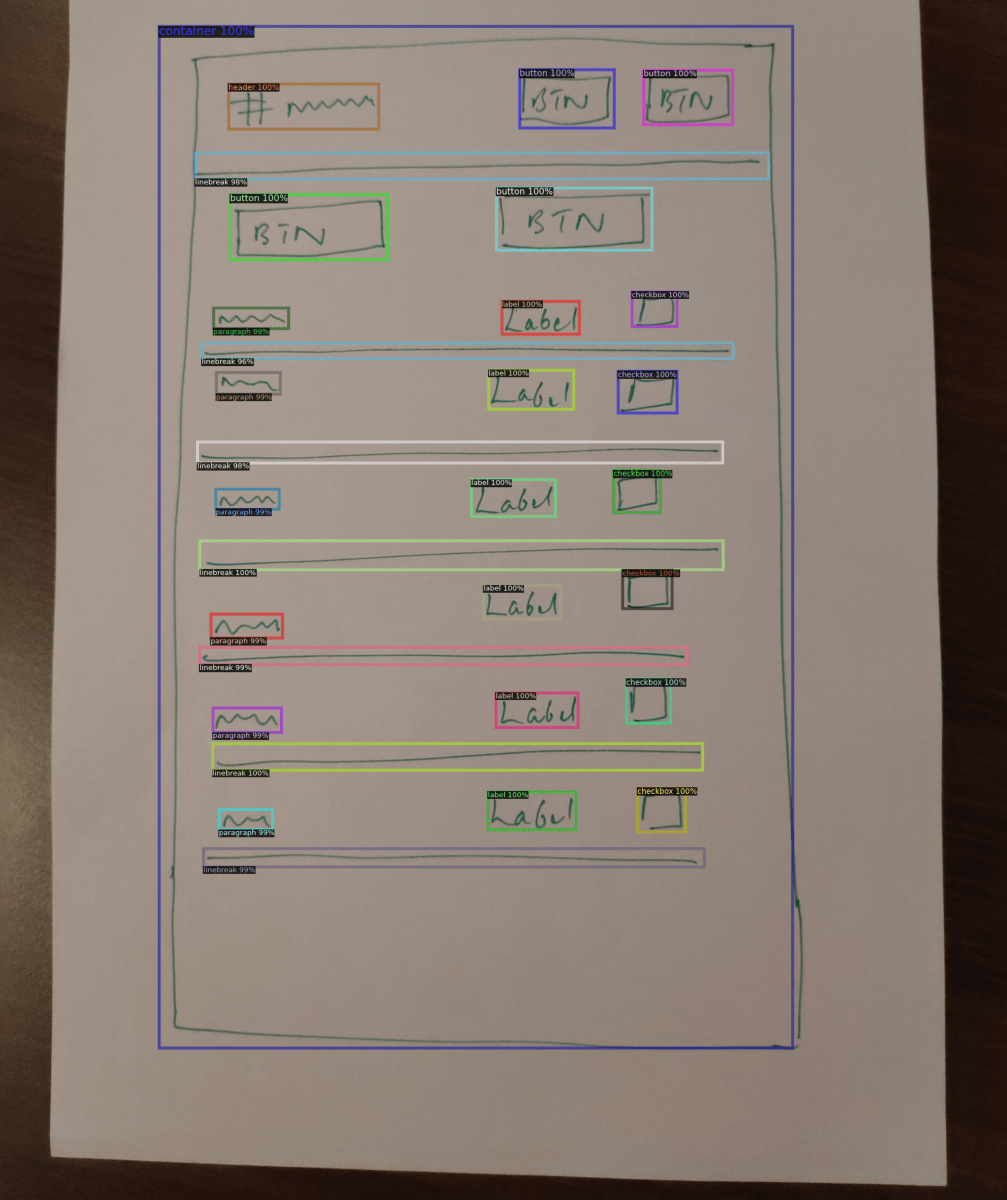} &       \includegraphics[width=\imgwd]{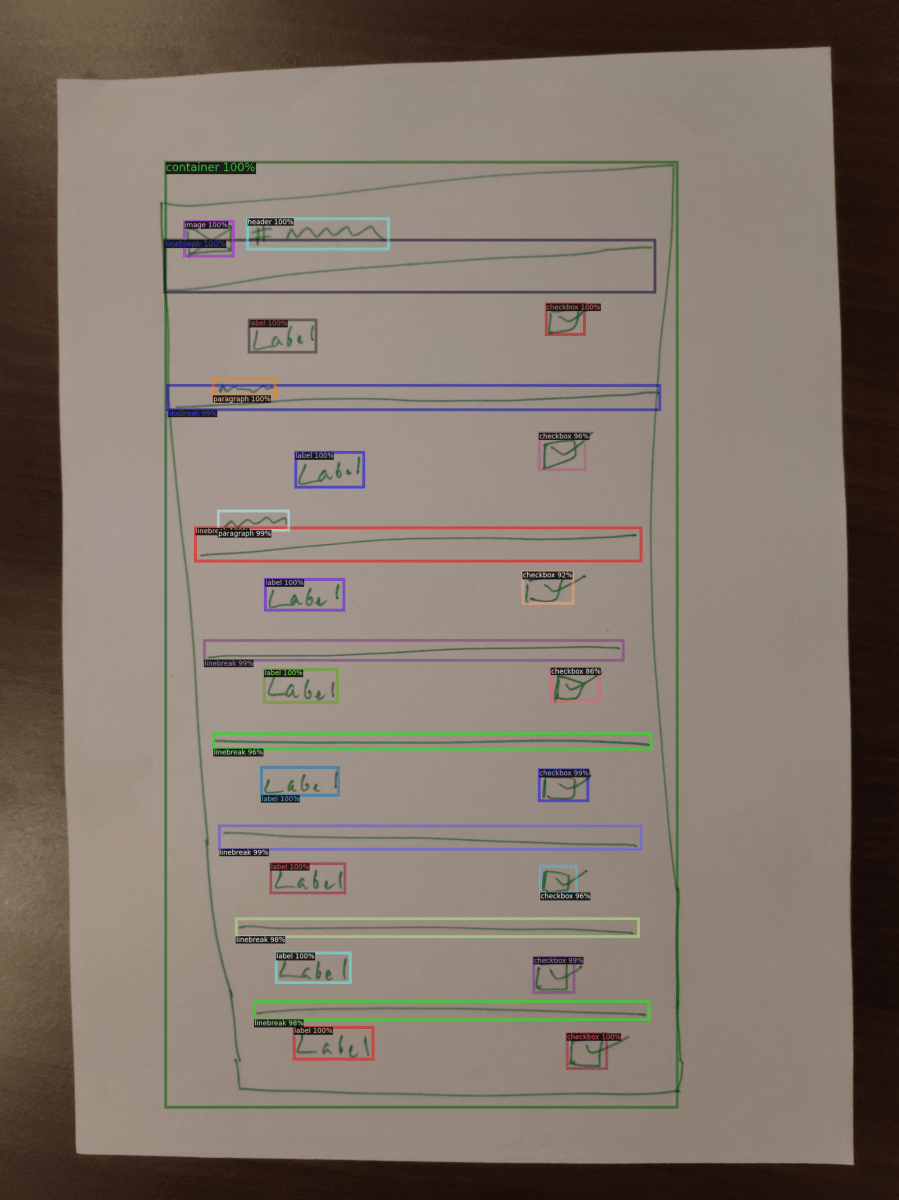} &  \includegraphics[width=\imgwd]{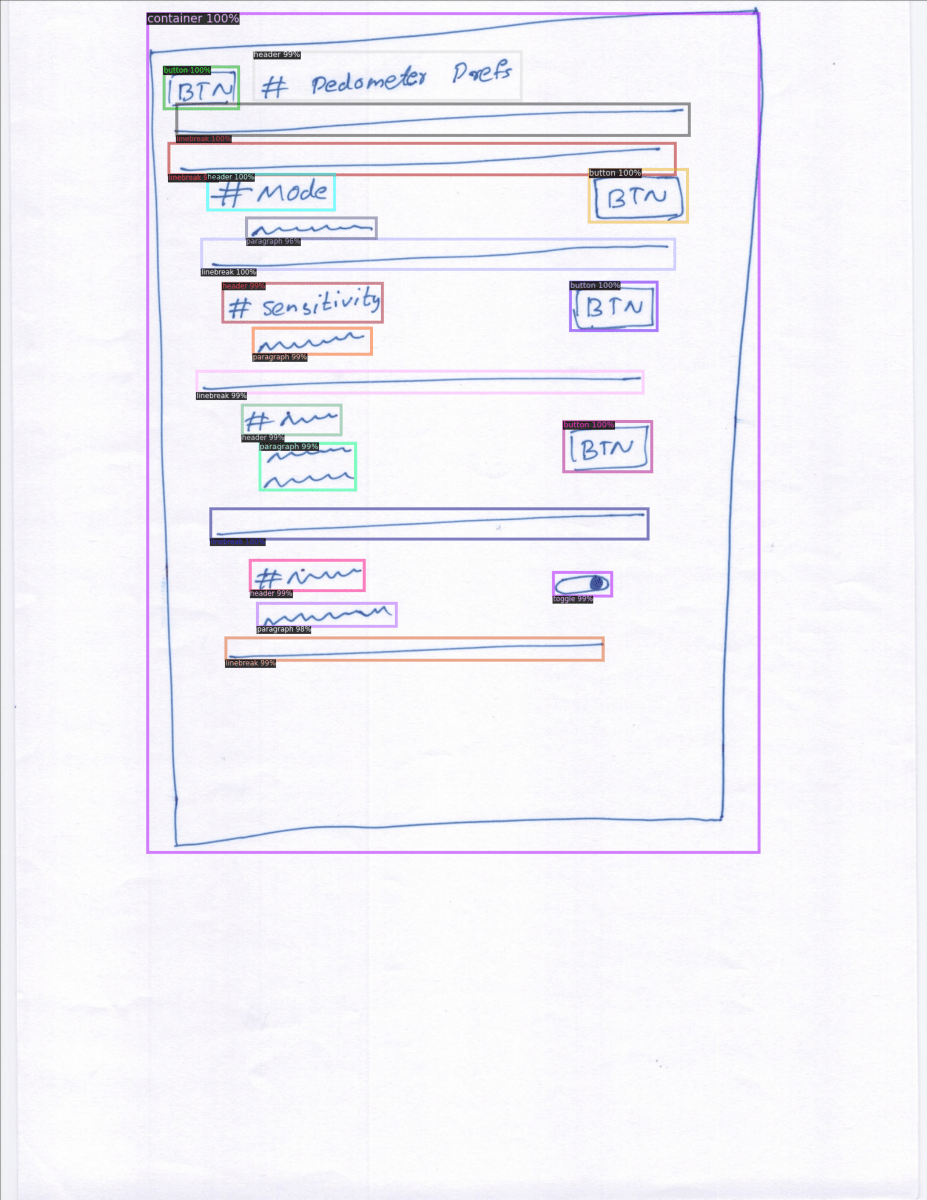}\\
    \end{tabular}   
    \caption{DrawnUI UX parsed by Magic Layouts - 1.}
    \label{fig:dui-1}
    \vspace{-3mm}
\end{figure*}

\newcommand{\imgwl}{7.5cm}

\begin{figure*}[!h]
    \centering
    \begin{tabular}{ccc}
      \includegraphics[width=\imgwl]{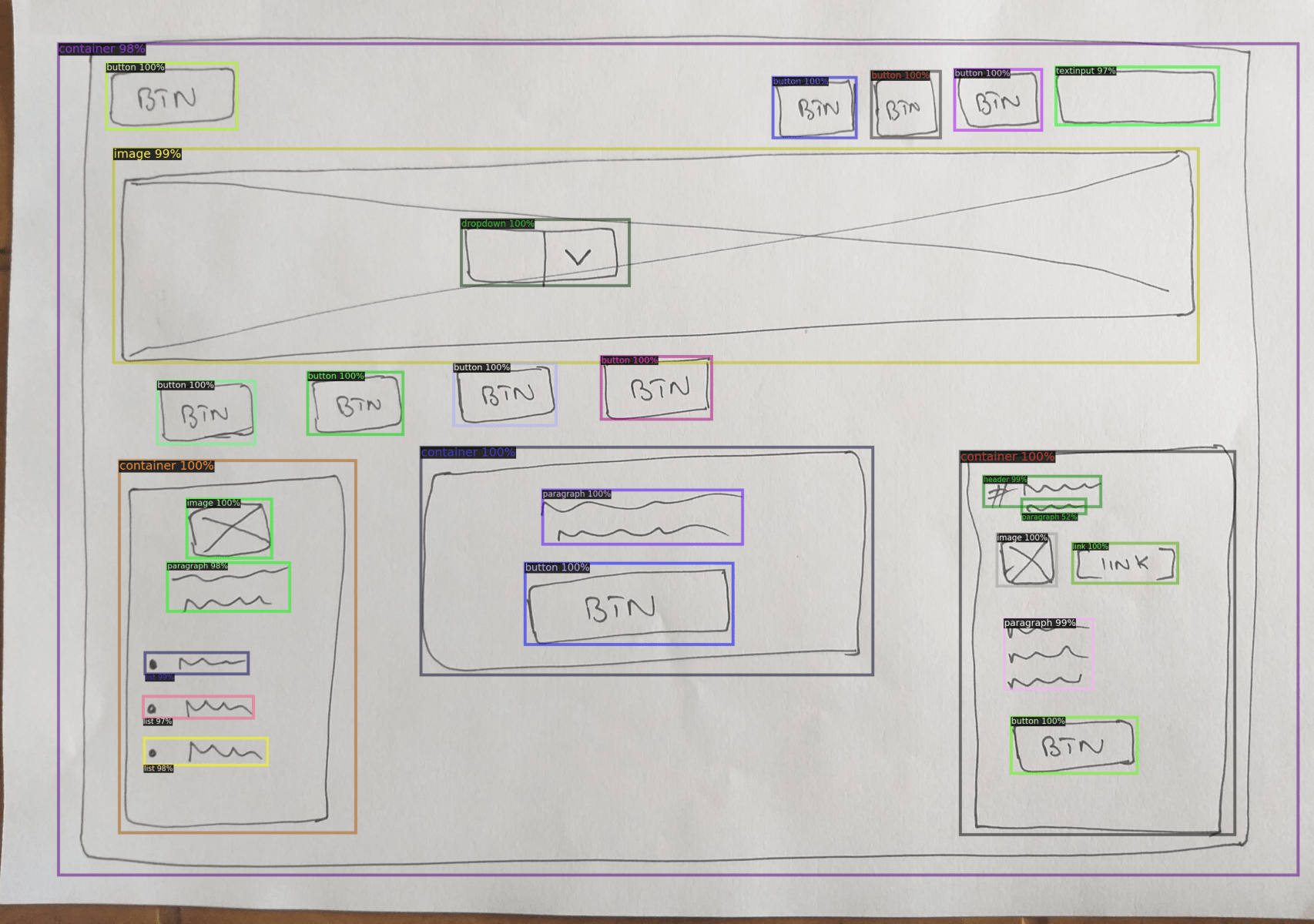} &  \includegraphics[width=\imgwl]{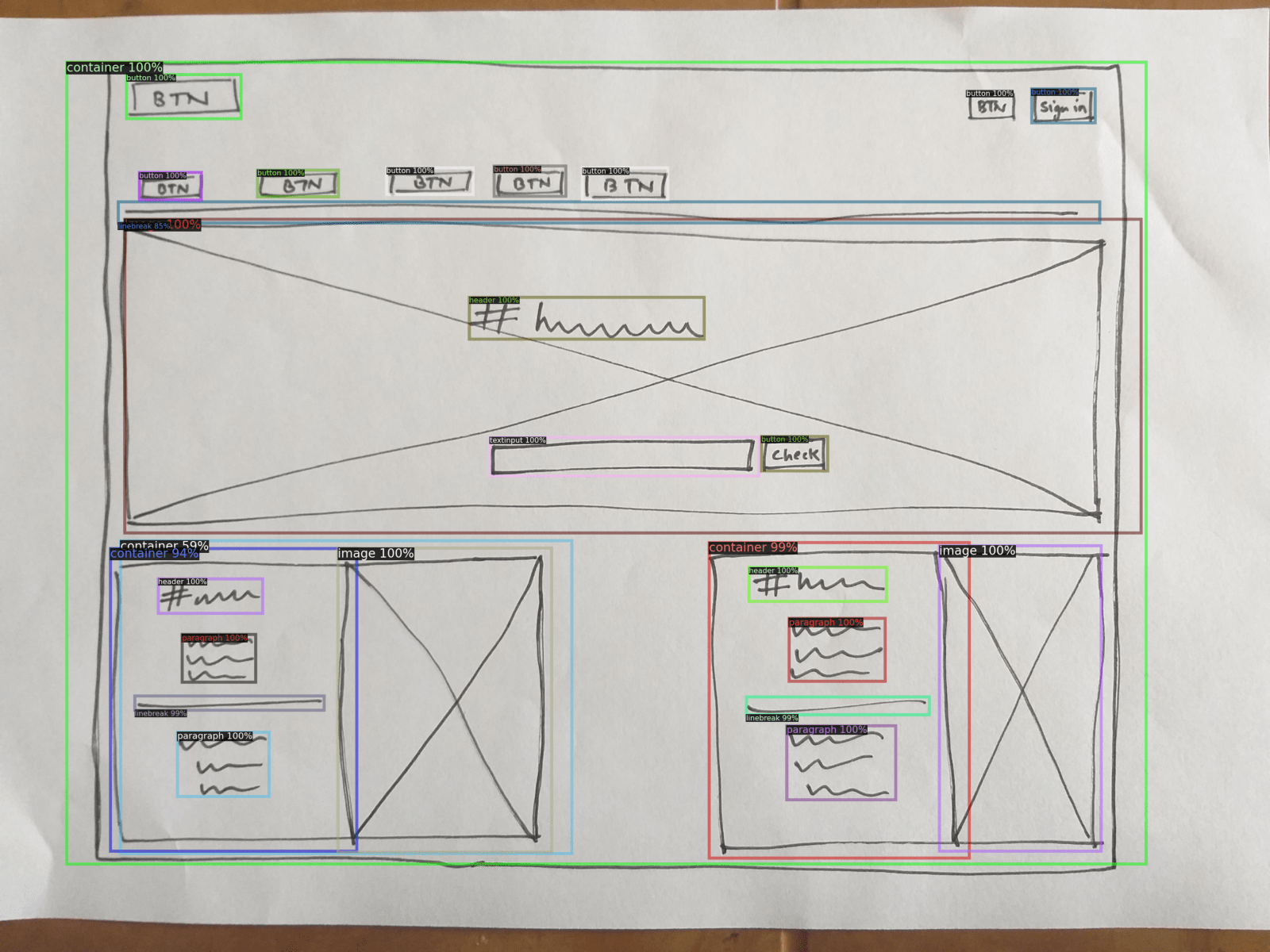} \\
      \includegraphics[width=\imgwl] {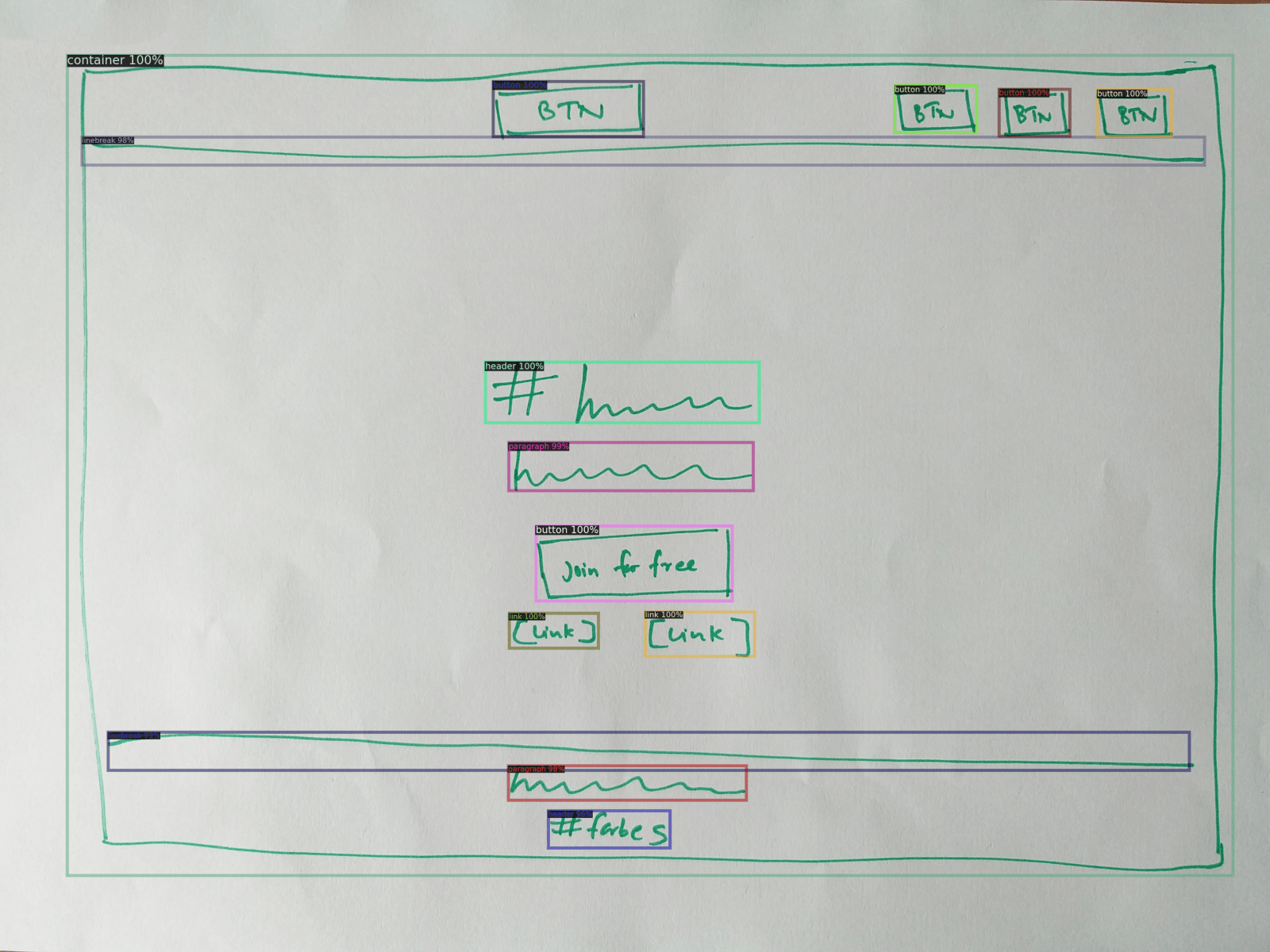}&   \includegraphics[width=\imgwl]{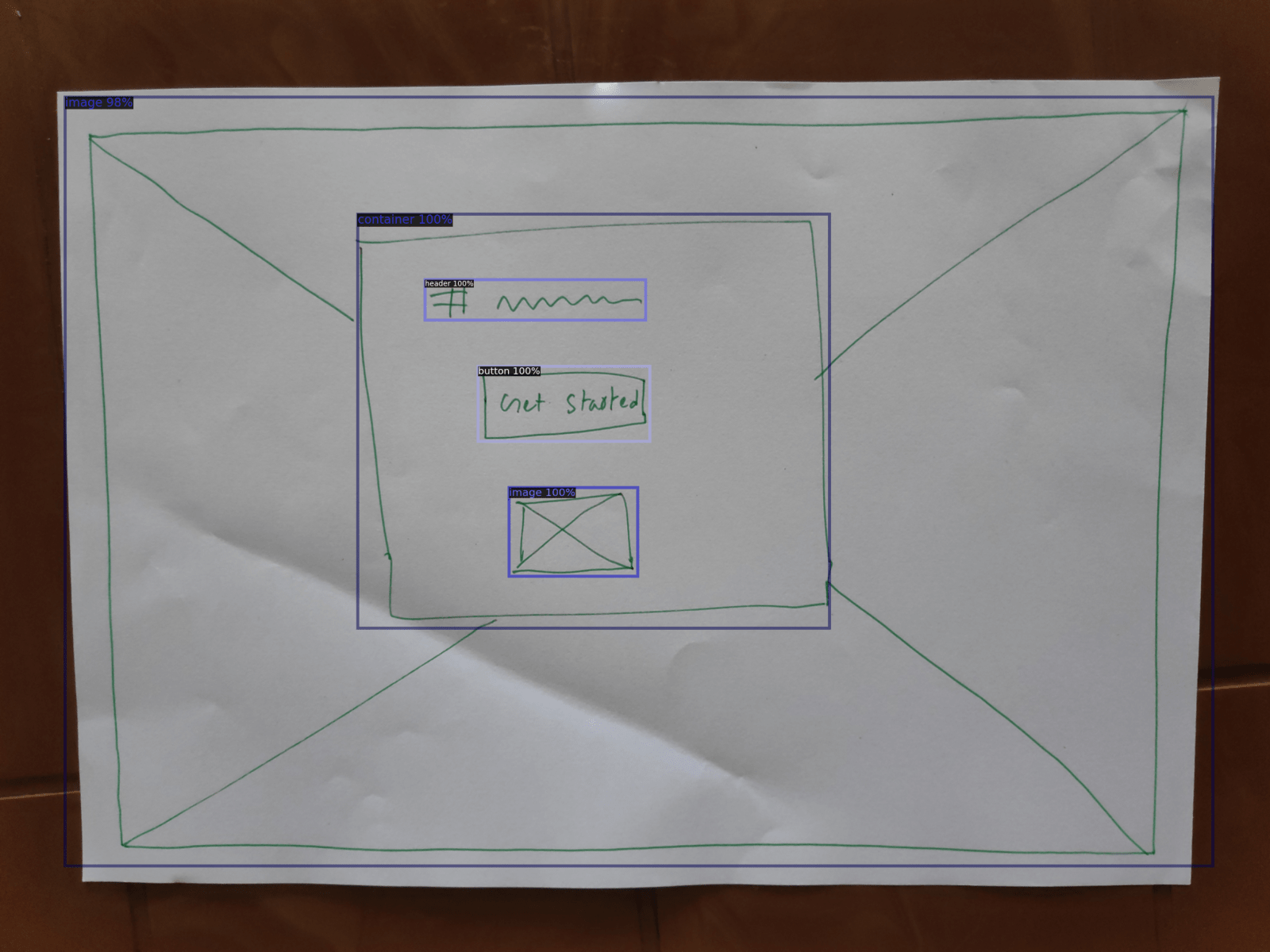} \\
      \includegraphics[width=\imgwl]{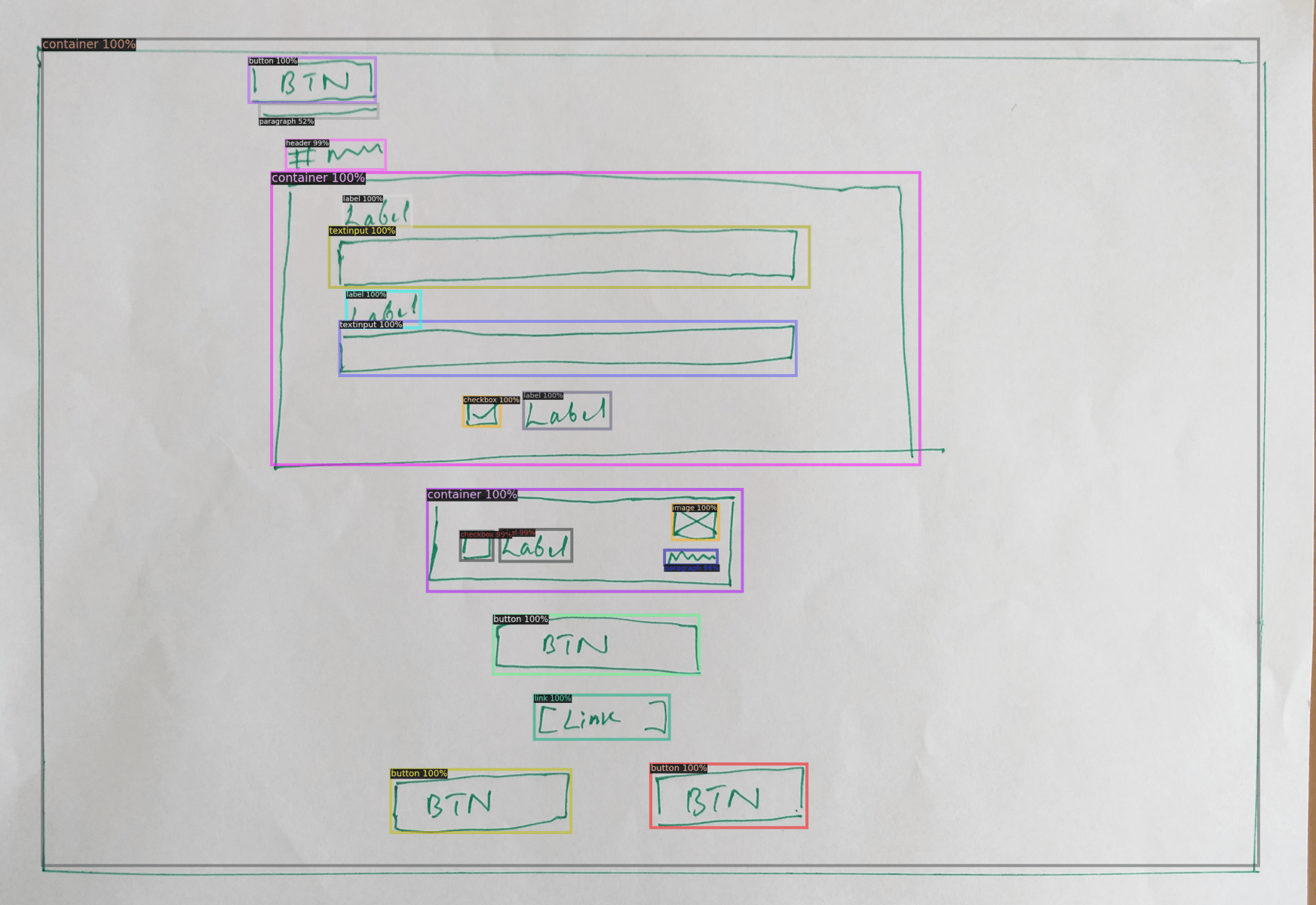} &   \includegraphics[width=\imgwl]{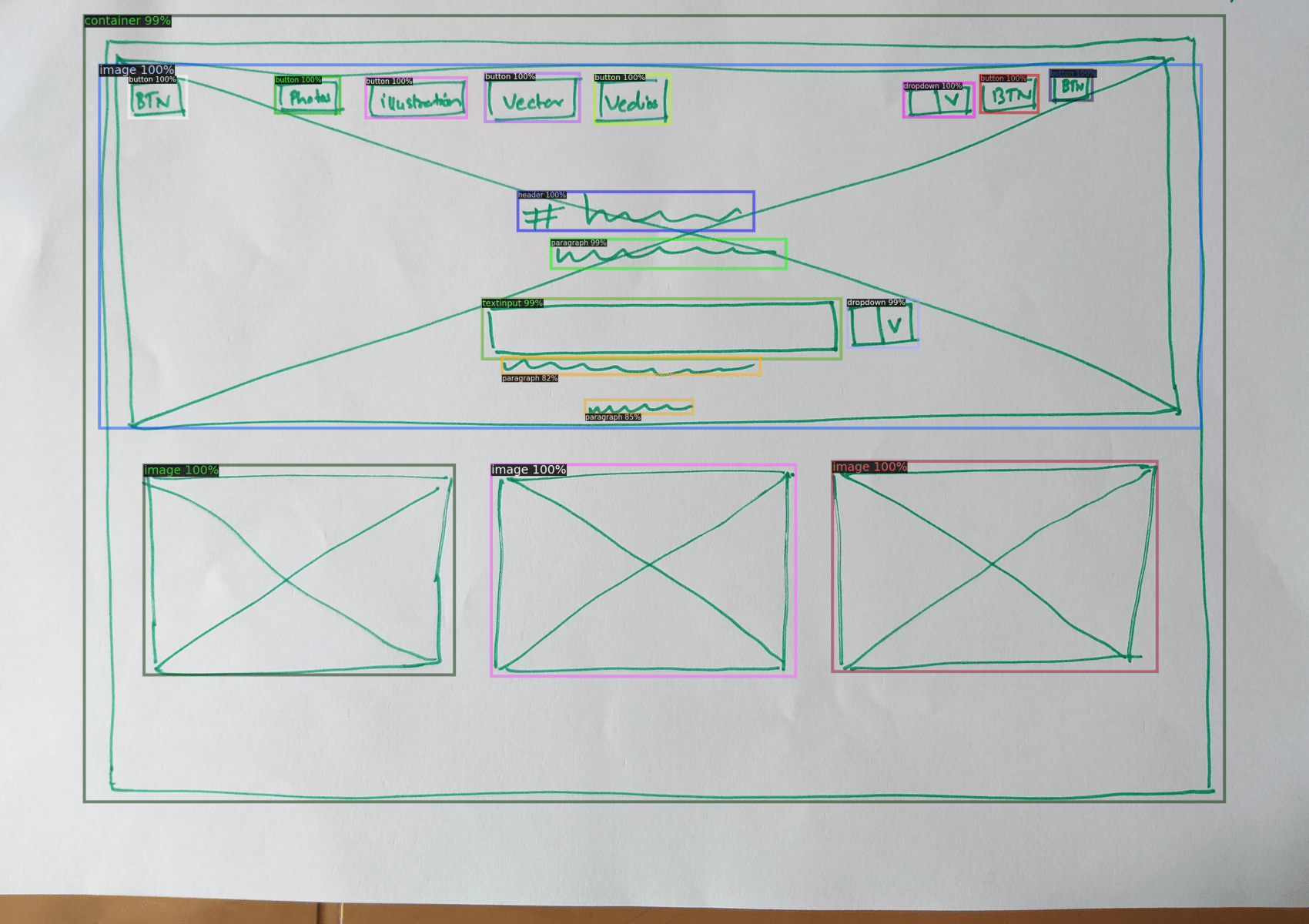} \\
      \includegraphics[width=\imgwl]{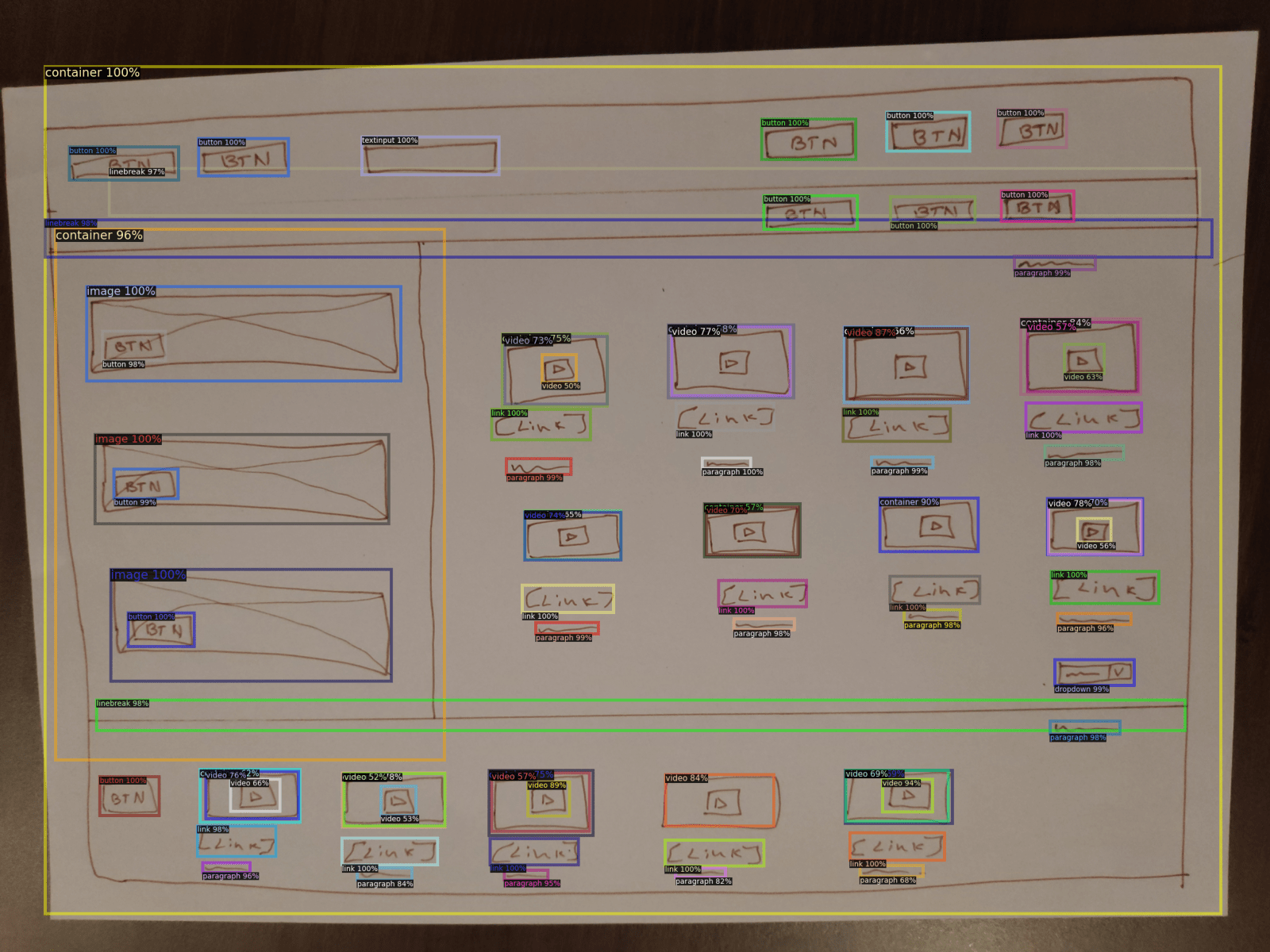} &    \includegraphics[width=\imgwl]{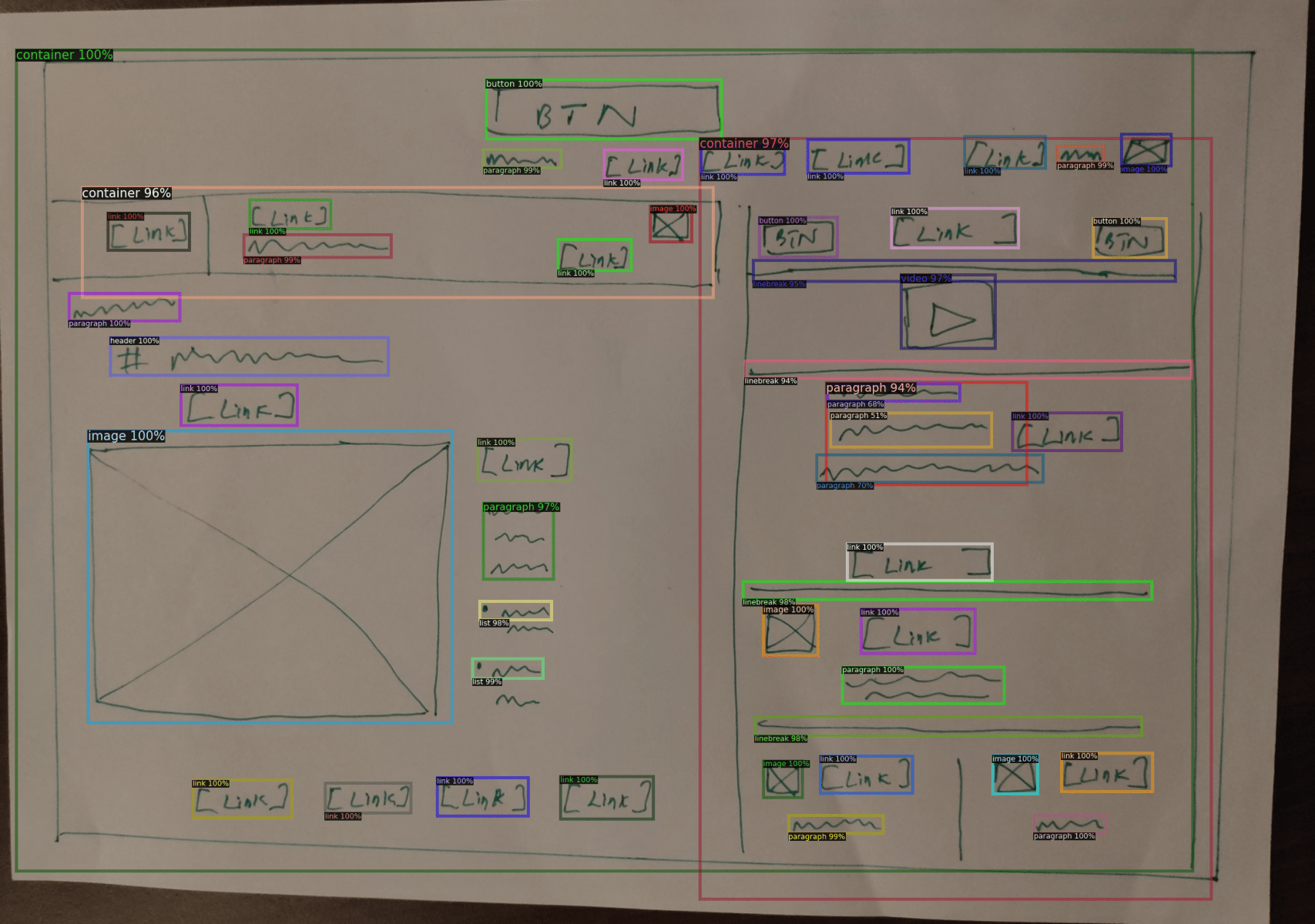} \\
    \end{tabular}   
    \caption{DrawnUI UX parsed by Magic Layouts - 2.}
    \label{fig:dui-2}
    \vspace{-3mm}
\end{figure*}


\end{document}